%% file: main.tex
\documentclass[10pt,twocolumn,letterpaper]{article}

\usepackage{cvpr}              

\input{preamble}

\definecolor{cvprblue}{rgb}{0.21,0.49,0.74}
\usepackage[pagebackref,breaklinks,colorlinks,allcolors=cvprblue]{hyperref}

\def\paperID{} 
\def\confName{CVPR}
\def\confYear{2026}

\title{\methodName: Universal Metric Depth Estimation for Any Camera}

\author{Girish Chandar Ganesan$^1$\quad Yuliang Guo$^2$\quad Liu Ren$^2$\quad Xiaoming Liu$^{1,3}$\\
{\small$^{1}$Michigan State University}
{\small$^{2}$Bosch Research North America \& Bosch Center for AI (BCAI)}\\
{\small$^{3}$University of North Carolina at Chapel Hill}\\
{\small\tt ganesang@msu.edu}\quad
{\small\tt [yuliang.guo2,liu.ren]@us.bosch.com}\quad
{\small\tt liuxm@cs.unc.edu}\\
{\small\url{https://girish1511.github.io/UniDAC}}\vspace{-3pt}
}

\begin{document}
\maketitle


\input{sec/0_abstract}    
\input{sec/1_intro}

\input{sec/2_related_works}
\input{sec/3_prelim}
\input{sec/4_methodology}
\input{sec/5_experiments}
\input{sec/6_conclusion}
\newpage
{
    \small
    \bibliographystyle{ieeenat_fullname}
    \bibliography{main}
}
\input{sec/X_suppl}

\end{document}

%% file: preamble.tex
\usepackage{pifont}
\usepackage{multirow}
\usepackage{tikz}
\usepackage[table]{xcolor}

\newcommand{\cmark}{\ding{51}}  

\newcommand{\thatIs}{\textit{i.e.}\xspace}

\newcommand{\kitti}{KITTI\xspace}
\newcommand{\nuscenes}{nuScenes\xspace}

\newcommand{\nyuDepth}{NYU-v2\xspace}

\newcommand{\argoverseTwo}{Argoverse2\xspace}

\newcommand{\kittiThreeSixty}{KITTI-360\xspace}

\newcommand{\ddad}{DDAD\xspace}
\newcommand{\aTwoDTwo}{A2D2\xspace}

\newcommand{\taskonomy}{Taskonomy\xspace}
\newcommand{\ibims}{IBims-1\xspace}

\newcommand{\hmThreeD}{HM3D\xspace}
\newcommand{\hypersim}{Hypersim\xspace}
\newcommand{\lyft}{LYFT\xspace}
\newcommand{\scannetpp}{ScanNet++\xspace}
\newcommand{\mThreeD}{Matterport3D\xspace}
\newcommand{\panoGVTwo}{Pano3D-GV2\xspace}
\newcommand{\aiMotive}{aiMotive\xspace}

\newcommand{\methodName}{UniDAC\xspace}

\newcommand{\image}{\mathbf{I}}

\newcommand{\depthMap}{\mathbf{D}}
\newcommand{\encoder}{\mathcal{E}}
\newcommand{\feature}{\mathbf{F}}

\newcommand{\loss}{\mathcal{L}}

\newcommand{\realDomain}{\mathbb{R}}
\newcommand{\transpose}{\intercal}
\newcommand{\relSym}{\text{rel}}
\newcommand{\metricSym}{\text{m}}
\newcommand{\scale}{s}
\newcommand{\shift}{t}

\newcommand{\query}{\mathbf{q}}

\newcommand{\complexDomain}{\mathbb{C}}
\newcommand{\pixel}{\mathbf{p}}
\newcommand{\rotMat}{\mathbf{R}}
\newcommand{\scaleMap}{\mathbf{S}}
\newcommand{\weightMap}{\mathbf{W}}
\newcommand{\deltaPixel}{\delta_\pixel}
\newcommand{\distMap}{\Delta}
\newcommand{\shiftMap}{\mathbf{T}}

\newcommand{\absRel}{A.Rel\xspace}

\newcommand{\rmse}{RMSE\xspace}

\newcommand{\siLog}{SI\textsubscript{log}\xspace}

\newcommand{\di}[1]{\delta_{#1}\xspace}
\newcommand{\dOne}{\di{1}\xspace}

\newcommand{\iDisc}{iDisc\xspace}
\newcommand{\unidepth}{UniDepth}

\newcommand{\metricThreeD}{Metric3D}

\newcommand{\uniKThreeD}{UniK3D\xspace}
\newcommand{\dac}{DAC\xspace}
\newcommand{\dacIn}{DAC\textsubscript{I}\xspace}
\newcommand{\dacOut}{DAC\textsubscript{O}\xspace}
\newcommand{\dacUni}{DAC\textsubscript{U}\xspace}

\newcommand{\Paragraph}[1]{\vspace{2mm}\noindent\textbf{#1.}\hspace{0.5mm}}

\newcommand{\lidar}{LiDAR\xspace}

%% file: sec/0_abstract.tex
\begin{abstract}
Monocular metric depth estimation (MMDE) is a core challenge in computer vision, playing a pivotal role in real-world applications that demand accurate spatial understanding.
Although prior works have shown promising zero-shot performance in MMDE, they often struggle with generalization across diverse camera types, such as fisheye and $360^\circ$ cameras.
Recent advances have addressed this through unified camera representations or canonical representation spaces, but they require either including large-FoV camera data during training or separately trained models for different domains.
We propose \methodName, an MMDE framework that presents universal robustness in all domains and generalizes across diverse cameras using a single model.
We achieve this by decoupling metric depth estimation into relative depth prediction and spatially varying scale estimation, enabling robust performance across different domains.
We propose a lightweight Depth-Guided Scale Estimation module that upsamples a coarse scale map to high resolution using the relative depth map as guidance to account for local scale variations.
Furthermore, we introduce RoPE-$\phi$, a distortion-aware positional embedding that respects the spatial warping in Equi-Rectangular Projections (ERP) via latitude-aware weighting.
\methodName achieves state of the art (SoTA) in cross-camera generalization by consistently outperforming prior methods across all datasets.
\end{abstract}

%% file: sec/1_intro.tex
\section{Introduction}
\label{sec:intro}

Depth estimation plays a pivotal role in bridging the gap between 2D and 3D vision, enabling a wide range of applications such as autonomous vehicles~\cite{wang2019pseudo, park2021pseudo}, robotics~\cite{dong2022towards,sabnis2011single}, and AR/VR~\cite{du2020depthlab, sari2023depth}.

\begin{figure}
    \centering
    \begin{tikzpicture}
        \draw (0,0) node[inner sep=0] {\includegraphics[width=\linewidth]{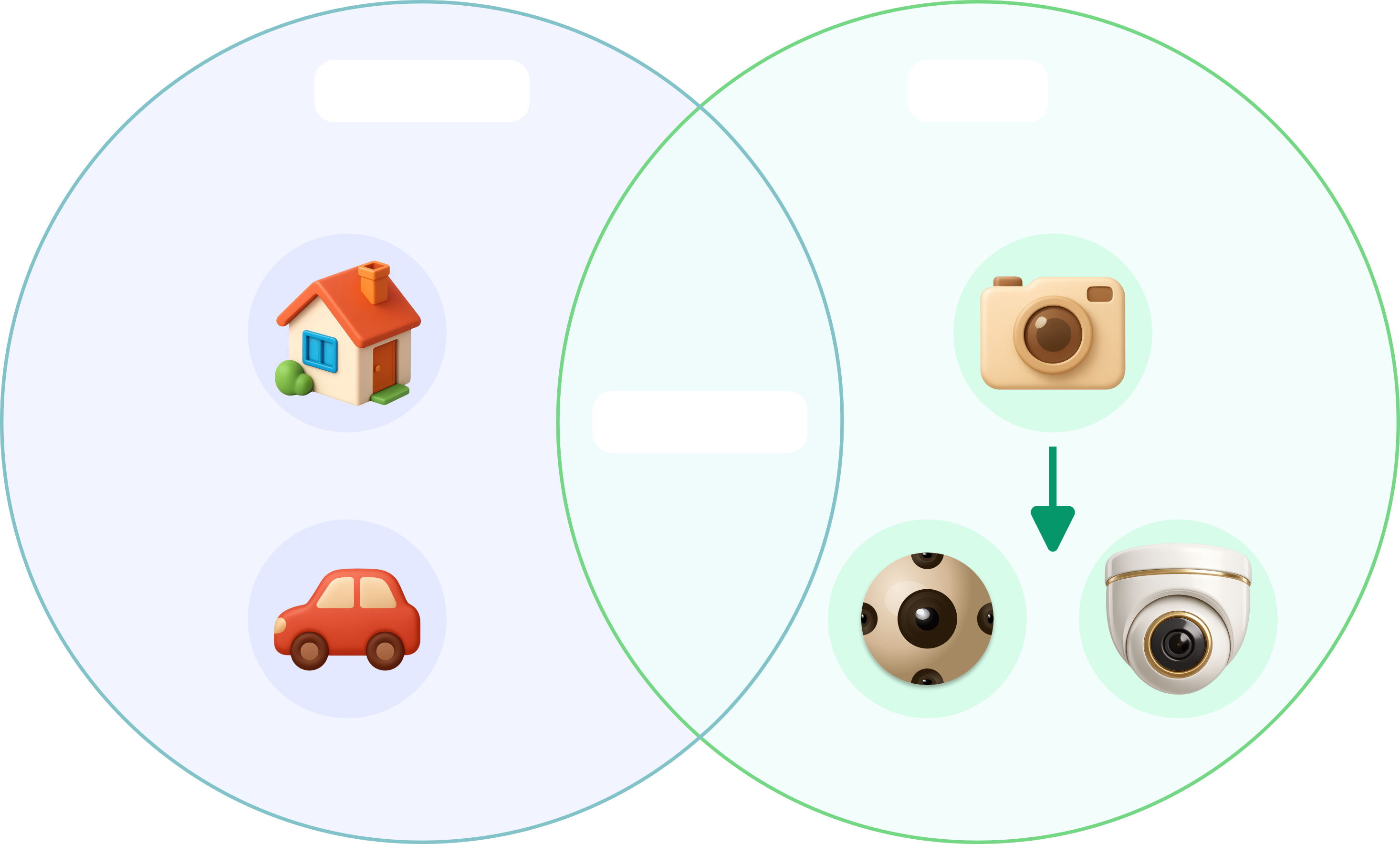}};


        

        \draw (1.65,1.98) node[inner sep=0, align=center] {\fontsize{10.0}{10}\selectfont \dac};
        \draw (-1.65,1.98) node[inner sep=0, align=center] {\fontsize{10.0}{10}\selectfont \uniKThreeD};
        \draw (0,0) node[inner sep=0, align=center] {\fontsize{10.0}{10}\selectfont \methodName};

        \draw (-2.1,1.25) node[inner sep=0, align=center] {\fontsize{8.0}{10}\selectfont Indoor};
        \draw (-2.1,-0.45) node[inner sep=0, align=center] {\fontsize{8.0}{10}\selectfont Outdoor};
        \draw (2.1,1.25) node[inner sep=0, align=center] {\fontsize{8.0}{10}\selectfont Persp.};
        \draw (1.4,-0.43) node[inner sep=0, align=center] {\fontsize{8.0}{10}\selectfont $360^\circ$};
        \draw (2.8,-0.45) node[inner sep=0, align=center] {\fontsize{8.0}{10}\selectfont Fisheye};
        
        \draw (-1.5,-3) node[inner sep=0, text width=3cm, align=center] {\fontsize{10.0}{10}\selectfont Universal Domain Robustness};
        \draw (1.9,-3) node[inner sep=0, text width=2.5cm, align=center] {\fontsize{10.0}{10}\selectfont Cross-Camera Generalization};
    \end{tikzpicture}
    
    \caption{We propose \textbf{\methodName}, a universal, domain-agnostic metric depth estimation framework that generalizes to any camera. Unlike prior methods that either rely on large-FoV data during training or require separate models for indoor and outdoor domains, \methodName is trained solely on perspective images yet generalizes effectively to large-FoV inputs, leveraging a universal model to robustly handle both indoor and outdoor environments.}
    \label{fig:teaser}
\end{figure}
\begin{figure*}
    \centering
    \begin{tikzpicture}
        \draw (0,0) node[inner sep=0] {\includegraphics[width=\linewidth]{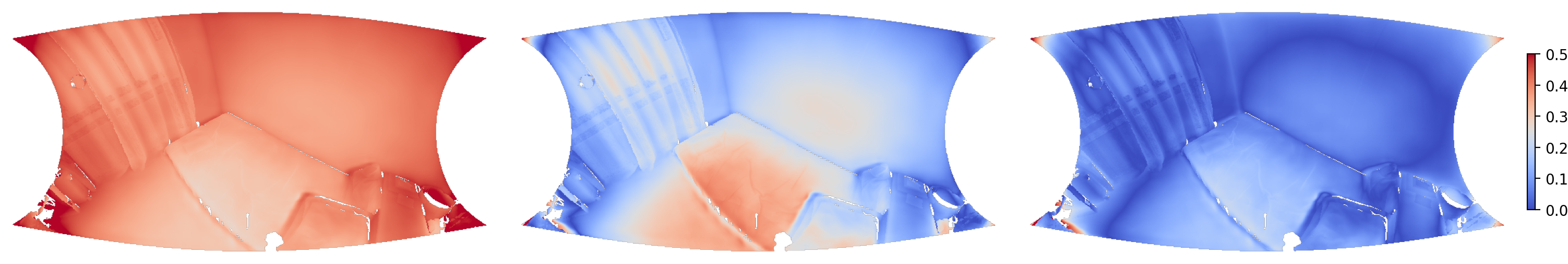}};
        \draw (-6,-1.5) node[inner sep=1, fill=white, fill opacity=0.75, text opacity=1, anchor=center] {\fontsize{8.0}{10}\selectfont (a) No Scaling};
        \draw (-0.25,-1.5) node[inner sep=1, fill=white, fill opacity=0.75, text opacity=1, anchor=center] {\fontsize{8.0}{10}\selectfont (b) Median Scaling};
        \draw (-0.25,1) node[inner sep=1, fill=white, fill opacity=0.75, text opacity=1, anchor=center] {\fontsize{8.0}{10}\selectfont $\scale \in \realDomain$};
        \draw (5.4,-1.5) node[inner sep=1, fill=white, fill opacity=0.75, text opacity=1, anchor=center] {\fontsize{8.0}{10}\selectfont (c) Depth-Guided Scaling};
        \draw (5.4,1) node[inner sep=1, fill=white, fill opacity=0.75, text opacity=1, anchor=center] {\fontsize{8.0}{10}\selectfont $\scaleMap \in \realDomain^{H\times W}$};
    \end{tikzpicture}
    \caption{We show the Abs.Rel error between the predicted relative depth and the ground truth by performing (a) no scaling, (b) median scaling, and (c) depth guided scaling. Theoretically, the relative and metric depth are aligned with a single scale $\scale$. Practically, (b) we observe that the irregularities in the relative depth cannot be compensated with a single scalar $\scale$. Thus, to tackle this, (c) we propose a Depth-Guided Scale Estimation module that predicts a high-resolution scale map $\scaleMap$ respecting local variations.}
    \label{fig:motivation}
\end{figure*}
With the advent of deep learning, early works \cite{piccinelli2023idisc, bhat2021adabins, lee2019big, yuan2022neural,eigen2014depth} achieve promising results. 
However, they fail to generalize across diverse scenes due to scale-depth ambiguity.
Recent methods \cite{ranftl2020towards, yin2021learning, yang2024depth, yang2024depthv2, ke2024repurposing} address this limitation by predicting depth up to an unknown scale in addition to leveraging large-scale data \cite{ranftl2020towards, yang2024depth, yang2024depthv2} and diffusion priors \cite{ke2024repurposing, fu2024geowizard, ke2025marigold}.
However, despite their success, these up-to-scale methods encounter limitations in real-world applications where precise spatial measurements are essential, necessitating the inherently challenging and ill-constrained task of monocular metric depth estimation.

Recent works \cite{guizilini2023towards, piccinelli2024unidepth, piccinelli2025unidepthv2, hu2024metric3d, yin2023metric3d, wang2025moge} have demonstrated zero-shot performance in MMDE by conditioning upon camera parameters~\cite{piccinelli2024unidepth, piccinelli2025unidepthv2}, adopting a canonical space \cite{yin2023metric3d, hu2024metric3d} or explicitly estimating the scale \cite{wang2025moge}, thereby overcoming the scale-depth ambiguity.
However, despite their remarkable performance, these methods have been developed with images from perspective cameras and thereby struggle with large FoV images captured from fisheye or 360$^\circ$ cameras.

Recently, \cite{piccinelli2025unik3d, guo2025depth} demonstrate remarkable MMDE performance on diverse cameras.
UniK3D~\cite{piccinelli2025unik3d} introduces a unified angular representation based on spherical harmonics to represent diverse camera models and subsequently condition the metric depth prediction.
While UniK3D achieves strong cross-dataset generalization with a single model covering both indoor and outdoor domains, it relies on training with a diverse set of camera models, including large-FoV data that closely resemble the test distribution.
In contrast, DAC~\cite{guo2025depth} achieves strong cross-camera generalization from perspective-only training by adopting an Equi-Rectangular Projection (ERP) as a canonical representation space, along with FoV-aligned and multi-resolution training strategies.
However, DAC requires separate models for indoor and outdoor domains, and its performance degrades when using a single unified model across both domains—partly due to its smaller-scale training compared to UniK3D.

In this work, we ask: \textit{Can we design a monocular metric depth estimation (MMDE) framework that generalizes effectively across all camera types and scene domains, without substantially increasing model or dataset scale?} 
We propose \methodName, a unified MMDE framework that generalizes across both diverse camera geometries and visual domains within a single model, as illustrated in \cref{fig:teaser}.
Our key insight is to achieve scene–camera unification by decomposing metric depth estimation into relative depth prediction and scale estimation - two components that inherently rely on different contextual scopes and benefit differently from pre-trained foundation model backbones.
To this end, we partition encoder features into early and late-stage representations: the former captures local structural details well-suited for relative depth, while the latter encodes global, scene-level context crucial for estimating scale.

Although metric and relative depth are theoretically connected by a single global scalar (scene scale), in practice, relative depth often exhibits spatially non-uniform scaling across the scene, as illustrated in \cref{fig:motivation}.
To handle this, we introduce a Depth-Guided Scale Estimation (DGSE) module that first predicts a coarse, low-resolution scale map from global encoder features.
We then upsample this map using the predicted relative depth as a non-parametric guidance signal, ensuring that spatially coherent regions share consistent scale values—while adding negligible computational overhead.

To better harness the potential of modern transformer architectures for large FoV images, we propose RoPE-$\phi$, a distortion-aware rotary positional embedding specifically designed for the Equi-Rectangular Projection (ERP) domain.
By incorporating latitude-aware weighting, RoPE-$\phi$ ensures that positional distances more faithfully correspond to geodesic proximity on the sphere.
Comprehensive experiments demonstrate that \methodName achieves robust generalization across both scenes and cameras, substantially outperforming prior SoTA approaches.

We summarize our contributions as follows:
\begin{itemize}
    \item[\cmark] We propose \methodName, a unified framework for monocular metric depth estimation that generalizes across diverse camera models and scene types using a single model.
    \item[\cmark] We design a Depth-Guided Scale Estimation module that generates a spatially adaptive scale map with minimal computational overhead. 
    \item[\cmark] We introduce RoPE-$\phi$, a distortion-aware RoPE, that respects the ERP geometry via latitude-based weighting.
    \item[\cmark] Extensive experiments show that \methodName achieves state-of-the-art cross-camera generalization in MMDE.
\end{itemize}

%% file: sec/2_related_works.tex
\section{Related Works}
\label{sec:rel_works}

\subsection{Perspective Monocular Depth Estimation}
Monocular depth estimation can be broadly categorized into affine-invariant, up-to-scale depth estimation, and absolute, metric depth estimation.

Prior works~\cite{ranftl2020towards, yin2021learning, yang2024depth, yang2024depthv2, ke2024repurposing} achieve notable generalization performance by leveraging large-scale labeled data ~\cite{ranftl2020towards, yin2021learning}, extensive unlabeled data ~\cite{yang2024depth, yang2024depthv2}, or strong generative priors~\cite{ke2024repurposing} of pre-trained diffusion models~\cite{rombach2022high}. 
However, despite their success, these methods encounter limitations in real-world applications where accurate metric depth is essential.

Recent works~\cite{guizilini2023towards, hu2024metric3d, yin2023metric3d, piccinelli2024unidepth, piccinelli2025unidepthv2} exhibit robustness in the zero-shot monocular metric depth estimation.
~\cite{guizilini2023towards,yin2023metric3d,hu2024metric3d} implicitly infer the scale of the scene by instilling the ground-truth camera parameters~\cite{guizilini2023towards} or mapping the camera parameters to a canonical space~\cite{yin2023metric3d,hu2024metric3d}.
UniDepth~\cite{piccinelli2024unidepth, piccinelli2025unidepthv2} implicitly predicts the scale in the form of camera parameters and applies them as a condition for the metric depth estimation, achieving incredible zero-shot metric depth performance without the need for intrinsic parameters during inference.
Despite these advances, none of these methods achieves satisfactory zero-shot performance on large FoV images.

\subsection{Large FoV Monocular Depth Estimation}

Fisheye and $360^\circ$ capture rich contextual information due to large FoV and can thereby benefit the task of depth estimation \cite{journals/ral/JiangSZDH21/unifuse,conf/cvpr/LiGY0DR22/omnifusion,conf/eccv/ShenLLNZZ22/panoformer,conf/iccv/YunSLLR23/egformer,conf/cvpr/AiCCSW23/hrdfuse}.
Due to the scarcity of large-FoV depth data, initial works \cite{conf/nips/SuG17/deformcnn360, zhu2019deformableconvnetsv2deformable,xiong2024efficient,journals/ral/JiangSZDH21/unifuse,rey2022360monodepth,conf/cvpr/LiGY0DR22/omnifusion,conf/eccv/ShenLLNZZ22/panoformer,conf/iccv/YunSLLR23/egformer, feng2023simfirsimpleframeworkfisheye} are constrained to in-domain setting.
These works handle distortions by adjusting the convolutional kernels through deformable CNNs \cite{conf/nips/SuG17/deformcnn360, zhu2019deformableconvnetsv2deformable,xiong2024efficient}, partitioning ERP into segments \cite{journals/ral/JiangSZDH21/unifuse,rey2022360monodepth}, and employing transformers to model the distortions \cite{conf/cvpr/LiGY0DR22/omnifusion,conf/eccv/ShenLLNZZ22/panoformer,conf/iccv/YunSLLR23/egformer, feng2023simfirsimpleframeworkfisheye}.
Although these methods achieve remarkable in-domain performance, they struggle to generalize across diverse cameras.

\subsection{Cross-Camera Generalizable MMDE}
A key challenge in building a foundational model for diverse cameras is the lack of large-scale, large-FoV datasets. UniK3D~\cite{piccinelli2025unik3d} combines perspective and large-FoV datasets using a spherical harmonics-based angular representation to share information across camera types. However, its performance is limited by training-time camera diversity and lacks cross-camera generalization. Recent methods~\cite{guo2025depth, gangopadhyay2025extending} achieve strong cross-camera generalization using only perspective images. \cite{gangopadhyay2025extending} extends UniDepth~\cite{piccinelli2024unidepth} to fisheye images via steering tokens but is restricted to perspective and fisheye inputs. DAC~\cite{guo2025depth} introduces a canonical Equi-Rectangular Projection (ERP) space with pitch-aware augmentation to simulate large-FoV images from perspective data, achieving SOTA cross-camera results. However, it requires separate models for indoor and outdoor domains. In contrast, \methodName achieves cross-camera and cross-domain generalization with a single unified model.

%% file: sec/3_prelim.tex
\begin{figure*}[!t]
    \centering
    \begin{tikzpicture}
    \draw (0,0) node[inner sep=0] {\includegraphics[width=\textwidth]{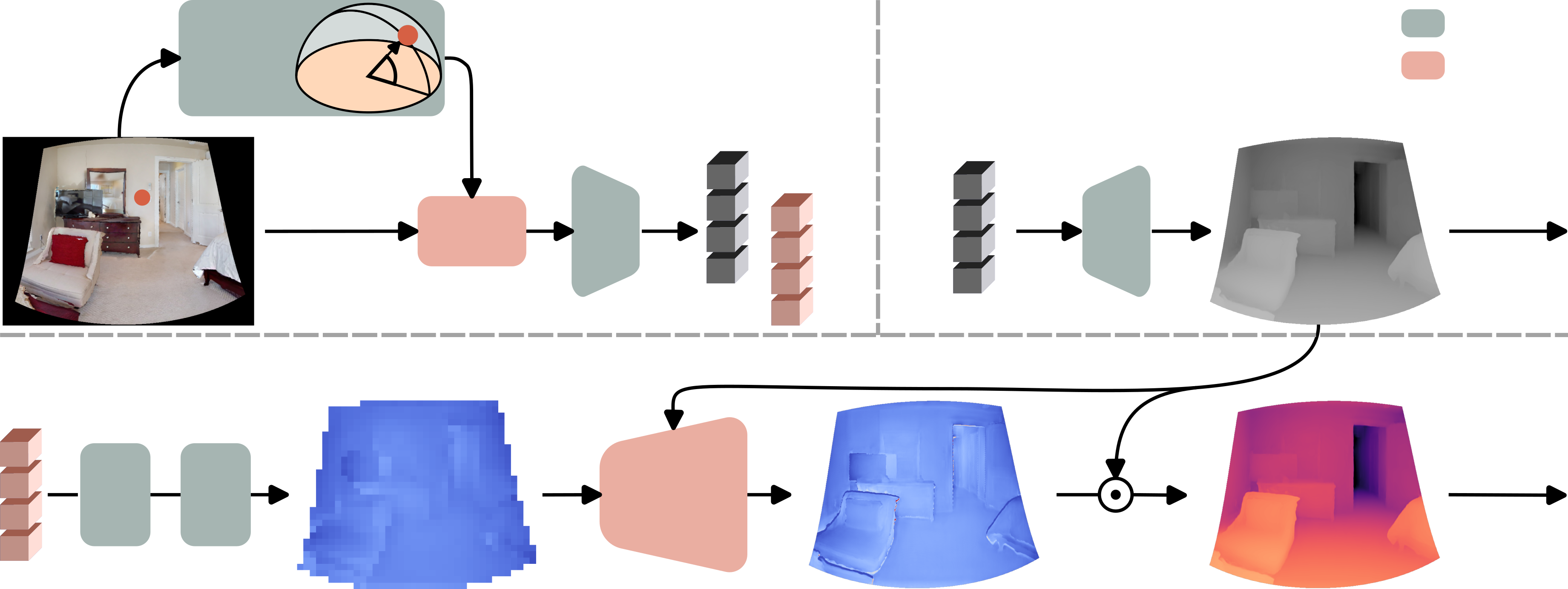}};
    \draw (-1.95,0.75) node[inner sep=0] {\fontsize{9.0}{10}\selectfont \textcolor{white}{Enc.}};
    \draw (4.4,0.75) node[inner sep=0, text width=2cm] {\fontsize{9.0}{10}\selectfont \textcolor{white}{Dec.}};
    \draw (-1.1,-2.25) node[inner sep=0, text width=1.5cm] {\fontsize{9.0}{10}\selectfont \textcolor{white}{Depth\\[-0pt] Guided\\[-2pt]Upsample}};
    \draw (-7.2,-2.2) node[inner sep=0.8, text width=1cm] {\fontsize{9.0}{10}\selectfont \textcolor{white}{Self\\[-2pt]Attn}};
    \draw (-5.9,-2.2) node[inner sep=0.8, text width=1.5cm] {\fontsize{9.0}{10}\selectfont \textcolor{white}{MLP}};
    \draw (-3.25,0.7) node[inner sep=0.8, text width=1.5cm] {\fontsize{9.0}{10}\selectfont \textcolor{white}{RoPE-$\phi$}};
    \draw (-1,2.8) node[inner sep=3, fill=white, fill opacity=0., text opacity=1, text width=2.65cm, line width=1, draw=gray, rounded corners=0.25cm] {\fontsize{10.0}{10}\selectfont Distortion-Aware Feature Extraction};
    \draw (3.5,2.8) node[inner sep=3, fill=white, fill opacity=0., text opacity=1, line width=1, draw=gray, rounded corners=0.25cm] {\fontsize{10.0}{10}\selectfont Relative Depth Estimation};
    \draw (-6.5,-0.85) node[inner sep=2.5, fill=white, fill opacity=0., text opacity=1, line width=1, draw=gray, rounded corners=0.25cm] {\fontsize{10.0}{10}\selectfont Depth-Guided Scale Estimation};
    \draw (8,3) node[inner sep=0, fill=white, fill opacity=0., text opacity=1] {\fontsize{9.0}{10}\selectfont Standard};
    \draw (8,2.5) node[inner sep=0, fill=white, fill opacity=0., text opacity=1] {\fontsize{9.0}{10}\selectfont Proposed};

    \draw (-8.65,1.55) node[inner sep=1, fill=white, fill opacity=0.75, text opacity=1, anchor=west] {\fontsize{10.0}{10}\selectfont $\image$};
    \draw (4.95,1.55) node[inner sep=1, fill=white, fill opacity=0.75, text opacity=1, anchor=west] {\fontsize{10.0}{10}\selectfont $\depthMap^\relSym$};
    \draw (4.95,-1.4) node[inner sep=1, fill=white, fill opacity=0.75, text opacity=1, anchor=west] {\fontsize{10.0}{10}\selectfont $\depthMap^\metricSym$};
    \draw (0.5,-1.4) node[inner sep=1, fill=white, fill opacity=0.75, text opacity=1, anchor=west] {\fontsize{10.0}{10}\selectfont $\scaleMap$};
    \draw (-5.15,-1.42) node[inner sep=1, fill=white, fill opacity=0.75, text opacity=1, anchor=west] {\fontsize{10.0}{10}\selectfont $\scaleMap_r$};

    \draw (-7.9,1.5) node[inner sep=1, fill=white, fill opacity=0.9, text opacity=1, anchor=west] {\fontsize{10.0}{10}\selectfont $\pixel = [u,v]$};
    \draw (-6.73,3) node[inner sep=1, fill=white, fill opacity=0., text opacity=1, anchor=west] {\fontsize{9.0}{10}\selectfont \textcolor{white}{Sph. Repr.}};
    \draw (-6.7,2.3) node[inner sep=1, fill=white, fill opacity=0., text opacity=1, anchor=west] {\fontsize{10.0}{10}\selectfont \textcolor{white}{$\phi = \frac{v\pi}{W}$}};
    \draw (-4.3,2.5) node[inner sep=1, fill=white, fill opacity=0.9, text opacity=1, anchor=west] {\fontsize{10.0}{10}\selectfont $\phi$};
    \draw (-4.4,1.5) node[inner sep=1, fill=white, fill opacity=0.75, text opacity=1, anchor=west] {\fontsize{10.0}{10}\selectfont $w(\phi)$};

    \draw (-0.3, 1.65) node[inner sep=1, fill=white, fill opacity=0., text opacity=1, anchor=west] {\fontsize{10.0}{10}\selectfont $\feature_l$};
    \draw (0.4, 1.15) node[inner sep=1, fill=white, fill opacity=0., text opacity=1, anchor=west] {\fontsize{10.0}{10}\selectfont $\feature_g$};

    \draw (7.7, 1.) node[inner sep=1, fill=white, fill opacity=0.75, text opacity=1, anchor=west] {\fontsize{10.0}{10}\selectfont $\loss_\relSym$};
    \draw (7.7, -1.9) node[inner sep=1, fill=white, fill opacity=0.75, text opacity=1, anchor=west] {\fontsize{10.0}{10}\selectfont $\loss_\metricSym$};
    \end{tikzpicture}
    
    \caption{\textbf{Overview of proposed method.} \methodName decouples metric depth estimation into relative depth and scale estimation. Relative depth relies on local scene information, while scene scale is domain-specific and depends on global scene information. Therefore, given an ERP image $\image$, we split the features from the encoder into local $\feature_l$ and the global features $\feature_g$. We predict the relative depth $\depthMap^\relSym$ using the local features $\feature_l$. We predict a scale map $\scaleMap$ from the global features $\feature_g$ to account for the irregularities in $\depthMap^\relSym$. We first predict a low-resolution scale map $\scaleMap_r$ and obtain the high-resolution $\scaleMap$ through our proposed Depth-Guided Scale (DGS) estimation module. The DGS upsamples $\scaleMap_r$ by using the $\depthMap^\relSym$ as a guide to ensure the upsampling process respects object boundaries. The final metric depth $\depthMap^\metricSym$ is calculated using $\depthMap^\relSym$ and $\scaleMap$ as shown in \cref{eq:final_metric_d}. We introduce distortion-aware positional embedding, termed RoPE-$\phi$, that applies a weight $w(\phi)$ to the RoPE rotations based on the latitude $\phi$. We train using two losses $\loss_\relSym$ and $\loss_\metricSym$ applied on $\depthMap^\relSym$ and $\depthMap^\metricSym$, respectively.}
    \label{fig:overview}
\end{figure*}

\section{Preliminaries}
\subsection{Metric Depth Decomposition}
\label{sec:met_decomp}
Given a metric depth map $\depthMap^\metricSym \in \realDomain^{H\times W}$, it can be decomposed into a relative depth map $\depthMap^\relSym \in \realDomain^{H\times W}$ and scale-shift scalars $\{\scale, \shift\} \in \realDomain$ as follows:
\begin{align}
    \depthMap^\metricSym = \scale \depthMap^\relSym + \shift.
    \label{eq:met_rel}
\end{align}
The relative depth map $\depthMap^\relSym$ contains local information, such as object shapes and boundaries, whereas the scale-shift scalars are global factors that depend on the scene as a whole, such as indoor or outdoor.
Therefore, the metric depth $\depthMap^\metricSym$ can be split into local and global components and are represented using $\depthMap^\relSym$ and $\{\scale, \shift\}$ respectively.

\subsection{RoPE}
\label{sec:vanilla_rope}
Rotary Positional Embedding (RoPE) was first introduced by \cite{su2024roformer} in the language domain.
RoPE embeds relative positional information among transformer tokens as position-dependent rotations.
Specifically, given the $n$-th token $\query_n \in \realDomain^{1\times d}$, RoPE rotates them using Euler's formula $e^{i\psi}$ to get $\query'_n = \query_ne^{in\psi}$, before applying the attention operations.
RoPE practically implements the rotations by first converting the tokens $\query_n$ to the complex domain, $\bar{\query_n} \in \complexDomain^{1\times d/2}$, in addition to using multiple frequencies $\psi_k$ across the channel dimensions.
The rotations for $N$ tokens are then compactly represented in a rotation matrix $\rotMat \in \complexDomain^{N\times d}$ defined as:
\begin{align}
    \rotMat(n,t) = e^{in\psi_k}.
    \label{eq:rope1d}
\end{align}
\cite{heo2024rotary} later proposes 2D-RoPE, designed to be compatible with vision transformers.
Given an image $\image \in \realDomain^{H\times W\times 3}$, \cite{heo2024rotary} replaces the 1D indexing $n$ to 2D pixel indices $\pixel_n = \{u_n,v_n\} \in \realDomain^{\{0,\dots,H\}\times\{0,\dots,W\}}$.
Thus, the rotation matrix in \cref{eq:rope1d} is modified to:
\begin{align}
    \centering
    \rotMat(n,2k) = e^{iu_n\psi_k}; \rotMat(n,2k+1) = e^{iv_n\psi_k},
    \label{eq:rot_mat_rope}
\end{align}
where, the frequencies $\psi_k$ are calculated as:
\begin{align}
    \psi_k = \{100^{-4k/d}: k\in \{0,\dots,d/4\}\}.
\end{align}

%% file: sec/4_methodology.tex
\section{\methodName}
\label{sec:method}
We propose \methodName, a unified framework for monocular metric depth estimation with cross-camera generalization.
We posit that one of the main issues with developing a unified model is the variation in depth ranges across different domains.
For example, the maximum depth is typically around 10 meters for indoor scenes, whereas it's 80 meters for outdoor scenes.
Thus, to prevent the model from getting confused trying to predict varied ranges, we decouple the metric depth prediction into domain-agnostic and domain-specific components, \thatIs, relative depth map and scene scale, respectively.
As mentioned in \cref{sec:met_decomp}, the relative depth map depends only on local pixel variations in the image; thus, we utilize features rich in local information to estimate relative depth as described in \cref{sec:rel_d_est}.
On the other hand, the scale and shift parameters depend on the scene as a whole, and we estimate it using the global information-rich features as detailed in \cref{sec:scale_est}.
To account for the irregularities in the predicted relative depth, we estimate a high-resolution scale map as opposed to a 1-D scale through our proposed Depth-Guided Scale Estimation module (\cref{sec:scale_est}).
Specifically, we achieve higher efficiency by first predicting a low-resolution scalar map (\cref{sec:patch_scale}) and followed by a depth-guided upsampling (\cref{sec:depth_guide_up}).
An overview of our pipeline is shown in \cref{fig:overview}.

\subsection{Relative Depth Estimation}
\label{sec:rel_d_est}
Given an input image $\image \in \realDomain^{H\times W\times3}$, we derive set of features $\mathcal{F} = \{\feature_l, \feature_g\} \in \realDomain^{h\times w\times C}$ from encoder $\encoder$, where $\feature_l,\feature_g$ are the local and global features respectively.
$\feature_l$ and $\feature_g$ are split as the outputs from the early and final layers of $\encoder$, respectively.
We then estimate the relative depth map $\bar{\depthMap^\relSym} \in \realDomain^{H\times W\times 1}$ from local features $\feature_l$ via a decoder $\mathcal{D}$.

The predicted relative depth map can have an arbitrary scale, which can affect the later scale estimation step.
We normalize $\hat{\depthMap^\relSym}$ by median scaling to get $\depthMap^\relSym$:
\begin{align}
    \depthMap^\relSym = \frac{\hat{\depthMap^\relSym}}{\hat{\scale}}; \;\;
    \hat{\scale} &= \texttt{Median}(\hat{\depthMap^\relSym}).
\end{align}

\subsection{Depth-Guided Scale Estimation}
\label{sec:scale_est}

The predicted relative depth map $\depthMap^\relSym$ requires global scalars, \thatIs, scale $\scale$ and shift $\shift$,  to be converted to metric depth.
The global scalars are low-dimensional terms dependent on the scene as a whole and are not affected much by local variations in the scene.
Therefore, we utilize the global features $\feature_g$ from \cref{sec:rel_d_est} to estimate $\scale$ and $\shift$.
Theoretically, the metric and relative depth are related to each other by a couple of 1-D scalars $\{\scale, \shift\}$ as shown in \cref{eq:met_rel}.
However, in practice, the relative depth predicted by a network may be slightly stretched or compressed in different regions due to local errors or occlusions.
Thus, to adjust for the irregularities in $\depthMap^\relSym$, we predict a scalar map $\scaleMap \in \realDomain^{H\times W}$.
A naive solution is to predict $\scaleMap$ through a series of transpose convolutions, but this would incur an additional computational cost.
Moreover, predicting a high-dimensional output $\scaleMap$ is challenging and may lead to poor estimates.
Thus, we propose a low-resolution scale prediction followed by a lightweight non-parametric depth-guided upsampling to estimate $\scaleMap$.
We predict $\shift$ from the \texttt{CLS} token of the $\feature_g$ by passing it through a shallow MLP.

\subsubsection{Patch-level Scale Estimation}
\label{sec:patch_scale}
The global features are low-resolution features corresponding to non-overlapping patches on the image $\image$.
Since we want similar features to have similar scales, we employ a self-attention on $\feature_g$ followed by a shallow-MLP to estimate a low-resolution scale map $\scaleMap_r \in \realDomain^{h\times w}$. Thus, we get $\scaleMap_r = \texttt{MLP}(\texttt{SelfAttn}(\feature_g))$.

$\scaleMap_r$ is a low-resolution spatial map that must be upsampled to obtain $\scaleMap$, while maintaining low computational overhead; hence, we adopt a non-parametric approach.
Nearest-neighbor upsampling provides a simple solution, but it does not respect boundaries while upsampling.
To ensure non-parametric upsampling while respecting inter-object boundaries, we utilize the predicted relative depth $\depthMap^\relSym$.

\begin{figure}
    \begin{tikzpicture}
        \draw (0,0) node[inner sep=0] {\includegraphics[width=\linewidth]{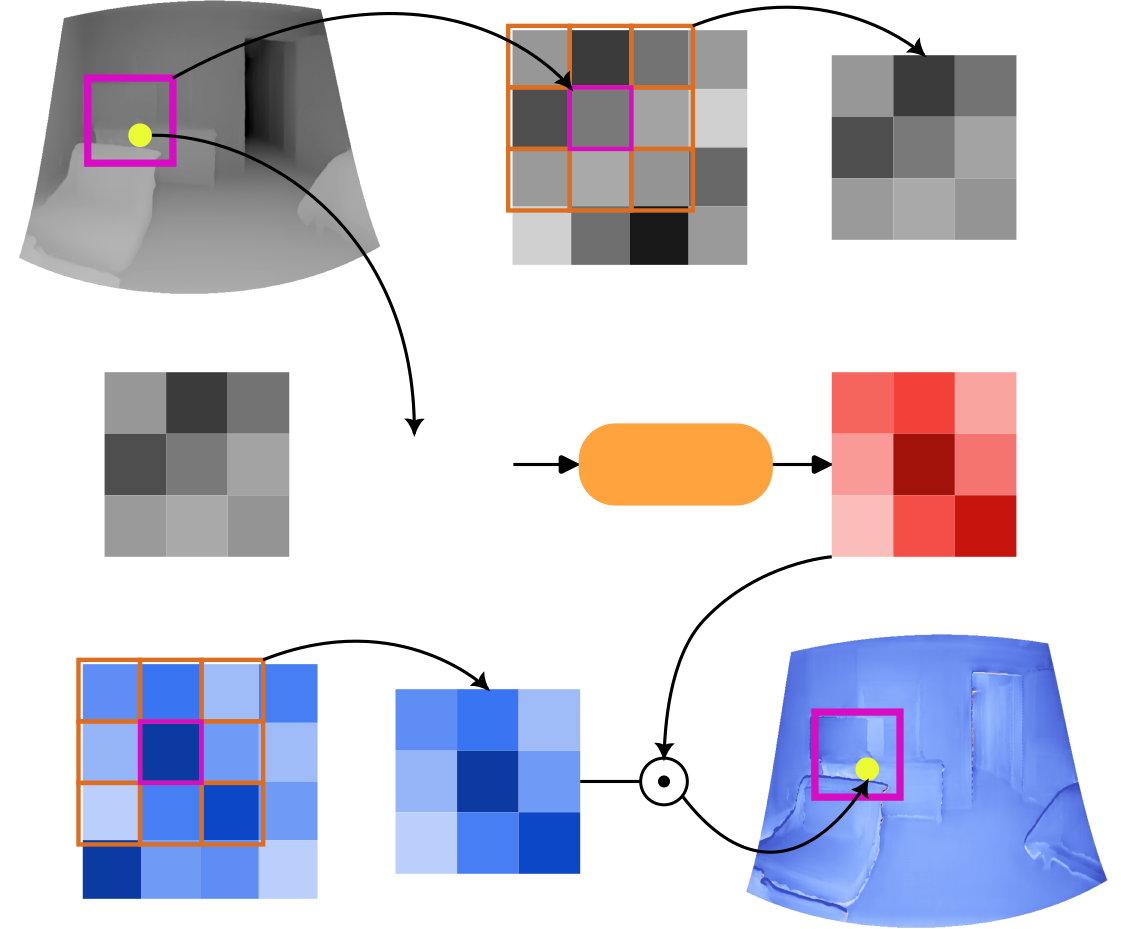}};
        \draw (-2.5,3.8) node[inner sep=0] {\fontsize{10.0}{10}\selectfont $\depthMap^\relSym$};
        \draw (-3.3,2.6) node[inner sep=0.8, fill=white, fill opacity=0.7, text opacity=1] {\fontsize{8.0}{10}\selectfont $\pixel$};
        \draw (0.5,3.8) node[inner sep=0] {\fontsize{10.0}{10}\selectfont $\depthMap^\relSym_r$};
        \draw (0.3,2.55) node[inner sep=0.8, fill=white, fill opacity=0.7, text opacity=1] {\fontsize{8.0}{10}\selectfont $\pixel_r$};
        \draw (2.7,3.8) node[inner sep=0] {\fontsize{10.0}{10}\selectfont $\depthMap^\relSym_r[\pixel_r+\delta_\pixel]$};
        \draw (-3.65,2.55) node[inner sep=0.8, fill=white, fill opacity=0.7, text opacity=1] {\fontsize{8.0}{10}\selectfont $r$};
        \draw (-3.2,2.1) node[inner sep=0.8, fill=white, fill opacity=0.7, text opacity=1] {\fontsize{8.0}{10}\selectfont $r$};

        \draw (-1,0) node[inner sep=0.8, fill=white, fill opacity=0.7, text opacity=1] {\fontsize{8.0}{10}\selectfont $\depthMap^\relSym[\pixel]$};
        \draw (-1.7,0) node[inner sep=0.8] {\fontsize{10.0}{10}\selectfont $-$};
        \draw (-3.7,0) node[inner sep=0.8] {\fontsize{10.0}{10}\selectfont $\Bigg($};
        \draw (-0.5,0) node[inner sep=0.8] {\fontsize{10.0}{10}\selectfont $\Bigg)$};
        \draw (0.88,0) node[inner sep=0.8] {\fontsize{8.0}{10}\selectfont SoftMax};
        \draw (2.7,1) node[inner sep=0.8] {\fontsize{8.0}{10}\selectfont $\weightMap[\pixel]$};

        \draw (-2.6,-1.1) node[inner sep=0] {\fontsize{8.0}{10}\selectfont $\scaleMap_r$};
        \draw (-2.898,-2.15) node[inner sep=0.8, fill=white, fill opacity=0.7, text opacity=1] {\fontsize{8.0}{10}\selectfont $\pixel_r$};
        \draw (-0.5,-1.1) node[inner sep=0] {\fontsize{8.0}{10}\selectfont $\scaleMap_r[\pixel_r+\delta\pixel]$};
        \draw (2.7,-1.1) node[inner sep=0] {\fontsize{8.0}{10}\selectfont $\scaleMap[\pixel]$};
        \draw (2.1,-2.1) node[inner sep=0.8, fill=white, fill opacity=0.7, text opacity=1] {\fontsize{8.0}{10}\selectfont $\pixel$};
    \end{tikzpicture}
    \centering
    
    \caption{\textbf{Depth-Guided Upsampling.} We leverage the predicted relative depth $\depthMap^\relSym$ as a guide to upsample the predicted low-resolution scale map $\scaleMap_r \in \realDomain^{\frac{H}{r}\times\frac{W}{r}}$ to get $\scaleMap \in \realDomain^{H\times W}$. We compare $\depthMap^\relSym$ and its downsampled version $\depthMap_r$ to get the local information in the form of weights $\weightMap \in \realDomain^{H\times W\times 9}$. We compare the spatial mapping between $\scaleMap$ and $\scaleMap_r$ and combine it with $\weightMap$ to obtain $\scaleMap$. The Depth-Guided Upsampling is non-parametric and thus does not add computational overhead.}
    \label{fig:dgu}
\end{figure}
\subsubsection{Depth-Guided Upsampling}
\label{sec:depth_guide_up}
The relative depth map $\depthMap^\relSym$ already has local boundary information, which we leverage as a guide to upsample $\scaleMap_r$.
We first obtain a low-resolution relative depth map $\depthMap^\relSym_r$ by applying median pooling on $\depthMap^\relSym$ with a kernel size and a stride of $r$.
Thus, every pixel $\pixel = [u,v]$ in $\depthMap^\relSym$ would be mapped to pixel $\pixel_r$ in $\depthMap^\relSym_r$ such that $\pixel_r = [\lfloor\frac{u}{r}\rfloor, \lfloor\frac{v}{r}\rfloor]$.
Then, with a neighborhood $\Omega =\{-1,0,1\}^2$ we obtain a distance matrix $\distMap \in \realDomain^{H\times W\times 9}$ defined as follows:
\begin{align}
    \distMap[\pixel] = \{|\depthMap^\relSym[\pixel] - \depthMap^\relSym_r[\pixel_r+\deltaPixel] | : \forall \; \deltaPixel \in \Omega\}.
\end{align}
We apply softmax for every pixel over the negative distance values in $\distMap$ to get a per-pixel weight map such that $\weightMap[\pixel] = \texttt{softmax}(-\distMap[\pixel]) \in \realDomain^{H\times W\times 9}$.
$\distMap$ and $\weightMap$ contain the pseudo mapping from low-resolution data ($\depthMap_r$) to high-resolution data ($\depthMap$).
We use this information as the routing signal to get the high-resolution scale data $\scaleMap$ from low-resolution scale data $\scaleMap_r$.
Finally, the scalar map $\scaleMap$ is calculated as the weighted summation over the low-resolution scales as follows:
\begin{align}
    \scaleMap[\pixel] = \weightMap[\pixel,:]^{\transpose}\mathcal{N}(\scaleMap_r, \pixel_r),
\end{align}
where, $\mathcal{N}(\scaleMap_r, p_r) = \{\scaleMap_r[\pixel_r+\deltaPixel] : \forall \; \deltaPixel \in \Omega\}$ is the neighborhood extractor.

The final metric depth $\depthMap^\metricSym$ is then estimated using the Hadamard product $\odot$ as follows:
\begin{align}
\depthMap^\metricSym = \scaleMap \odot\depthMap^\relSym  + \shift. 
\label{eq:final_metric_d}
\end{align}

\begin{figure}[!t]
    \centering
    \begin{tikzpicture}
        \draw (0,0) node[inner sep=0] {\includegraphics[width=0.9\linewidth]{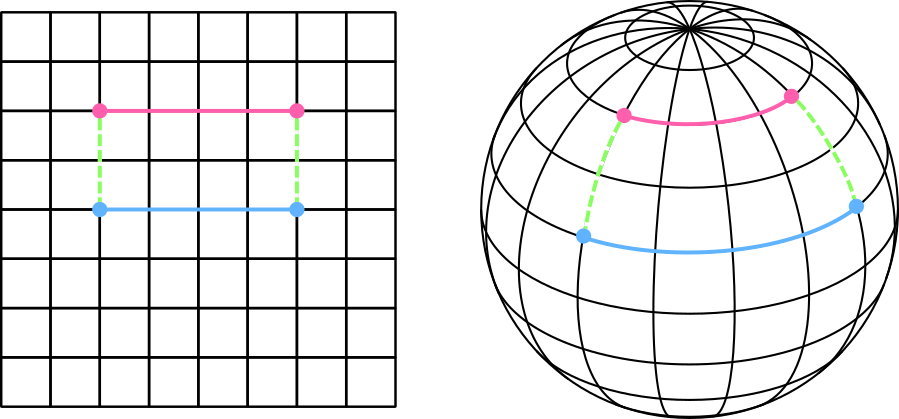}};
        \draw (-4,0.9) node[inner sep=0.8] {\fontsize{10.0}{10}\selectfont $\phi_1$};
        \draw (-4,0) node[inner sep=0.8] {\fontsize{10.0}{10}\selectfont $\phi_2$};
        \draw (-2.8,1.9) node[inner sep=0.8] {\fontsize{10.0}{10}\selectfont $\theta_1$};
        \draw (-1.2,1.9) node[inner sep=0.8] {\fontsize{10.0}{10}\selectfont $\theta_2$};


        \draw (-3,1.1) node[inner sep=0.8, fill=white, fill opacity=0.9, text opacity=1] {\fontsize{10.0}{10}\selectfont $\pixel_{11}$};
        \draw (-1,1.1) node[inner sep=0.8, fill=white, fill opacity=0.9, text opacity=1] {\fontsize{10.0}{10}\selectfont $\pixel_{12}$};
        \draw (-3,-0.2) node[inner sep=0.8, fill=white, fill opacity=0.9, text opacity=1] {\fontsize{10.0}{10}\selectfont $\pixel_{21}$};
        \draw (-1,-0.2) node[inner sep=0.8, fill=white, fill opacity=0.9, text opacity=1] {\fontsize{10.0}{10}\selectfont $\pixel_{22}$};

        \draw (1.9,1.2) node[inner sep=0.8, fill=white, fill opacity=0.9, text opacity=1] {\fontsize{10.0}{10}\selectfont $\mathcal{G}(\pixel_{11}, \pixel_{12})$};
        \draw (2.4,-0.7) node[inner sep=0.8, fill=white, fill opacity=0.9, text opacity=1] {\fontsize{10.0}{10}\selectfont $\mathcal{G}(\pixel_{21}, \pixel_{22})$};

        \draw (-2,-2) node[inner sep=0.8] {\fontsize{10.0}{10}\selectfont (a)};
        \draw (2,-2) node[inner sep=0.8] {\fontsize{10.0}{10}\selectfont (b)};
    \end{tikzpicture}
    \caption{\textbf{Motivation for RoPE-$\phi$.} We show the difference between (a) the pixel distance in ERP and (b) the corresponding geodesic distance on the curvature of the sphere. Although $|\pixel_{11}-\pixel_{12}| = |\pixel_{21}-\pixel_{22}|$ in the ERP, we see that $\mathcal{G}(\pixel_{11},\pixel_{12}) < \mathcal{G}(\pixel_{21},\pixel_{22})$ on the sphere. Geodesic distance respects the actual separation in the 3D space. Thus, we modify 2D-RoPE to reflect the geodesic distance to get RoPE-$\phi$.}
    \label{fig:rope_geodesic}
\end{figure}

\subsection{RoPE-\texorpdfstring{$\phi$}{phi}}
\label{sec:rope_phi}
We take a step back and observe the encoder $\encoder$.
$\encoder$ employs 2D-RoPE as the positional embedding scheme as defined in \cref{sec:vanilla_rope}.
The 2D-RoPE design ensures the relative positional embedding is dependent solely on the pixel locations $\pixel=[u,v]$.
While this is desired for tasks involving perspective images, it is not suitable for ERP representation, which encodes spherical data.
ERP is a 2D representation of the surface of the sphere.
The pixel distance in ERP is the warped distance between points on the curvature of a sphere, referred to as the geodesic distance.
We propose RoPE-$\phi$, an enhanced 2D-RoPE that respects the geodesic distance while applying the rotations in 2D-RoPE.

Every pixel in an ERP denotes a direction in the 3D space.
Specifically, given an ERP image $\image \in \realDomain^{H\times W\times 3}$, a pixel $\pixel_n=\{u_n,v_n\}$ denotes direction $\{\theta_n=\frac{u_n}{H}, \phi_n=\frac{v_n}{W}\}$, where $\theta_n,\phi_n$ are the longitude and latitude respectively.
Given two pixels $\{\pixel_1, \pixel_2\}$ = \{$\{\theta_1, \phi_1\},\{\theta_2, \phi_2\}\}$ in $\image$, and $\Delta\theta=|\theta_1-\theta_2|; \Delta\phi=|\phi_1-\phi_2|$, their geodesic distance on unit-sphere $\mathcal{G}(\pixel_1, \pixel_2)$ is defined as:
\begin{align}
    \mathcal{G}(\pixel_1, \pixel_2) = \smash{\arccos}{(\sin\phi_1\!\sin\phi_2\!+\!\cos\phi_1\!\cos\phi_2\Delta\theta)}.
    \label{eq:geodesic}
\end{align}
We observe from \cref{eq:geodesic} that when $\Delta\theta=0$, \thatIs, when the pixels are along the same longitude, $\mathcal{G}\propto\Delta\phi$, aligning with 2D-RoPE design.
However, when $\Delta\phi=0$, \thatIs, when the pixels are along the same latitude, we can approximate that $\mathcal{G}\propto\cos\phi\Delta\theta$.
In other words, given a pair of pixels on the same latitude $\phi$ and fixed longitudinal distance $\Delta\theta$, the geodesic distance reduces as we move towards the pole, even though the pixel distance in ERP remains the same, as shown in \cref{fig:rope_geodesic}.
With this observation, we add a latitude-based cosine weighting to the 2D-RoPE and modify the rotation matrix $\rotMat$ in \cref{eq:rot_mat_rope} to get $\rotMat_\phi$ as follows:
\begin{align}
    \centering
    \rotMat_\phi[n] = \rotMat[n]^{w(\phi_n)} \;;\; w(\phi) = \delta + (1-\delta)\cos\phi,
    \label{eq:rot_mat_rope_phi}
\end{align}
where $\delta$ controls the attenuation of weight towards the poles such that $w(\phi) \in [\delta,1]$.


\begin{table*}[!t]
    \centering
    \caption{\textbf{Zero-shot evaluation of universal domain robustness}. We evaluate all unified methods on a zero-shot setting across both indoor and outdoor domains. All models are trained on a mix of indoor and outdoor datasets. \uniKThreeD is trained on large FoV images, while the rest of the methods are only trained on perspective images. [Key: \textbf{Best}, \underline{Second Best}, S: Swin-L~\cite{liu2021swin}, V: ViT-L~\cite{Dosovitskiy2020VIT}]}
    \resizebox{0.95\linewidth}{!}{%
    \begin{tabular}{l|c|ccc|ccc|ccc}
    \toprule
    \multirow{2}*{\textbf{Methods}}  & \textbf{Dataset} & \multicolumn{3}{c|}{\textbf{\scannetpp}} & \multicolumn{3}{c|}{\textbf{\panoGVTwo}} & \multicolumn{3}{c}{\textbf{\kittiThreeSixty}}\\
     & \textbf{Size} & $\delta_1 \uparrow$ & \absRel $\downarrow$ & \rmse $\downarrow$ & $\delta_1 \uparrow$ & \absRel $\downarrow$ & \rmse $\downarrow$ & $\delta_1 \uparrow$ & \absRel $\downarrow$ & \rmse $\downarrow$\\
    \midrule
    \uniKThreeD\cite{piccinelli2025unik3d} - V & $7.94$M & \underline{$0.651$} & \underline{$0.253$} & \underline{$0.285$} & \boldmath$0.785$ & \underline{$0.170$} & \underline{$0.400$} & \underline{$0.817$} & \underline{$0.244$} & \boldmath$2.400$\\ 
    \midrule
    \metricThreeD v2~\cite{hu2024metric3d} - V & $16.20$M & $0.536$ & $0.223$ & $0.895$ & $0.404$ & $0.307$ & $0.855$ & $0.716$ & $0.200$ & $4.580$\\
    \unidepth~\cite{piccinelli2024unidepth} - V & $3.83$M & $0.364$ & $0.497$ & $1.166$ & $0.247$ & $0.789$ & $1.268$ & $0.481$ & $0.294$ & $6.564$\\
    
    \dacUni\cite{guo2025depth} - S & $0.79$M & $0.658$ & $0.233$ & $0.464$ & $0.684$ & $0.203$ & $0.507$ & $0.708$ & $0.186$ & $5.079$ \\
    \methodName - V & $1.45$M & \boldmath$0.918$ & \boldmath$0.097$ & \boldmath$0.277$ & \underline{$0.768$} & \boldmath$0.161$ & \boldmath$0.394$ & \boldmath$0.836$ & \boldmath$0.141$ & \underline{$3.977$}\\
    \end{tabular}
    }
    \label{tab:uni_exp}
\end{table*}

\begin{table*}[!t]
    \centering
    \caption{\textbf{Zero-shot evaluation of cross-camera generalization}. We evaluate all methods on a zero-shot setting across diverse cameras. *$_\text{I}$, *$_\text{O}$ indicate models trained only on the indoor and outdoor datasets, respectively. [Key: \textbf{Best}, \underline{Second Best}, R: ResNet101~\cite{he2016deep}, S: Swin-L~\cite{liu2021swin}, V: ViT-L~\cite{Dosovitskiy2020VIT}]}
    \resizebox{\textwidth}{!}{%
    \begin{tabular}{l|ccc|ccc|ccc|ccc}
    \toprule
    \multirow{2}*{\textbf{Methods}}  & \multicolumn{3}{c|}{\textbf{\scannetpp}} & \multicolumn{3}{c|}{\textbf{\panoGVTwo}} & \multicolumn{3}{c|}{\textbf{\mThreeD}} & \multicolumn{3}{c}{\textbf{\kittiThreeSixty}}\\
     & $\delta_1 \uparrow$ & \absRel $\downarrow$ & \rmse $\downarrow$ & $\delta_1 \uparrow$ & \absRel $\downarrow$ & \rmse $\downarrow$ & $\delta_1 \uparrow$ & \absRel $\downarrow$ & \rmse $\downarrow$ & $\delta_1 \uparrow$ & \absRel $\downarrow$ & \rmse $\downarrow$\\
    \midrule
    \dacIn~\cite{guo2025depth} - R & $0.852$ & $0.132$ & $0.309$ & $0.812$ & $0.139$ & $0.478$ & $0.773$ & $0.156$ & $0.619$ & $0.082$ & $0.464$ & $10.046$\\
    \dacIn~\cite{guo2025depth} - S & $0.854$ & $0.128$ & $0.287$ & $0.729$ & $0.184$ & $0.483$ & $0.723$ & $0.179$ & $0.591$ & $0.245$ & $0.365$ & $9.508$\\
    \dacOut~\cite{guo2025depth} - R & $0.256$ & $0.901$ & $2.312$ & $0.340$ & $0.616$ & $1.713$ & $0.390$ & $0.548$ & $1.630$ & $0.786$ & $0.156$ & $4.361$\\
    \dacOut~\cite{guo2025depth} - S & $0.109$ & $$1.412$$ & $3.539$ & $0.323$ & $0.870$ & $2.056$ & $0.330$ & $0.924$ & $2.164$ & $0.822$ & $0.149$ & $3.751$\\
    \midrule
    \dacUni\cite{guo2025depth} - S & \underline{$0.658$} & \underline{$0.233$} & \underline{$0.464$} & \underline{$0.684$} & \underline{$0.203$} & \underline{$0.507$} & \underline{$0.662$} & \underline{$0.215$} & \underline{$0.662$} & \underline{$0.708$} & \underline{$0.186$} & \underline{$5.079$} \\
    \methodName - V & \boldmath$0.918$ & \boldmath$0.097$ & \boldmath$0.277$ & \boldmath$0.768$ & \boldmath$0.161$ & \boldmath$0.394$ & \boldmath$0.745$ & \boldmath$0.175$ & \boldmath$0.442$ & \boldmath$0.836$ & \boldmath$0.141$ & \boldmath$3.977$\\
    \end{tabular}
    
    }
    \vspace{-1mm}
    \label{tab:full_exp}
\end{table*}

\subsection{Optimization}
We train \methodName by optimizing relative depth loss $\loss_\relSym$ and metric depth loss $\loss_\metricSym$ on the predicted relative depth map $\depthMap^\relSym$ and metric depth map $\depthMap^\metricSym$, respectively.
The losses $\loss_\relSym, \loss_\metricSym$ are both \siLog loss widely in MMDE.
Given predicted depth map $\depthMap$ and corresponding ground-truth depth map $\bar{\depthMap}$, we estimate the logarithmic error $\epsilon_\pixel = \ln{\bar{\depthMap}[\pixel]} - \ln{\depthMap[\pixel]}$.
The \siLog is then computed as follows:
\begin{align}
    \loss_{\siLog} 
= \sqrt{\frac{1}{n} \sum_{\pixel} (\epsilon_\pixel)^2 
- \frac{\lambda}{n^2} \left( \sum_{\pixel} (\epsilon_\pixel) \right)^2}.
\label{eq:silog1}
\end{align}
Using expectation $\mathbb{E}$ and variance $\mathbb{V}$, we rearrange \cref{eq:silog1} to get the following:
\begin{align}
    \loss_{\siLog} 
= \sqrt{\mathbb{V} [\epsilon_\pixel] +
(1 - \lambda) \mathbb{E}^2[\epsilon_\pixel]}.
\label{eq:silog2}
\end{align}
We can observe from \cref{eq:silog2} that by setting $\lambda = 1$ we get a purely scale-invariant loss.
The standard \siLog used in MMDE sets the $\lambda=0.85$ allowing a mixture of scale-invariant and scale-variant components.
We utilize this observation to define the losses $\loss_\relSym$ and $\loss_\metricSym$ as follows:
\begin{align}
    \loss_\relSym = \loss_{\siLog}^{\lambda=1} \;;\; \loss_\metricSym = \loss_{\siLog}^{\lambda=0.85}.
\end{align}

%% file: sec/5_experiments.tex

\section{Experiments}
\label{sec:exp}

\subsection{Setup}
\Paragraph{Training Datasets}
Our training set comprises three indoor and four outdoor datasets captured from perspective cameras.
The indoor datasets consist of \hmThreeD~\cite{ramakrishnan2021habitat}, \hypersim~\cite{roberts:2021}, and \taskonomy~\cite{zamir2018taskonomy}, while the outdoor datasets consists of \ddad~\cite{guizilini20203d}, \lyft~\cite{houston2021one}, \argoverseTwo~\cite{wilson2023argoverse}, and \aTwoDTwo~\cite{geyer2020a2d2}.
We use the tiny version of \hmThreeD and \taskonomy provided by OmniData~\cite{eftekhar2021omnidata} to streamline training.
We exclude images of rear-center camera in \aTwoDTwo and front-camera in \argoverseTwo as explained in \cref{sec:abl_train_data}.
Altogether, the training set comprises 1.1M perspective images; 670K from indoor datasets and 780K from outdoor datasets.

\Paragraph{Testing Datasets}
We evaluate the cross-camera generalization of \methodName on two fisheye datasets, namely, \scannetpp~\cite{yeshwanth2023scannet++} and \kittiThreeSixty~\cite{liao2022kitti}, and two 360$^\circ$ datasets, namely, \panoGVTwo~\cite{albanis2021pano3d} and \mThreeD~\cite{chang2017matterport3d}.

\Paragraph{Baselines}
We compare \methodName to the following works:
\begin{itemize}
    \item \uniKThreeD~\cite{piccinelli2025unik3d}: A SoTA method in zero-shot MMDE that does not require any camera parameters in inference.
    \item \dac~\cite{guo2025depth}: A SoTA method for cross-camera generalization that is trained only on perspective images. \dacIn, \dacOut denote models trained on indoor and outdoor datasets, respectively.  \dacUni denotes a model trained from scratch on combined indoor and outdoor datasets. Since it was not released by \cite{guo2025depth}, we train it using the same setup as our method for a fair comparison.
\end{itemize}
We additionally include \unidepth~\cite{piccinelli2024unidepth} and \metricThreeD v2~\cite{hu2024metric3d} evaluated in \dac \cite{guo2025depth} for comparison. 

\Paragraph{Evaluation Metrics}
We employ standard metrics  in MMDE, {\it i.e.}, percentage of inliers ($\di{i} \uparrow$) with a threshold of $1.25^i$, absolute mean relative error (\absRel$\downarrow$), and root mean squared error (\rmse$\downarrow$).

\Paragraph{Implementation}
We implement \methodName in PyTorch~\cite{paszke2019pytorch} and CUDA~\cite{nickolls2008scalable}.
We initialize ViT-L~\cite{Dosovitskiy2020VIT} backbone with DINO~\cite{simeoni2025dinov3} pre-trained weights as our encoder and use DPT~\cite{ranftl2021vision} as our decoder.
We utilize the AdamW optimizer initialized with hyperparameters $\beta_1=0.9$, $\beta_2=0.999$, and an initial learning rate of 0.0001.
Cosine Annealing, proposed by~\cite{loshchilov2016sgdr}, is employed as the learning rate scheduler with the final learning rate set to one-tenth of the initial learning rate.
We train the model for 120k iterations with a batch size of 128.
We follow the training setup of \cite{guo2025depth} and apply FoV-alignment, multi-resolution sampling, and ERP augmentations.
All ablations are performed by training on \hmThreeD and \kittiThreeSixty datasets using the ViT-B backbone.


\subsection{Comparison with SoTA}
\label{sec:sota_comp}
\cref{tab:uni_exp} compares \methodName with \cite{piccinelli2024unidepth,hu2024metric3d,guo2025depth,piccinelli2025unik3d} in universal domain robustness on large FoV datasets.
We observe that \methodName outperforms all the prior methods trained with perspective images on both indoor and outdoor datasets and sets the SoTA in cross-camera generalization.
\uniKThreeD's \cite{piccinelli2025unik3d} training set contains images from large FoV cameras, and thus does not constitute a fair comparison to the other methods trained only on perspective images.
However, \methodName still outperforms \uniKThreeD \cite{piccinelli2025unik3d}, demonstrating the strong generalization prowess of \methodName.
We omit the evaluation on \mThreeD~\cite{chang2017matterport3d} from \cref{tab:uni_exp} since it is present in the training set of \cite{piccinelli2025unik3d} and would be an unfair comparison to other methods.

We observe from \cref{tab:uni_exp} that \methodName achieves ${\sim}26\%$ improvement in $\dOne$ metric in the \scannetpp dataset compared to both \uniKThreeD and \dacUni.
While we improve the \panoGVTwo by ${\sim}8.4\%$ over \dacUni, we achieve similar performance as that of \uniKThreeD.
As mentioned previously, \uniKThreeD contains \mThreeD, a $360^\circ$ dataset similar to \panoGVTwo, and we still come close to \uniKThreeD on the \panoGVTwo dataset, demonstrating the robustness of our method.
We outperform prior methods on \kittiThreeSixty, including \uniKThreeD by ${\sim}2\%$, even though \cite{piccinelli2025unik3d} contains \aiMotive \cite{matuszka2023aimotive}, an outdoor fisheye dataset in its training set.

\subsubsection{Comparison with \dac}
\cref{tab:full_exp} compares \methodName against unified and domain specific \dac models across all testing datasets.
As expected, \dacIn~\cite{guo2025depth} and \dacOut~\cite{guo2025depth} performance drops catastrophically when evaluated using outdoor and indoor datasets, respectively.
Although \dacUni~\cite{guo2025depth} is trained on both indoor and outdoor data, it fails to generalize well since it tries to learn a single global scale for both domains.
\methodName outperforms \dacUni, consistently and significantly across indoor and outdoor domains, underscoring the effect of decoupling metric depth into relative and scale estimation.

In addition to outperforming \dacUni, \cref{tab:full_exp} shows that \methodName outperforms \dacIn on \scannetpp.
Although \methodName beats \dacIn with the Swin backbone on \panoGVTwo and \mThreeD, we fall short of beating \dacIn with the ResNet backbone.
We believe this is because transformers struggle with scale equivariance as noted in \cite{guo2025depth}.
While comparing between transformer backbones, \thatIs, Swin and ViT, we outperform \dacIn on \panoGVTwo and \mThreeD with ${\sim}4\%$ and ${\sim}2.2\%$ respectively.

\begin{figure*}
    \centering
    \begin{minipage}[t]{\linewidth}
        \centering
        \begin{tikzpicture}
            \draw (0,0) node[inner sep=0] {\includegraphics[width=0.23\linewidth]{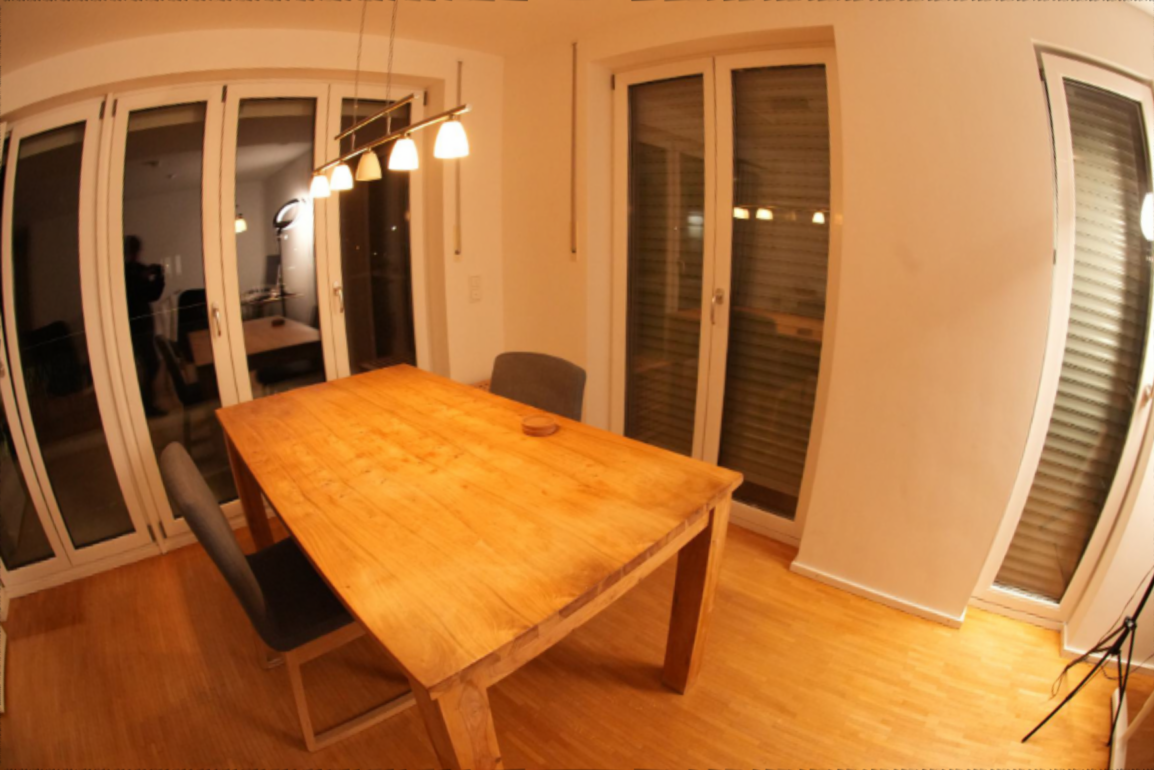}};
            \draw (-1.96,1.1) node[inner sep=1, anchor=west, fill=white, fill opacity=0.8, text opacity=1] {\fontsize{8.0}{10}\selectfont {\scannetpp~\cite{yeshwanth2023scannet++}}};
        \end{tikzpicture}
        \includegraphics[width=0.23\linewidth]{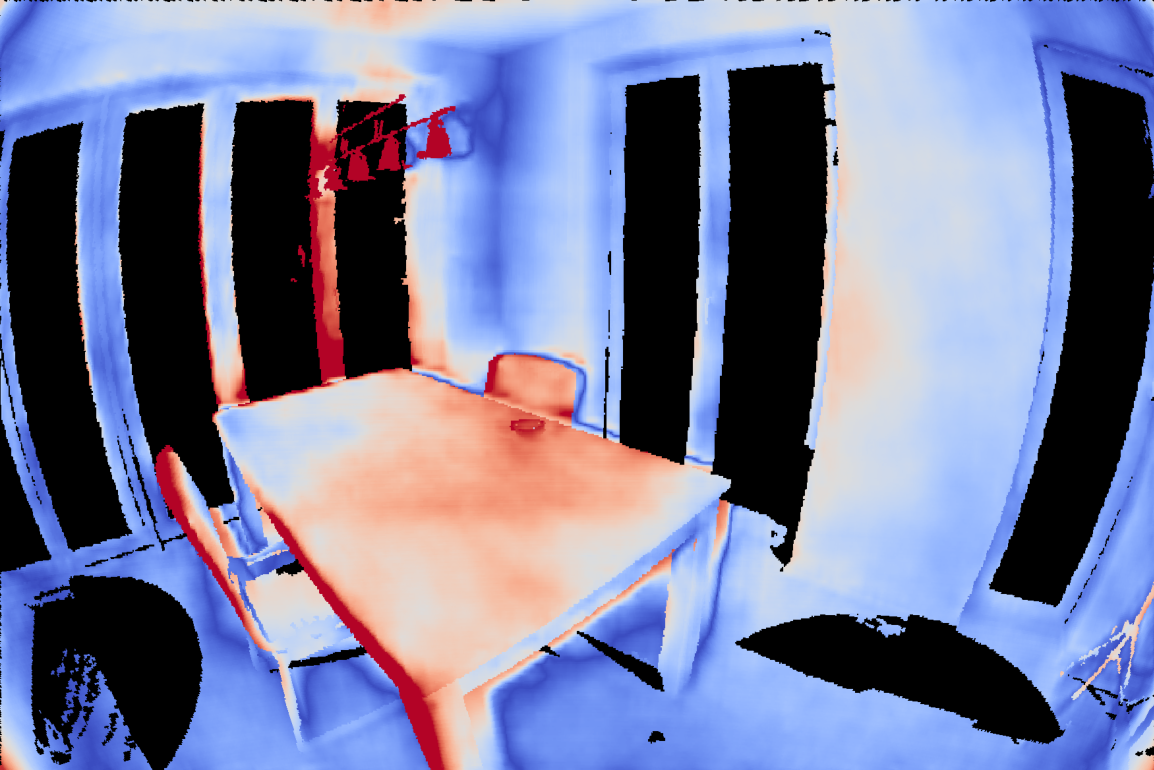}
        \includegraphics[width=0.23\linewidth]{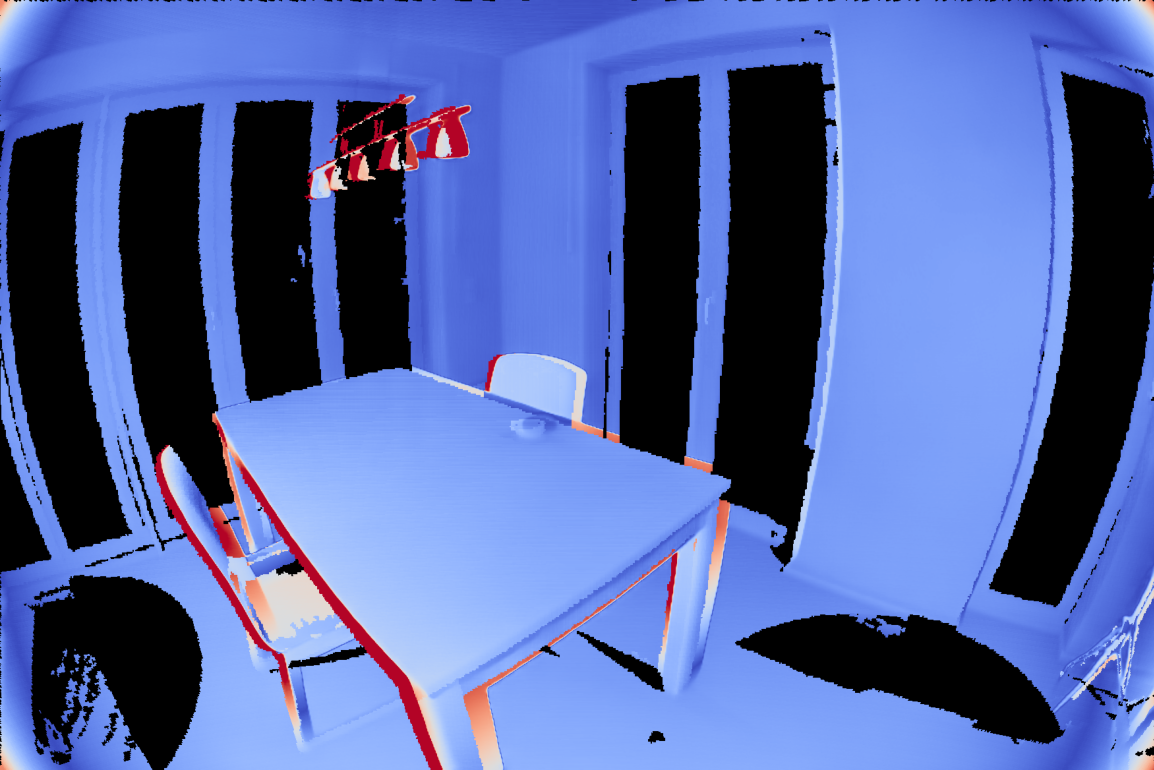}
        \includegraphics[width=0.23\linewidth]{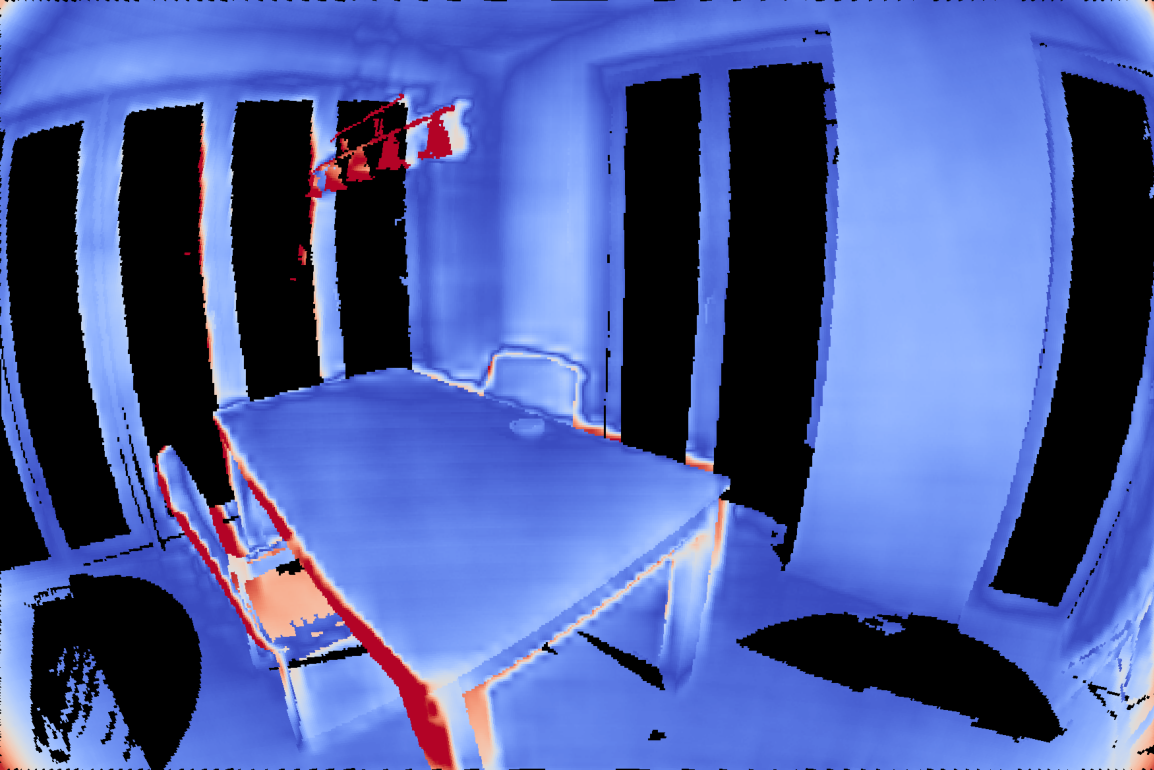}
        \raisebox{0.7ex}{\includegraphics[width=0.05\linewidth]{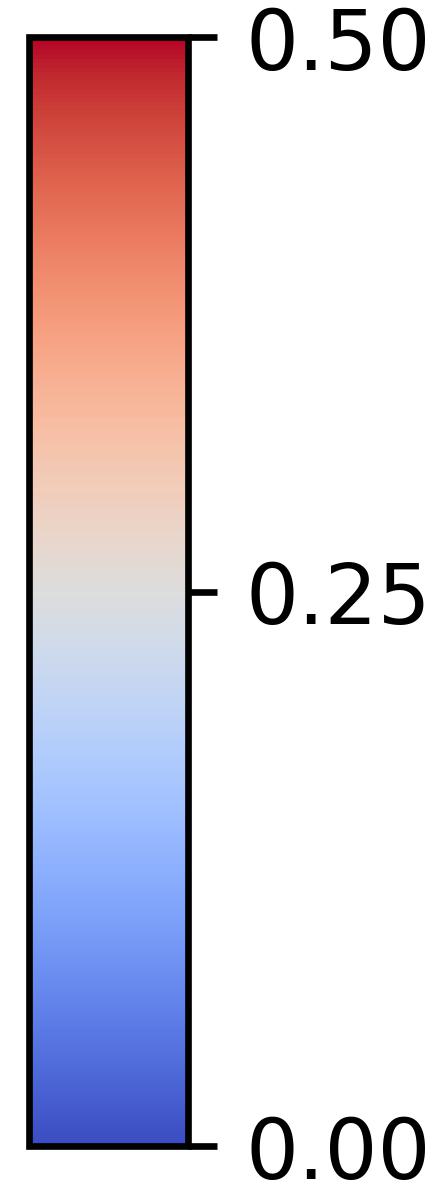}}
    \end{minipage}
    \begin{minipage}[t]{\linewidth}
        \centering
        \includegraphics[width=0.23\linewidth]{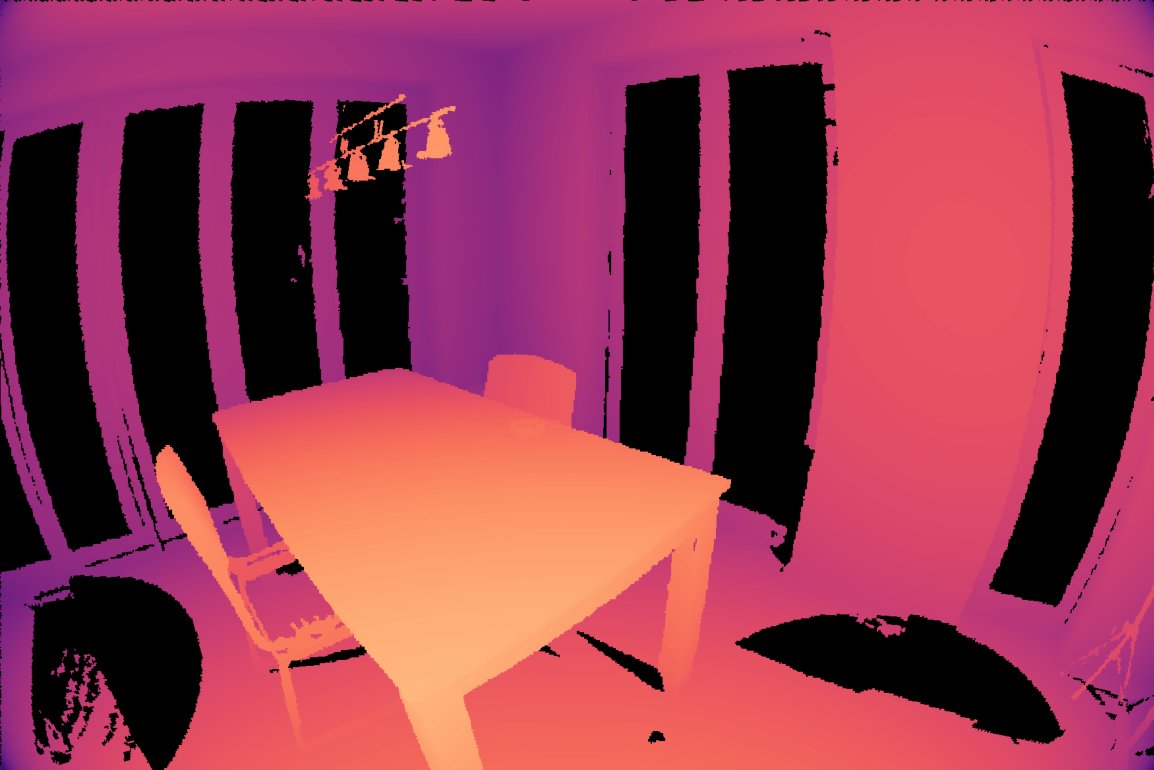}
        \includegraphics[width=0.23\linewidth]{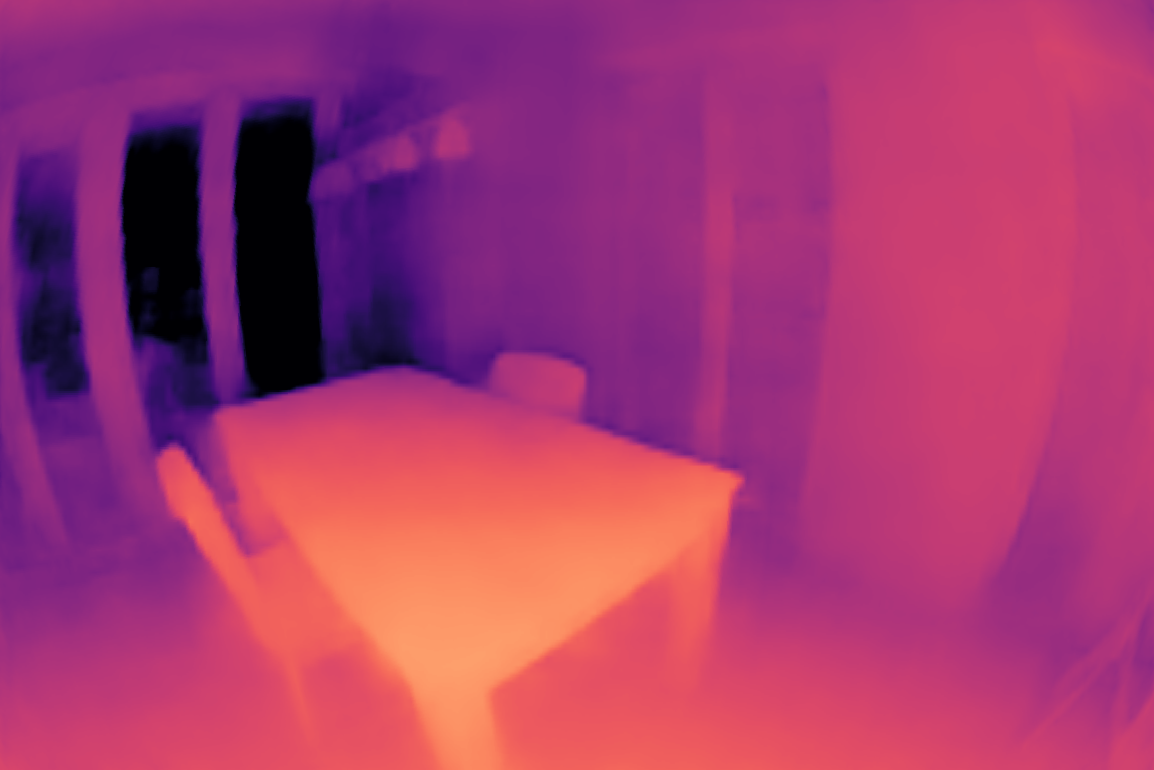}
        \includegraphics[width=0.23\linewidth]{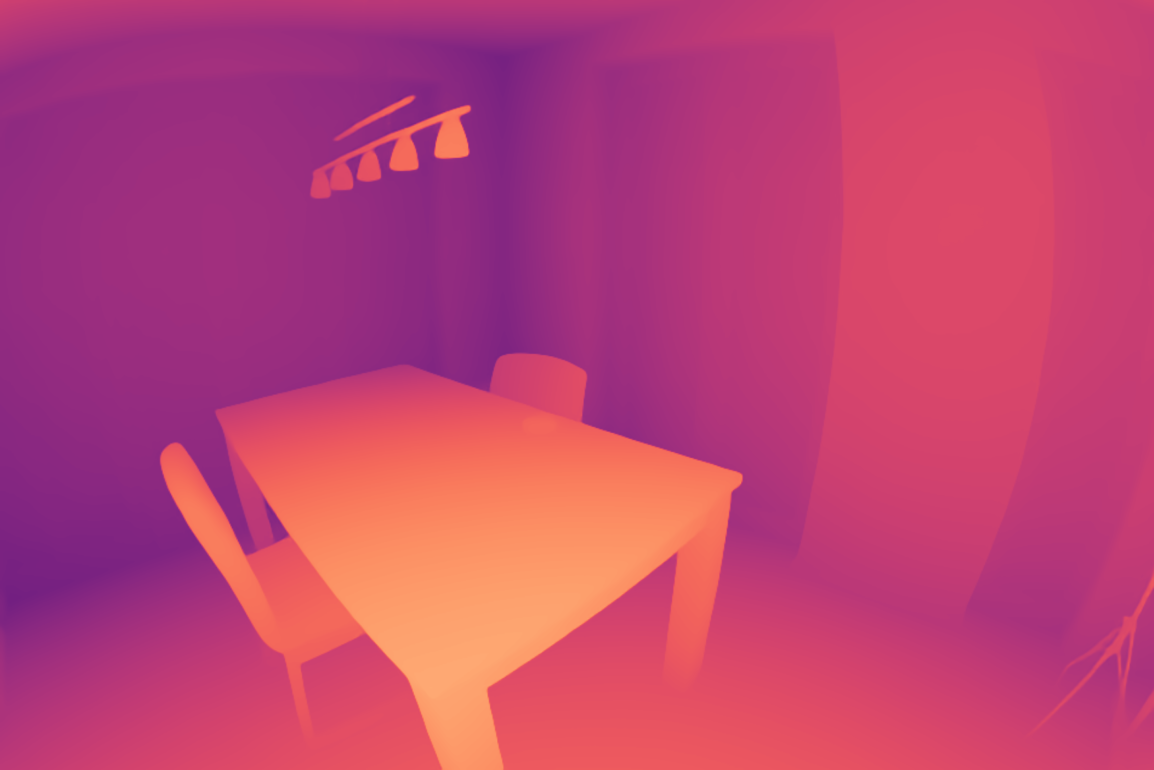}
        \includegraphics[width=0.23\linewidth]{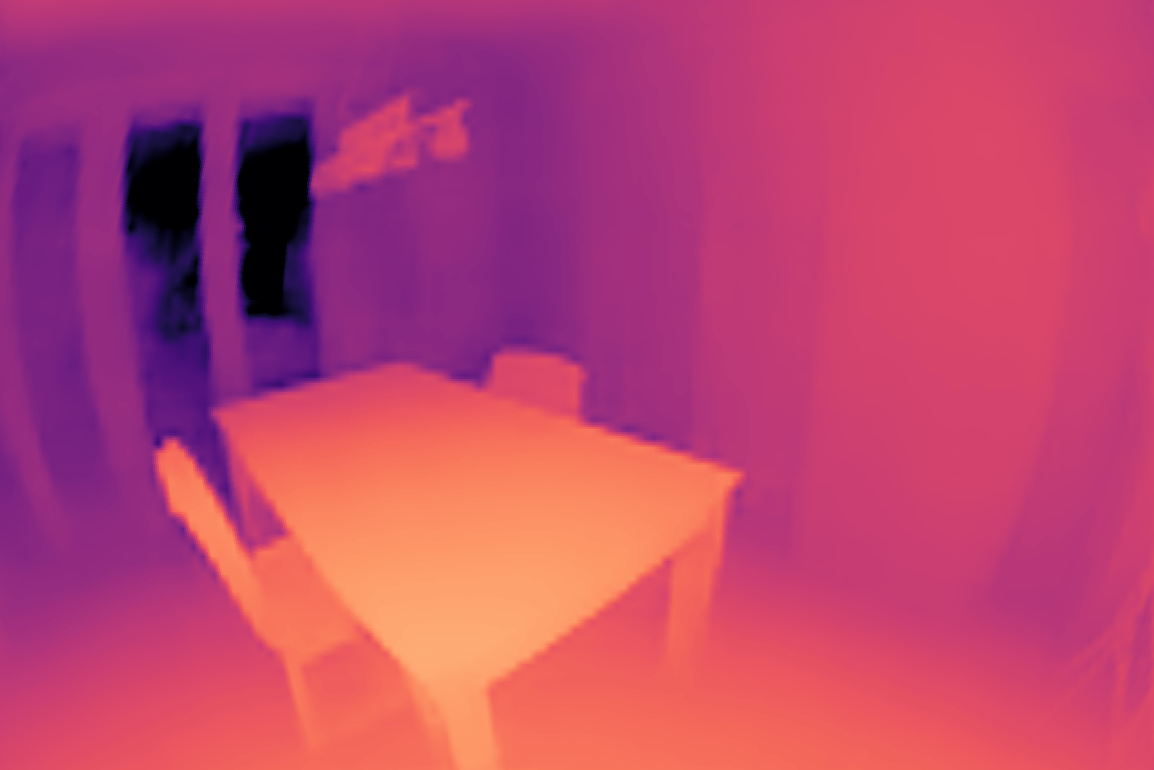}
        \raisebox{0.5ex}{\includegraphics[width=0.05\linewidth]{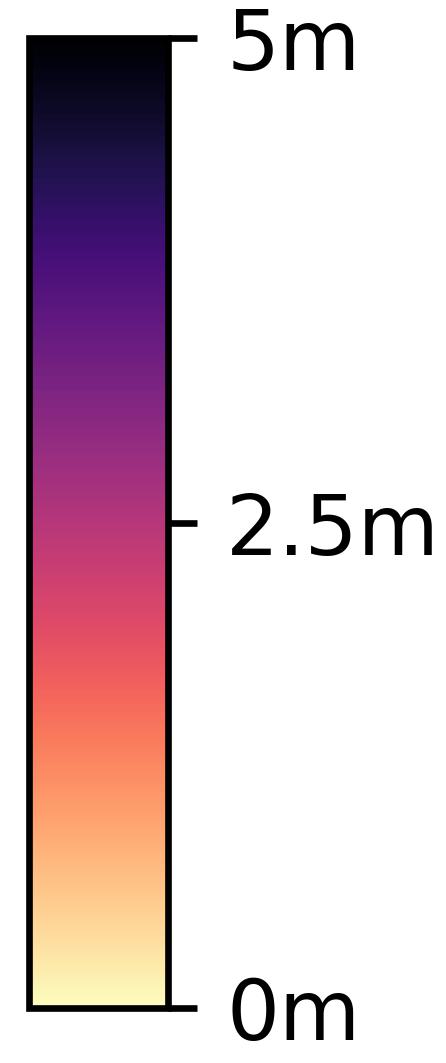}}
    \end{minipage}
    \begin{minipage}[t]{\linewidth}
        \centering
        \begin{tikzpicture}
            \draw (0,0) node[inner sep=0] {\includegraphics[width=0.23\linewidth]{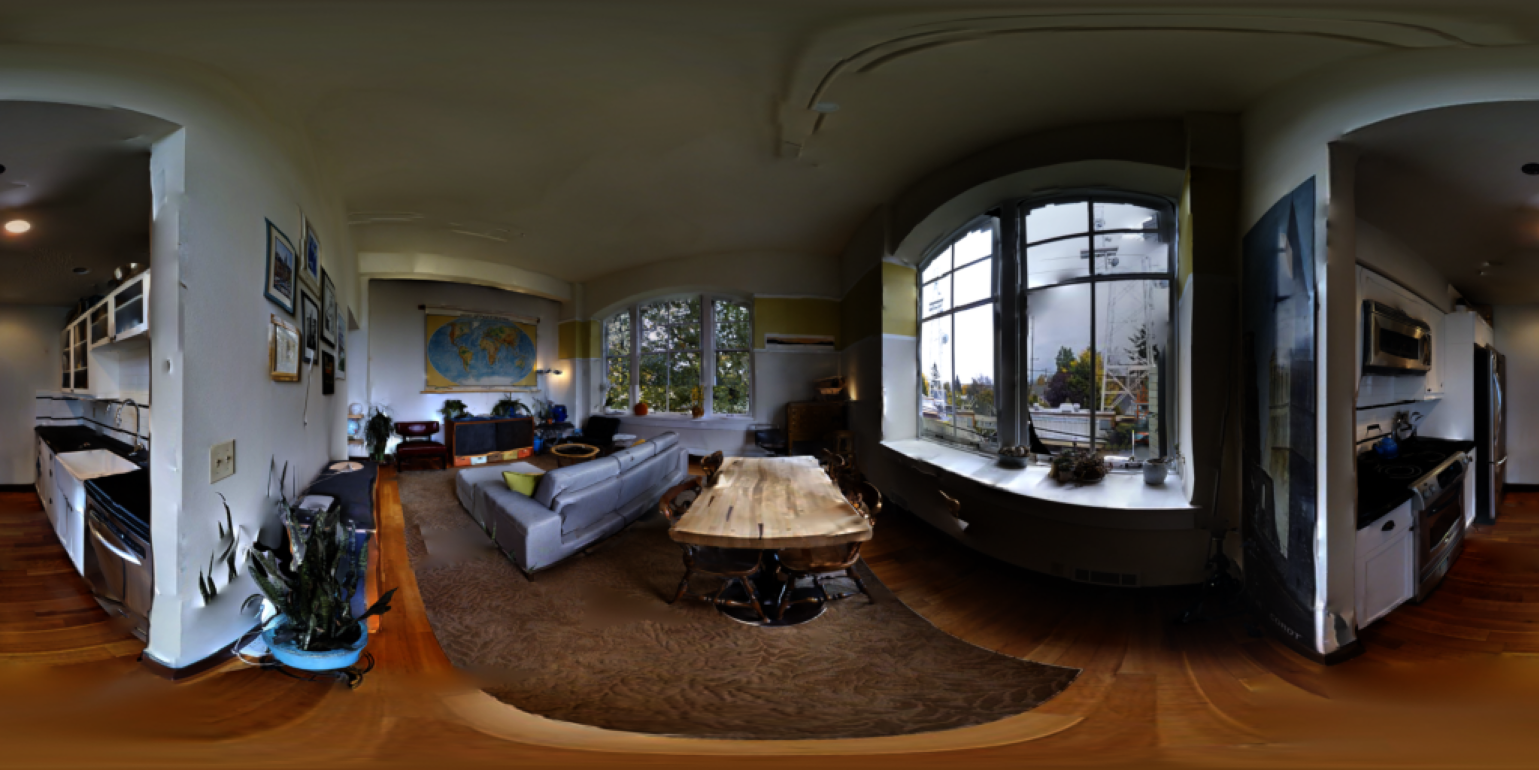}};
            \draw (-1.96,0.8) node[inner sep=1, anchor=west, fill=white, fill opacity=0.8, text opacity=1] {\fontsize{8.0}{10}\selectfont {\panoGVTwo~\cite{albanis2021pano3d}}};
        \end{tikzpicture}
        \includegraphics[width=0.23\linewidth]{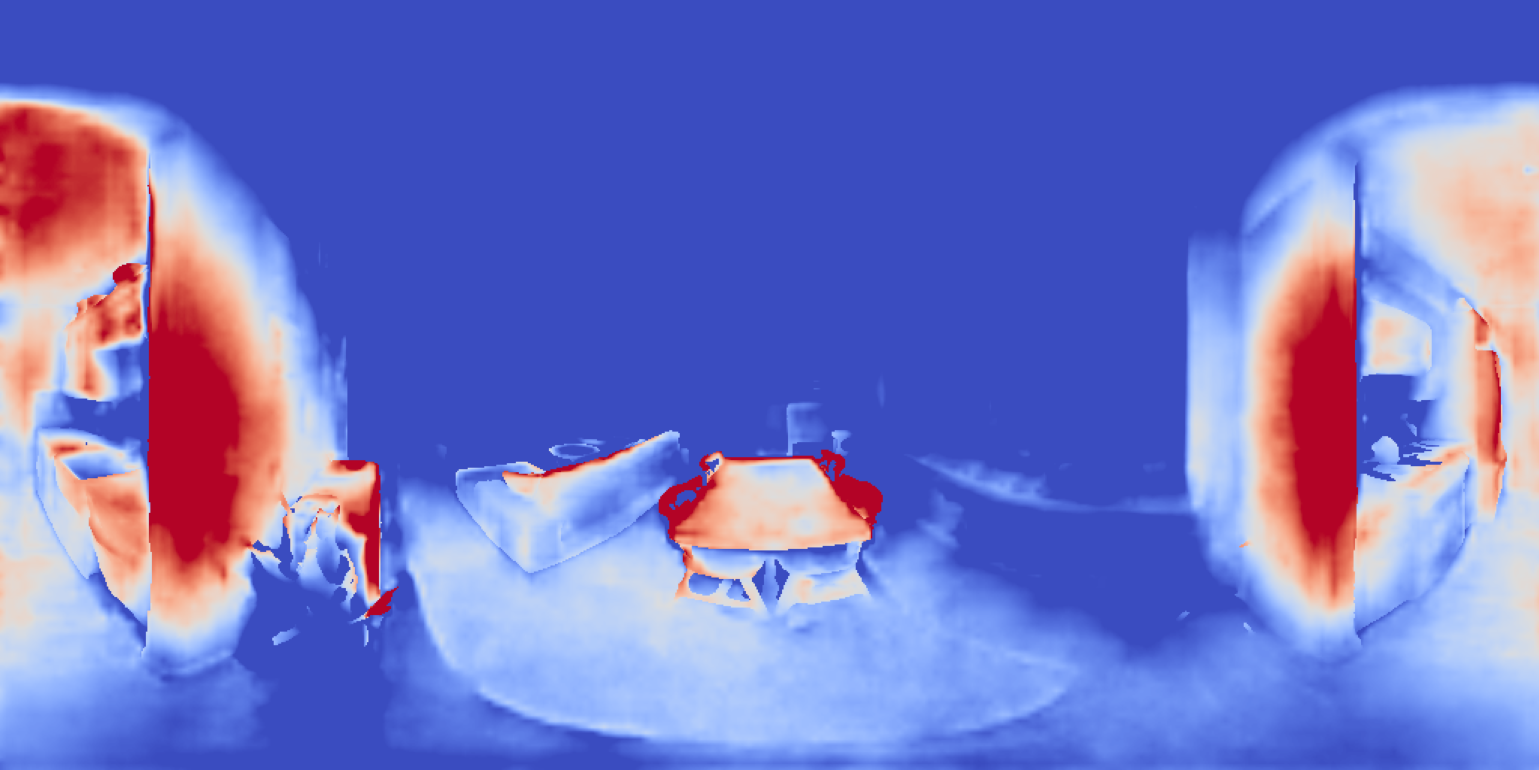}
        \includegraphics[width=0.23\linewidth]{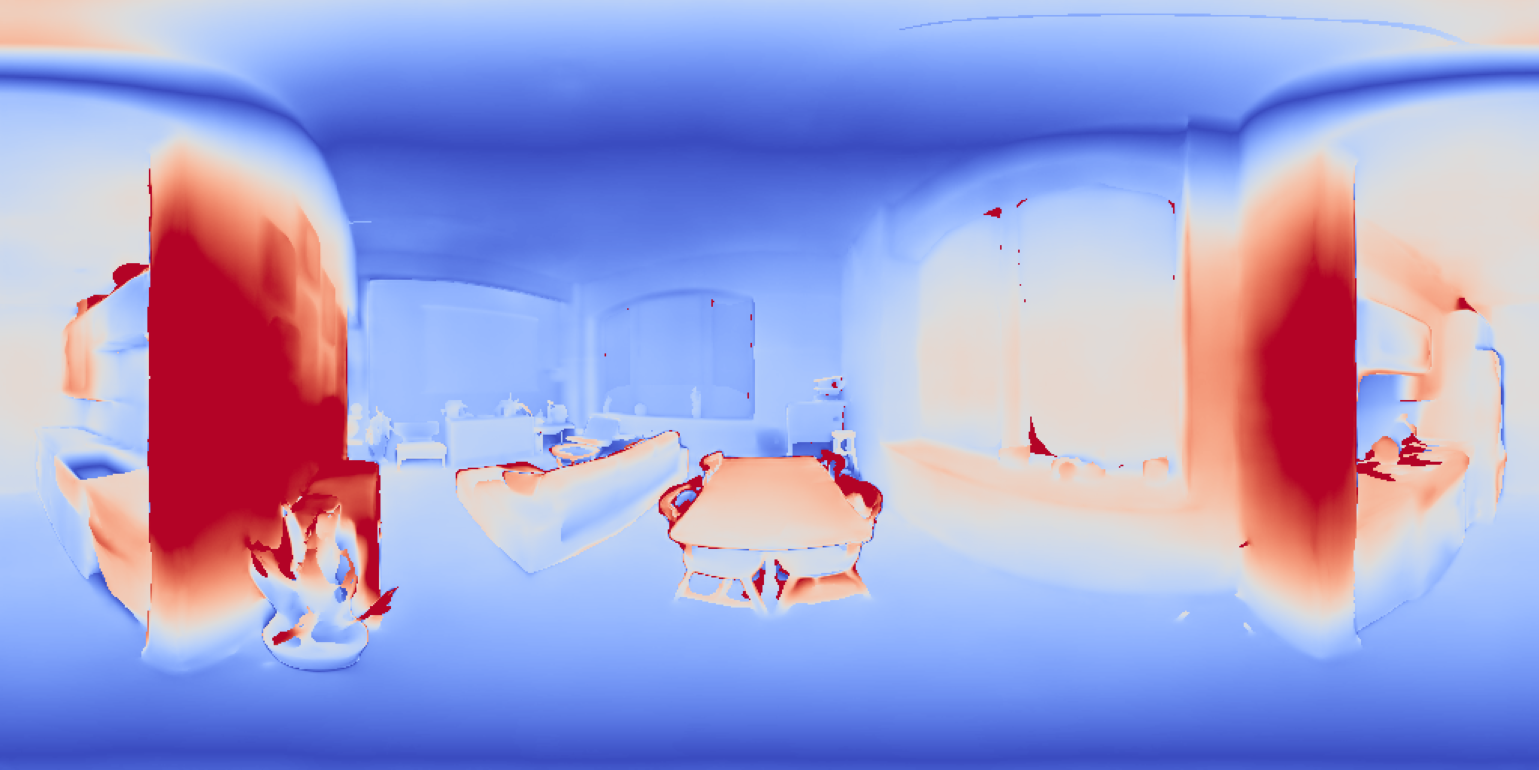}
        \includegraphics[width=0.23\linewidth]{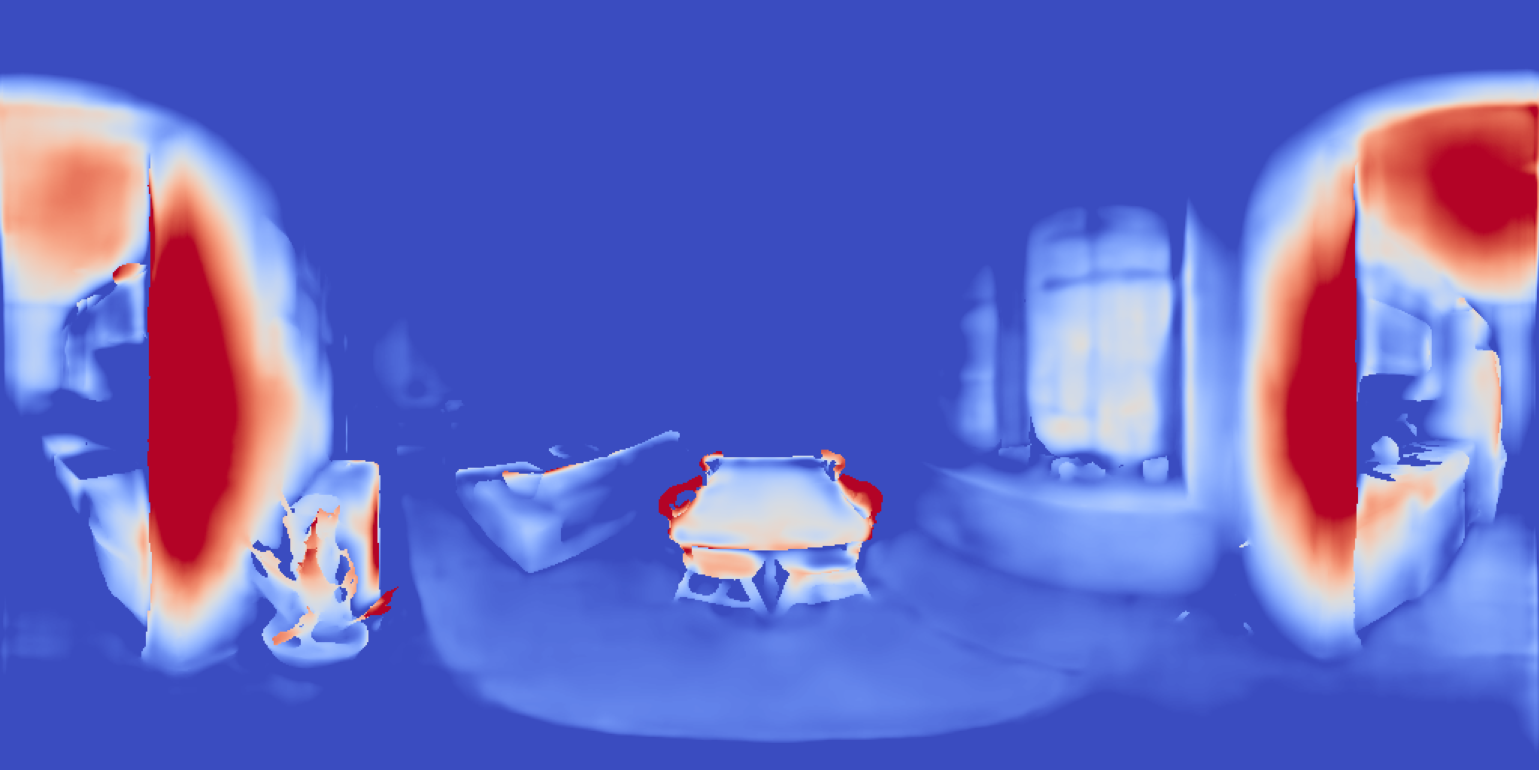}
        \raisebox{2ex}{\includegraphics[width=0.05\linewidth]{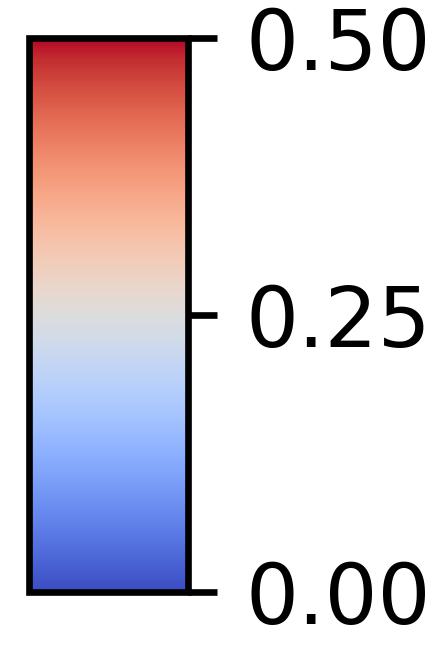}}
    \end{minipage}
    \begin{minipage}[t]{\linewidth}
        \centering
        \includegraphics[width=0.23\linewidth]{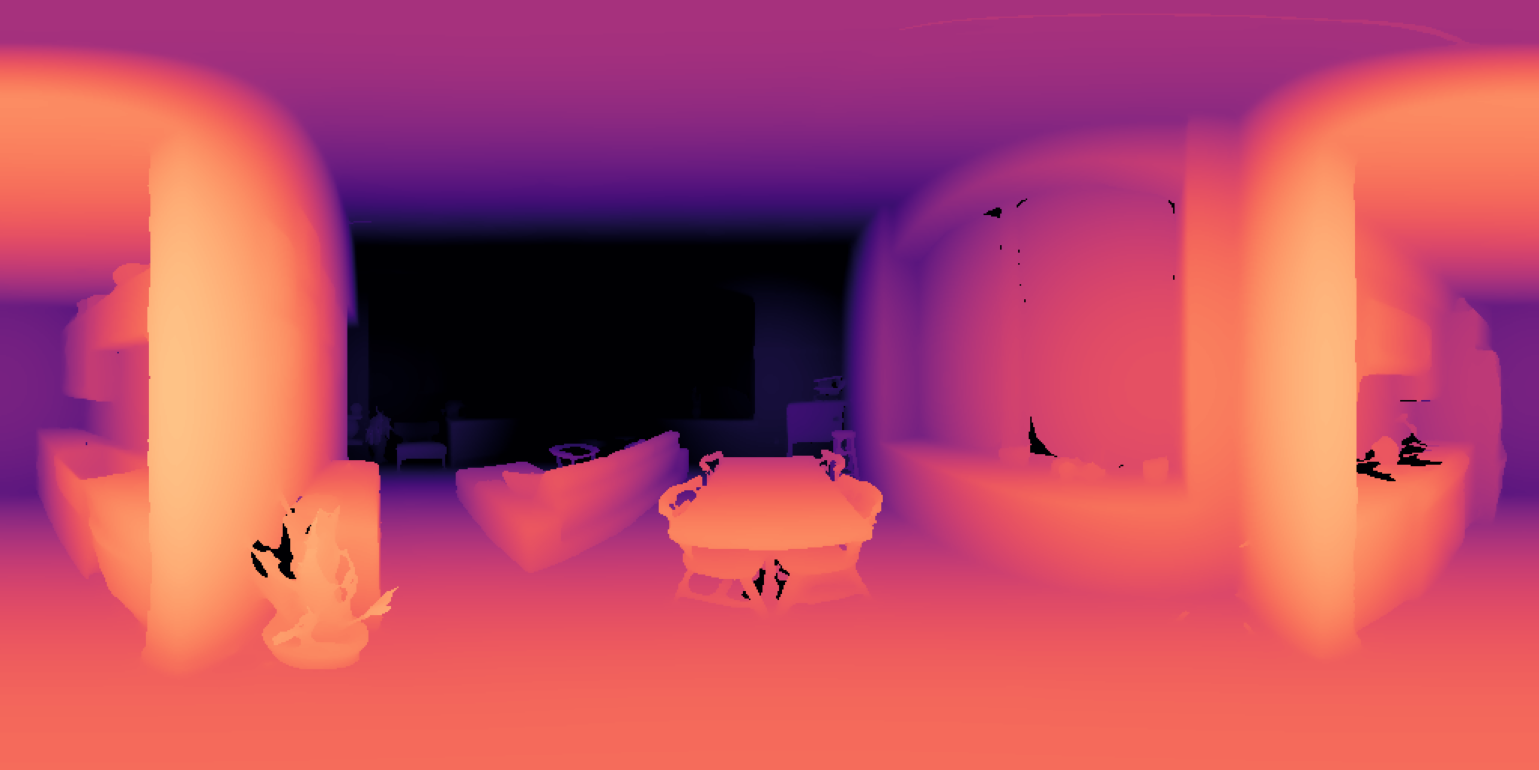}
        \includegraphics[width=0.23\linewidth]{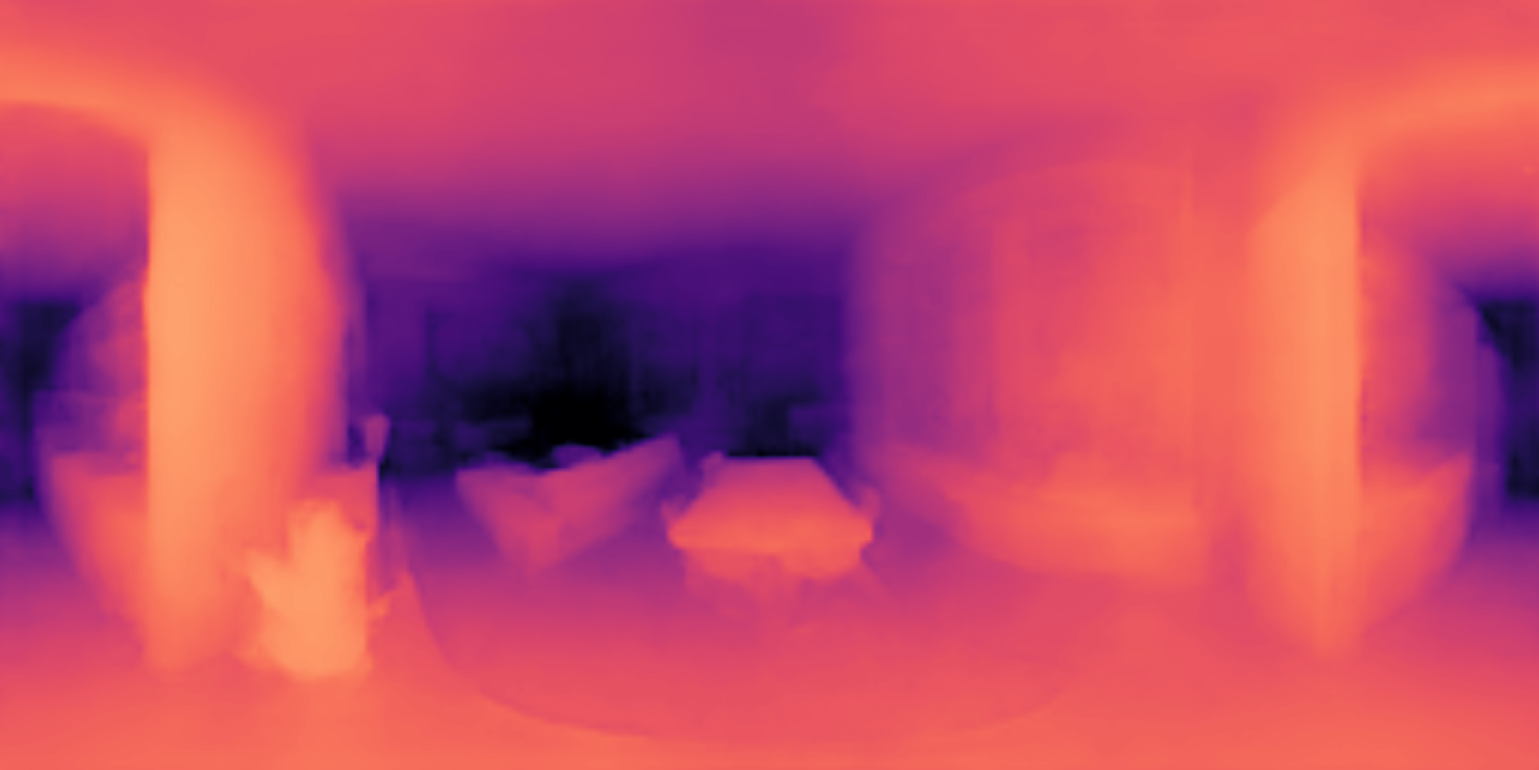}
        \includegraphics[width=0.23\linewidth]{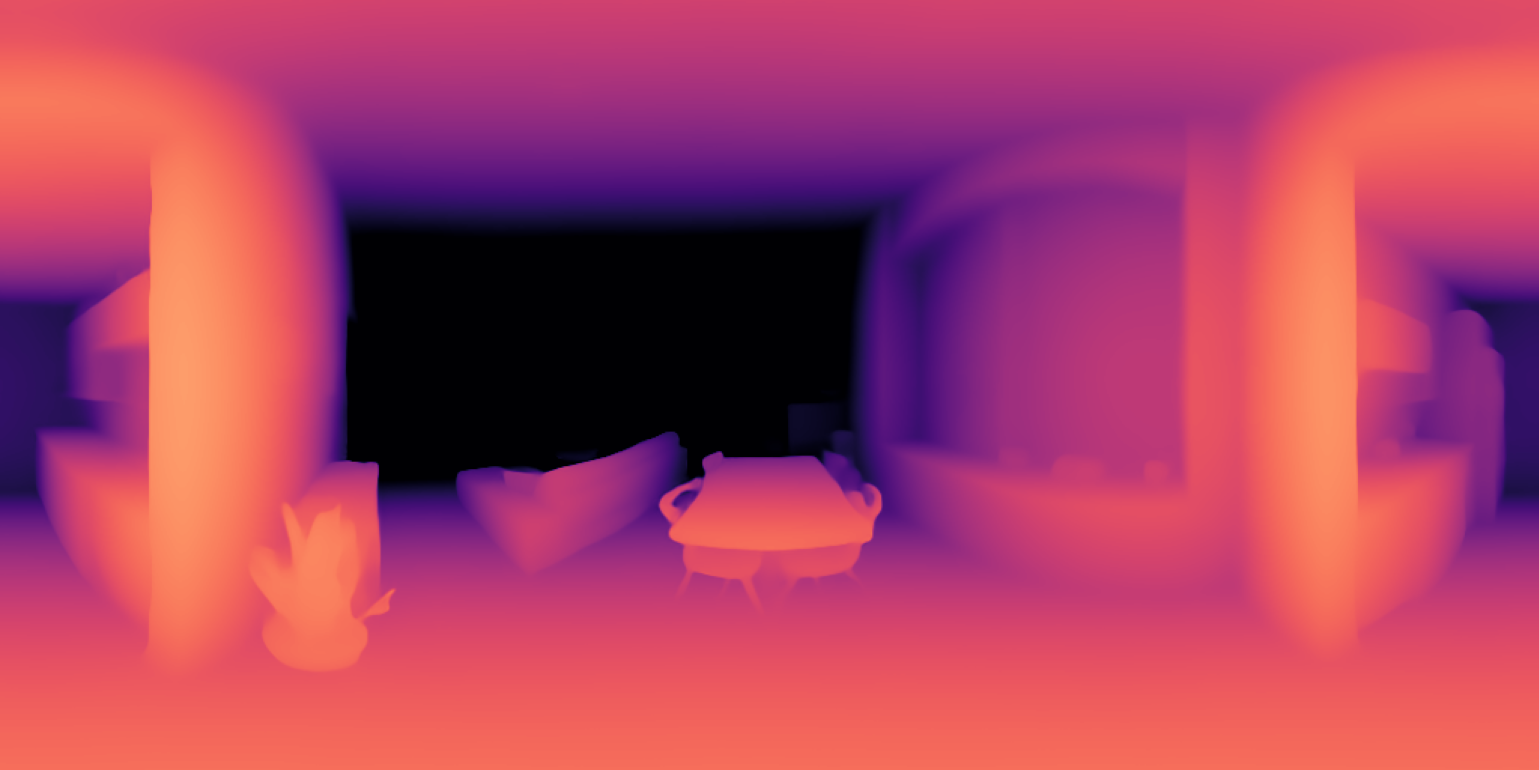}
        \includegraphics[width=0.23\linewidth]{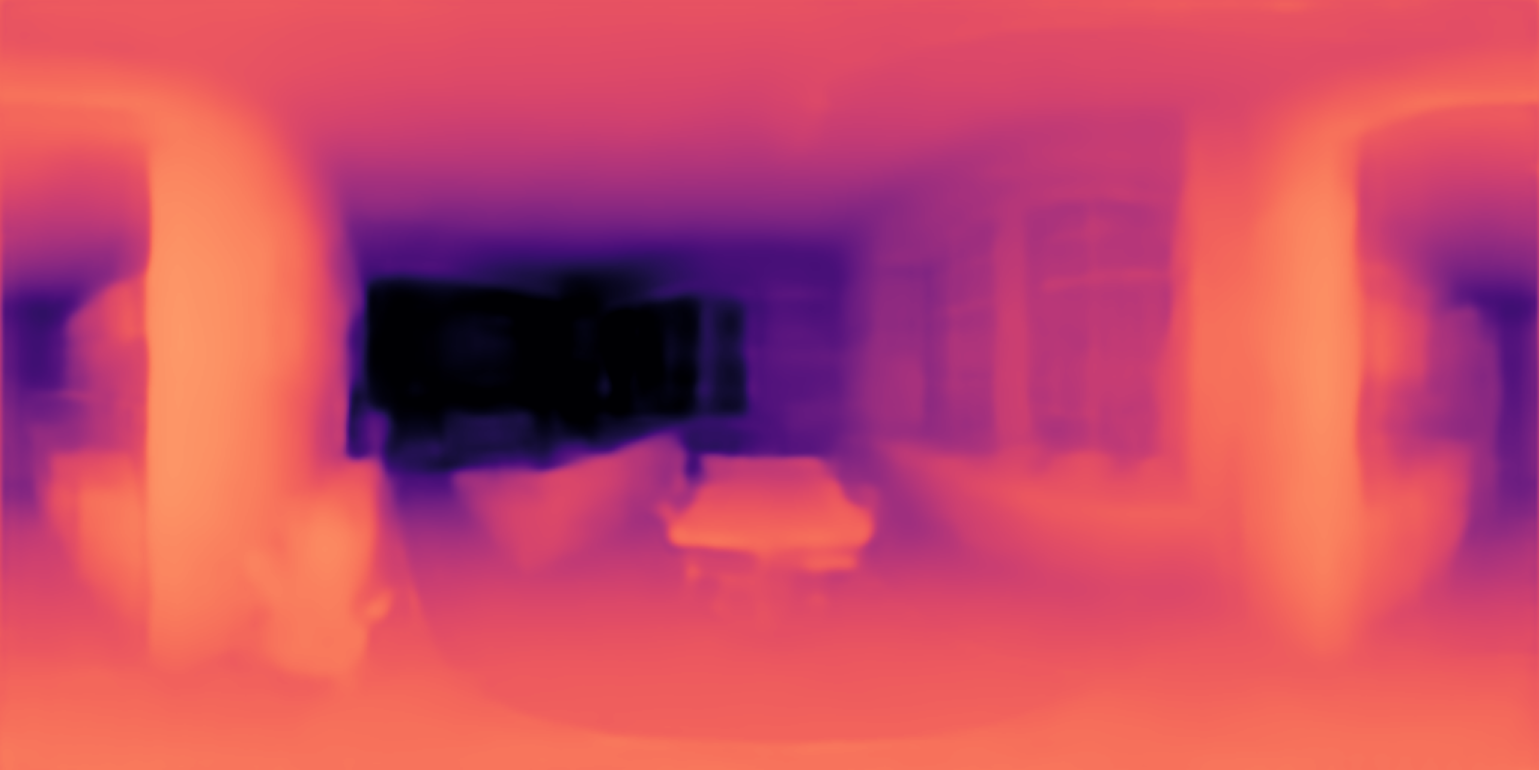}
        \raisebox{2ex}{\includegraphics[width=0.05\linewidth]{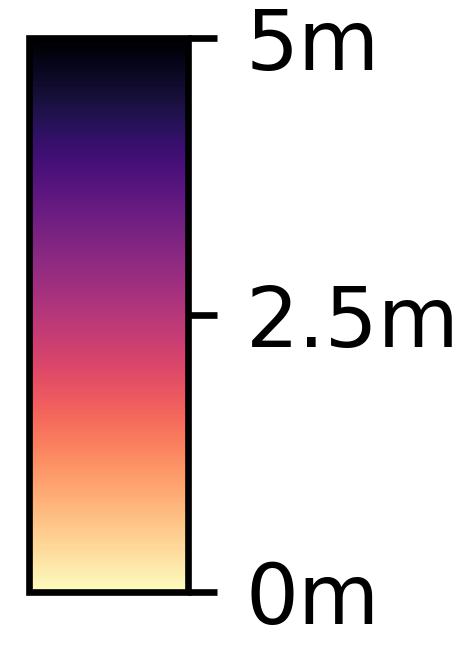}}
    \end{minipage}
    \begin{minipage}[t]{\linewidth}
        \centering
        \begin{tikzpicture}
            \draw (0,0) node[inner sep=0] {\includegraphics[width=0.23\linewidth]{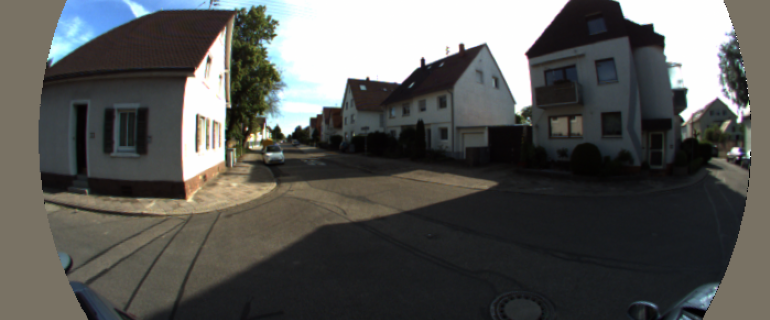}};
            \draw (-1.96,0.65) node[inner sep=1, anchor=west, fill=white, fill opacity=0.8, text opacity=1] {\fontsize{8.0}{10}\selectfont {\kittiThreeSixty~\cite{liao2022kitti}}};
        \end{tikzpicture}
        \includegraphics[width=0.23\linewidth]{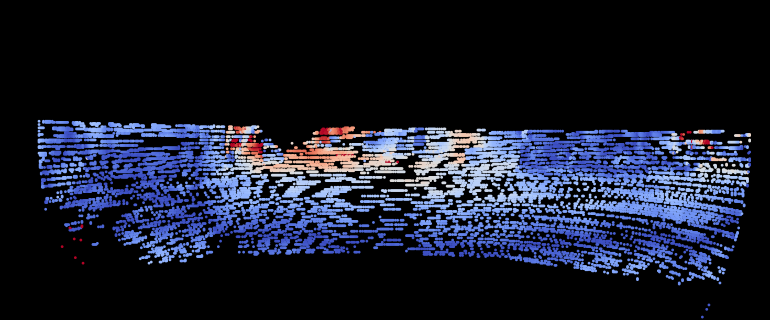}
        \includegraphics[width=0.23\linewidth]{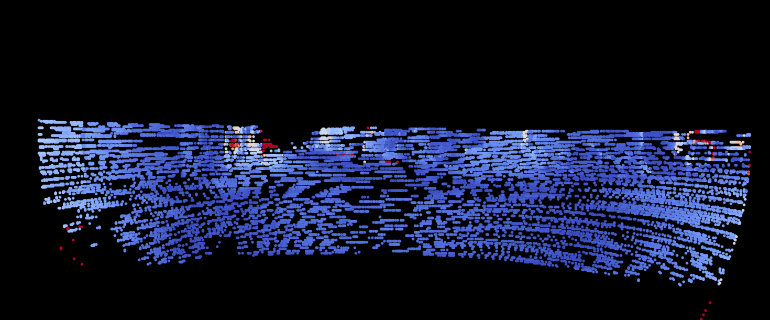}
        \includegraphics[width=0.23\linewidth]{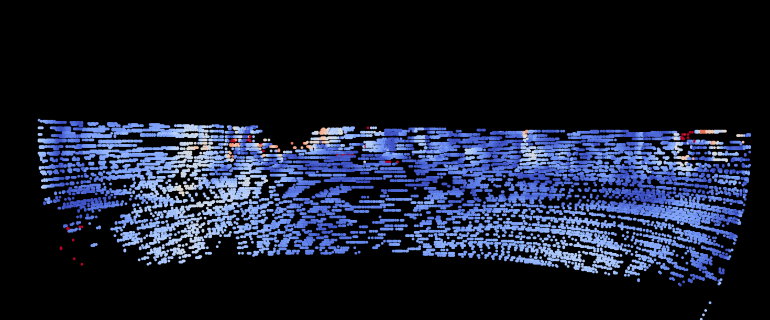}
        \raisebox{1.2ex}{\includegraphics[width=0.05\linewidth]{figures/qualitative/dac/kitti360/arel.jpg}}
    \end{minipage}
    \begin{minipage}[t]{\linewidth}
        \centering
        \begin{tikzpicture}
            \draw (0,0) node[inner sep=0] {\includegraphics[width=0.23\linewidth]{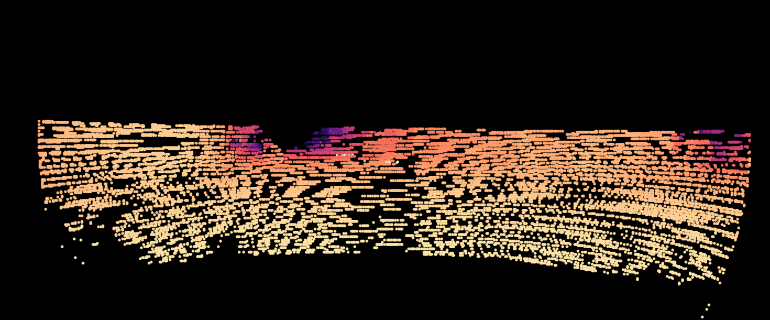}};
            \draw (0,-1.1) node[inner sep=0, align=center] {\fontsize{8.0}{10}\selectfont {RGB \& GT}};
        \end{tikzpicture}
        \begin{tikzpicture}
            \draw (0,0) node[inner sep=0] {\includegraphics[width=0.23\linewidth]{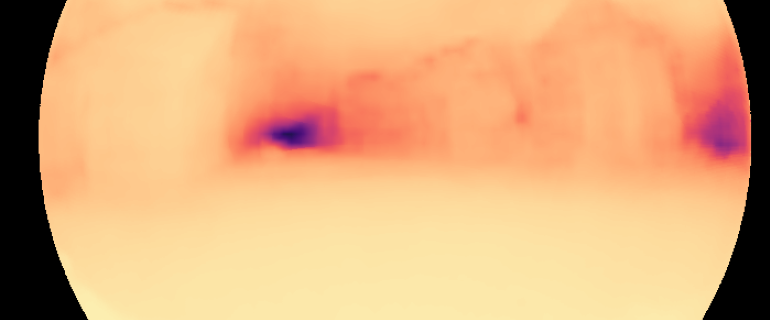}};
            \draw (0,-1.08) node[inner sep=0, align=center] {\fontsize{8.0}{10}\selectfont {\dacUni~\cite{guo2025depth}}};
        \end{tikzpicture}
        \begin{tikzpicture}
            \draw (0,0) node[inner sep=0] {\includegraphics[width=0.23\linewidth]{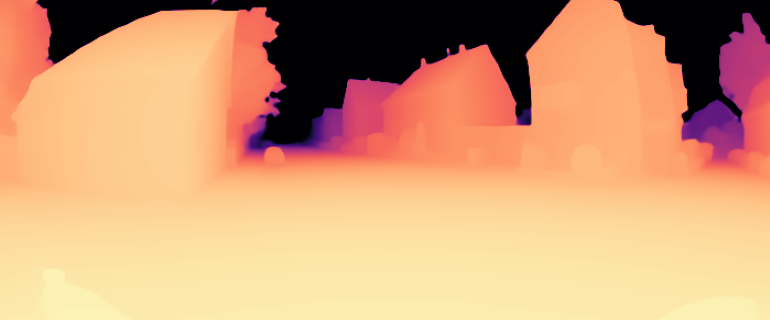}};
            \draw (0,-1.08) node[inner sep=0, align=center] {\fontsize{8.0}{10}\selectfont {\uniKThreeD~\cite{piccinelli2025unik3d}}};
        \end{tikzpicture}
        \begin{tikzpicture}
            \draw (0,0) node[inner sep=0, align=center] {\includegraphics[width=0.23\linewidth]{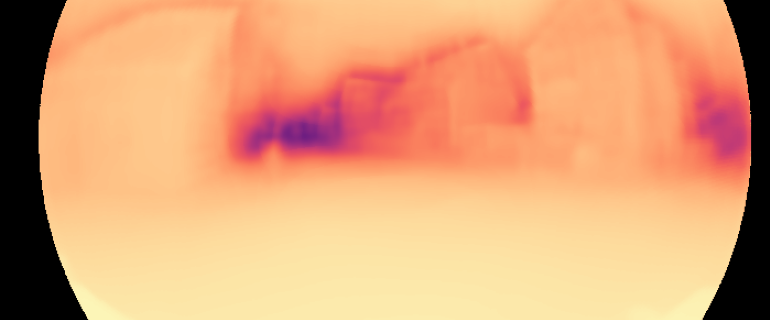}};
            \draw (0.1,-1.08) node[inner sep=0, align=center] {\fontsize{8.0}{10}\selectfont {\methodName\textcolor{white}{[]}}};
        \end{tikzpicture}
        \raisebox{3.5ex}{\includegraphics[width=0.05\linewidth]{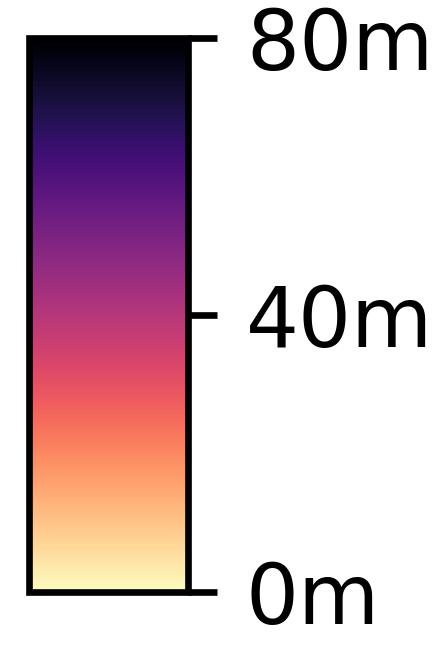}}
    \end{minipage}
    
    \caption{\textbf{Qualitative Results.} Every pair of consecutive rows corresponds to a single sample. Odd rows display the input RGB image, and \absRel error between predicted and GT depth maps. Even rows display the GT depth map and predicted depth maps. \methodName reduces error in the distorted region as seen around the table in \scannetpp compared to \dacUni and on the walls in \panoGVTwo compared to \uniKThreeD.\vspace{-1mm}}
    \label{fig:qual_res}
\end{figure*}
\begin{table}[!h]
    \centering
    \caption{\textbf{Ablation on Scale Estimation.} We study the impact of depth-guided scale estimation by comparing it against single-scale estimation. The model performances are compared in a zero-shot setting. `-' indicates we directly estimate metric depth without decoupling relative depth and metric scale. [Key: \textbf{Best}]}
    \resizebox{\linewidth}{!}{%
    \begin{tabular}{l|ccc|ccc}
    \toprule
    \multirow{2}*{\textbf{Scale}} & \multicolumn{3}{c|}{\textbf{\scannetpp}} & \multicolumn{3}{c}{\textbf{\kittiThreeSixty}} \\
     & $\delta_1 \uparrow$ & \absRel $\downarrow$ & \rmse $\downarrow$ & $\delta_1 \uparrow$ & \absRel $\downarrow$ & \rmse $\downarrow$\\
     \midrule
    $-$ & $0.782$ & $0.152$ & $0.401$ & $0.563$ & $0.286$ & $5.461$\\
    $\scale \in \realDomain$ & $0.773$ & $0.166$ & $0.426$ & $0.601$ & $0.245$ & $5.257$\\
    $\scaleMap \in \realDomain^{H\times W}$ & \boldmath$0.792$ & \boldmath$0.140$ & \boldmath$0.396$ & \boldmath$0.622$ & \boldmath$0.239$ & \boldmath$5.057$\\
    \end{tabular}
    }
    \label{tab:abl_scale}
\end{table}
\subsection{Ablation Studies}
\Paragraph{Impact of Scale Map}
We analyze the benefits of scale map estimation by comparing with single-scale estimation and direct metric-depth estimation without decoupling.
For direct metric-depth estimation, we remove the scale estimation module and use  the decoder's output as the metric depth.
We keep the relative depth estimation architecture the same, but replace the Depth-Guided Scale estimation module with a single scalar estimation.
We follow \cite{wang2025moge} and utilize the \texttt{[CLS]} token from the global features $\feature_g$ of the encoder.
We estimate the  scale as $\scale = \texttt{MLP}(\feature_g\texttt{[CLS]})\in \realDomain$, where $\texttt{MLP}$ is a shallow fully-connected network.

\cref{tab:abl_scale} shows that directly predicting metric depth performs well on \scannetpp but fails on \kittiThreeSixty.
This is due to strong indoor bias in the ablation data: \hmThreeD (310K) vs.\ \ddad (80K).
Predicting a scale map $\scaleMap$ gives $2\%$ performance gain over single-scale and thus validates our approach.

\Paragraph{Impact of RoPE-\texorpdfstring{$\phi$}{phi}}
We evaluate the impact of our proposed distortion-aware positional embedding RoPE-$\phi$ by comparing it with 2D RoPE defined in \cref{sec:vanilla_rope}.
We can observe from \cref{eq:rot_mat_rope_phi} that simply setting $\delta=1$ would give us 2D RoPE, thus indicating that 2D RoPE is a special case of RoPE-$\phi$.
We observe from \cref{tab:abl_rope} that \scannetpp benefits from RoPE-$\phi$ more than \kittiThreeSixty.
We believe that this is because \kittiThreeSixty contains valid depth along a narrow band around close to the equator as opposed to \scannetpp, which provides a dense-depth map spanning a large FoV.

\begin{table}[!h]
    \centering
    \caption{\textbf{Ablation on RoPE.} We evaluate the benefit of RoPE-$\phi$, comparing it against a model trained with 2D RoPE. The model performances are compared in a zero-shot setting. [Key: \textbf{Best}]}
    \resizebox{\linewidth}{!}{%
    \begin{tabular}{l|ccc|ccc}
    \toprule
    \multirow{2}*{\textbf{Pos. Emb}} & \multicolumn{3}{c|}{\textbf{\scannetpp}} & \multicolumn{3}{c}{\textbf{\kittiThreeSixty}} \\
     & $\delta_1 \uparrow$ & \absRel $\downarrow$ & \rmse $\downarrow$ & $\delta_1 \uparrow$ & \absRel $\downarrow$ & \rmse $\downarrow$\\
    \midrule
    RoPE & $0.750$ & $0.177$ & $0.452$ & $0.592$ & $0.254$ & $5.313$\\
    RoPE-$\phi$ & \boldmath$0.792$ & \boldmath$0.140$ & \boldmath$0.396$ & \boldmath$0.622$ & \boldmath$0.239$ & \boldmath$5.057$\\
    \end{tabular}
    }
    \label{tab:abl_rope}
\end{table}

%% file: sec/6_conclusion.tex
\section{Conclusion}
\label{sec:concl}

We presented \methodName, a unified framework for monocular metric depth estimation that universally generalizes across diverse camera models and scene domains using a single model. 
By decoupling metric depth into relative depth and scale map, we enable robust estimation across indoor and outdoor domains with a single model. 
Our proposed Depth-Guided Scale Estimation effectively fuses global and local information to produce scale-aware predictions with minimal overhead. 
Furthermore, we introduced RoPE-$\phi$, a distortion-aware positional encoding tailored for ERP images, enhancing the model’s performance on large FoV images. 
Extensive experiments demonstrate that \methodName achieves state-of-the-art performance in cross-camera and cross-domain generalization, establishing a strong foundation for scalable and universal depth perception systems.

%% file: sec/X_suppl.tex
\clearpage
\setcounter{page}{1}
\maketitlesupplementary











\section{Data}
\subsection{Training Data}
\label{sec:abl_train_data}
\cref{tab:train_data} provides an overview of the training datasets. 
In addition to the training datasets utilized in \dac~\cite{guo2025depth}, we add \argoverseTwo and \aTwoDTwo to balance the indoor and outdoor distribution in the training set.

We observe that out of seven cameras in \argoverseTwo, the front camera's aspect ratio is different than the rest of the six cameras.
Specifically, the resolution (H$\times$W) is $1550\times2048$ while the rest of the cameras have a resolution of $2048\times1550$.
Since the aspect ratio of the front camera is less than one, we omit images from the front camera for our training.
Furthermore, \argoverseTwo contains $1.5$M samples, and to prevent introducing bias to outdoor data, we randomly sample $300$K image-depth pairs to complete our training set.

\aTwoDTwo consists of six cameras, namely, front-center, front-right, front-left, side-right, side-left, and rear-center, with corresponding \lidar acquisitions.
We exclude the images from the rear-center cameras and add $350$K images obtained from the remaining five cameras to the training set.


We note that \uniKThreeD's~\cite{piccinelli2025unik3d} training set consists of more than $8$M samples, whereas our training set consists only $1.45$M samples.
Moreover, out of the $8$M images in the training set of \uniKThreeD, $72.4\%$ are perspective images, $27.27\%$ are fisheye images, and $0.33\%$ are ERP images.
In contrast, $100\%$ of \methodName's training set consists of perspective images as seen in \cref{tab:train_data}
\begin{table}[!h]
    \centering
    \caption{\textbf{Training Datasets.} List of training datasets with the following attributes: number of images, scene type, and acquisition method. [Key: Syn=Synthetic, Rec=Mesh reconstruction]}
    \resizebox{\linewidth}{!}{%
    \begin{tabular}{l|c|c|c}
    \toprule
    \textbf{Dataset} & \textbf{\#Images} & \textbf{Scene} & \textbf{Acquisition}\\
    \midrule
    \hmThreeD-tiny~\cite{ramakrishnan2021habitat} & $310$K & Indoor & Rec\\
    \taskonomy-tiny~\cite{zamir2018taskonomy} & $300$K & Indoor & RGB-D\\
    \hypersim~\cite{roberts:2021} & $54$K & Indoor & Syn\\
    \ddad~\cite{guizilini20203d} & $80$K & Outdoor & LiDAR\\
    \lyft~\cite{houston2021one}& $50$K & Outdoor & LiDAR\\
    \argoverseTwo~\cite{wilson2023argoverse} & $300$K & Outdoor & LiDAR\\
    \aTwoDTwo~\cite{geyer2020a2d2} & $350$K & Outdoor & LiDAR\\
    \bottomrule
    \end{tabular}}
    \label{tab:train_data}
\end{table}
\begin{table}[!h]
    \centering
    \caption{\textbf{Testing Datasets.} List of testing datasets with the following attributes: camera type, scene type, and acquisition method. [Key: Rec=Mesh reconstruction]}
    \resizebox{\linewidth}{!}{%
    \begin{tabular}{l|c|c|c}
    \toprule
    \textbf{Dataset} & \textbf{Cam.Type} & \textbf{Scene} & \textbf{Acquisition}\\
    \midrule
    \scannetpp~\cite{yeshwanth2023scannet++} & Fisheye & Indoor & Rec\\
    \mThreeD~\cite{chang2017matterport3d} & ERP & Indoor & Rec\\
    \panoGVTwo~\cite{albanis2021pano3d} & ERP & Indoor & Rec\\
    \kittiThreeSixty~\cite{liao2022kitti} & Fisheye & Outdoor & LiDAR\\
    \bottomrule
    \end{tabular}}
    \label{tab:test_data}
\end{table}

\subsection{Testing Data}
\cref{tab:test_data} details the testing data used to evaluate \methodName.
We follow the testing data setup of \dac and evaluate all our baselines on them.
While \scannetpp and \kittiThreeSixty are both fisheye datasets, they differ in their respective distortion models.
\scannetpp follows the KB~\cite{kannala2006generic} model while \kittiThreeSixty follows the MEI~\cite{mei2007single} model.
We adopt the lookup table provided by \dac for fast fisheye to ERP conversion.
\panoGVTwo and \mThreeD datasets consist of $360^\circ$ images, which are provided in the ERP images by default, and therefore we use them as is.

\begin{table}[!h]
    \centering
    \vspace{-3mm}
    \caption{\textbf{Effect of camera intrinsics on depth performance.} We study the impact of utilizing predicted and ground-truth \{P,G\} camera intrinsics during inference. [Key: \textbf{Best}, \underline{Second Best}]}
    \vspace{-3mm}
    \resizebox{\linewidth}{!}{%
    \begin{tabular}{l|ccc|ccc}
    \toprule
    \multirow{2}*{\textbf{Method}} & \multicolumn{3}{c|}{\textbf{\scannetpp}} & \multicolumn{3}{c}{\textbf{\kittiThreeSixty}} \\
     & $\delta_1 \uparrow$ & \absRel $\downarrow$ & \rmse $\downarrow$ & $\delta_1 \uparrow$ & \absRel $\downarrow$ & \rmse $\downarrow$\\
     \midrule
    \uniKThreeD & $0.651$ & $0.253$ & \boldmath$0.285$ & \underline{$0.817$} & $0.244$ & \boldmath$2.400$\\
    \midrule
    \methodName - P & $0.894$ & $0.110$ & $0.274$ & $0.706$ & $0.198$ & $4.397$\\
    \quad+\textit{\aTwoDTwo} & \underline{$0.917$} & \underline{$0.104$} & $0.279$ & $0.815$ & \underline{$0.154$} & $4.091$\\
    \midrule
    \methodName - G & $0.905$ & \underline{$0.104$} & $0.274$ & $0.757$ & $0.169$ & $4.470$\\
    \quad+\textit{\aTwoDTwo} & \boldmath$0.918$ & \boldmath$0.097$ & \underline{$0.277$} & \boldmath$0.836$ & \boldmath$0.141$ & \underline{$3.977$}\\
    \end{tabular}
    }
    \vspace{-3mm}
    \label{tab:abl_intr_pred}
\end{table}

\begin{table*}[!h]
    \centering
    \caption{\textbf{Zero-shot evaluation on perspective datasets.} We evaluate all unified models on perspective datasets. All models are trained on a mix of indoor and outdoor datasets. [Key: \textbf{Best}]}
    \resizebox{\linewidth}{!}{%
    \begin{tabular}{l|c|ccc|ccc|ccc|ccc}
    \toprule
    \multirow{2}*{\textbf{Method}} & \textbf{Training} & \multicolumn{3}{c|}{\textbf{\kitti}} & \multicolumn{3}{c|}{\textbf{\nyuDepth}} & \multicolumn{3}{c|}{\textbf{\nuscenes}} & \multicolumn{3}{c}{\textbf{\ibims}} \\
     & \textbf{Size} & $\delta_1 \uparrow$ & \absRel $\downarrow$ & \rmse $\downarrow$ & $\delta_1 \uparrow$ & \absRel $\downarrow$ & \rmse $\downarrow$ & $\delta_1 \uparrow$ & \absRel $\downarrow$ & \rmse $\downarrow$ & $\delta_1 \uparrow$ & \absRel $\downarrow$ & \rmse $\downarrow$\\
    \midrule
    \metricThreeD v2 & $16$M & \boldmath$0.974$ & \boldmath$0.053$ & \boldmath$2.493$ & $0.972$ & $0.067$ & $0.262$ & $0.841$ & $0.236$ & \underline{$9.400$} & $0.684$ & $0.207$ & $0.700$\\
    \unidepth & $3$M  & $0.964$ & $0.116$ & $2.788$ & \boldmath$0.988$ & \boldmath$0.052$ & \boldmath$0.194$ & \boldmath$0.846$ & \boldmath$0.127$ & \boldmath$4.560$ & $0.157$ & $0.410$ & $1.250$\\
    \midrule
    \uniKThreeD & $8$M & $0.833$ & $0.159$ & $4.323$ & $0.899$ & $0.133$ & $0.400$ & \underline{$0.840$} & \underline{$0.189$} & $10.830$ & \boldmath$0.919$ & \boldmath$0.104$ & \boldmath$0.406$\\
    \dacUni & $0.8$M & $0.767$ & $0.180$ & $5.332$ & $0.816$ & $0.140$ & $0.505$ & $0.631$ & $0.225$ & $8.321$ & $0.808$ & $0.370$ & $1.182$\\
    \methodName & $1.1$M & $0.872$ & $0.122$ & $3.784$ & $0.934$ & $0.093$ & $0.354$ & $0.801$ & $0.151$ & $6.335$ & \underline{$0.845$} & \underline{$0.129$} & \underline{$0.577$}\\
    \end{tabular}
    }
    \label{tab:persp_exp}
\end{table*}

\section{Comparison with UniK3D}
As mentioned in \cref{sec:sota_comp}, the comparison with \uniKThreeD~\cite{piccinelli2025unik3d} is not fair to \methodName, since \cite{piccinelli2025unik3d} is trained on large-FoV images.
However, we note that the comparison is also unfair towards \cite{piccinelli2025unik3d} since \methodName requires ground-truth camera parameters while \cite{piccinelli2025unik3d} doesn't.

For a fairer comparison, we employ AnyCalib~\cite{tirado2025anycalib}, an off-the-shelf camera intrinsics estimation model, and utilize the predicted intrinsics for ERP transformations. 
\cite{tirado2025anycalib} predicts intrinsics for KB~\cite{kannala2006generic} and UCM~\cite{kannala2006generic} camera models.
\scannetpp~\cite{yeshwanth2023scannet++} follows the KB model, whereas \kittiThreeSixty~\cite{liao2022kitti} follows the MEI~\cite{mei2007single} model, which is not handled by \cite{tirado2025anycalib}.
We approximate the MEI~\cite{mei2007single} model from UCM~\cite{geyer2000unifying} by setting the distortion parameters to zero.

\cref{tab:abl_intr_pred} provides the performance comparison between \cite{piccinelli2025unik3d} and \methodName using predicted and ground-truth intrinsics.
`+\textit{\aTwoDTwo}' denotes adding \aTwoDTwo~\cite{geyer2020a2d2} in the training data as detailed in \cref{sec:abl_train_data}.
We observe that even under this fairer comparison, we still outperform \cite{piccinelli2025unik3d} on \scannetpp~\cite{yeshwanth2023scannet++}.
We attribute the decrease in the performance on \kittiThreeSixty~\cite{liao2022kitti} to the approximation of predicted intrinsics from UCM~\cite{geyer2000unifying} to the MEI~\cite{mei2007single} model.
We believe that with better intrinsic estimation models that can handle the MEI~\cite{mei2007single} camera model, \methodName will retain its performance even with predicted intrinsics.

\begin{table}[!h]
    \centering
    \caption{\textbf{Ablation on Encoder Weight}. D2, D3 indicate the ViT encoders have been initialized with DINOv2 and DINOv3 pre-trained weights, respectively. [Key: \textbf{Best}]}
    \resizebox{\linewidth}{!}{%
    \begin{tabular}{l|ccc|ccc}
    \toprule
    \multirow{2}*{\textbf{Method}} & \multicolumn{3}{c|}{\textbf{\scannetpp}} & \multicolumn{3}{c}{\textbf{\kittiThreeSixty}} \\
     & $\delta_1 \uparrow$ & \absRel $\downarrow$ & \rmse $\downarrow$ & $\delta_1 \uparrow$ & \absRel $\downarrow$ & \rmse $\downarrow$\\
    \midrule
    \dacUni - D2 & $0.368$ & $0.305$ & $0.939$ & $0.334$ & $0.318$ & $6.872$\\
    \dacUni - D3 & $0.707$ & $0.211$ & $0.471$ & $0.428$ & $0.342$ & $5.305$\\
    \methodName - D2 & $0.533$ & $0.242$ & $0.703$ & $0.445$ & $0.287$ & $6.924$\\
    \methodName - D3 & \boldmath$0.792$ & \boldmath$0.140$ & \boldmath$0.396$ & \boldmath$0.622$ & \boldmath$0.239$ & \boldmath$5.057$\\
    \end{tabular}
    }
    \label{tab:abl_enc}
\end{table}

\section{Evaluation on Perspective Datasets}
We compare \methodName against our baselines on four perspective datasets, \kitti~\cite{geiger2012we}, \nyuDepth~\cite{silberman2012indoor}, \ibims~\cite{koch2018evaluation}, and \nuscenes~\cite{caesar2020nuscenes}.
While \cite{geiger2012we,silberman2012indoor,koch2018evaluation} provide artifact-free depthmaps in their official dataset, we utilize \cite{zhu2024replay} to estimate artifact-free depthmaps for \cite{caesar2020nuscenes}.
We observe from \cref{tab:persp_exp} that \methodName outperforms \uniKThreeD and \dac on two important perspective benchmarks, namely, \kitti and \nyuDepth, demonstrating the generalization capability of \methodName even in perspective datasets, beyond large-FoV datasets.

\begin{table}[!h]
    \centering
    \caption{\textbf{Ablation on Shift Estimation.} We study the impact of depth-guided shift map estimation. [Key: \textbf{Best}]}
    \vspace{-3mm}
    \resizebox{\linewidth}{!}{%
    \begin{tabular}{l|ccc|ccc}
    \toprule
    \multirow{2}*{\textbf{Shift}} & \multicolumn{3}{c|}{\textbf{\scannetpp}} & \multicolumn{3}{c}{\textbf{\kittiThreeSixty}} \\
     & $\delta_1 \uparrow$ & \absRel $\downarrow$ & \rmse $\downarrow$ & $\delta_1 \uparrow$ & \absRel $\downarrow$ & \rmse $\downarrow$\\
     \midrule
    $\shift \in \realDomain$ & 
    $0.792$ & $0.140$ & $0.396$ & $0.622$ & $0.239$ & $5.057$\\
    $\shiftMap \in \realDomain^{H\times W}$ & \boldmath$0.798$ & \boldmath$0.139$ & \boldmath$0.393$ & \boldmath$0.630$ & \boldmath$0.235$ & \boldmath$4.985$\\
    \end{tabular}
    }
    \label{tab:abl_shift}
\end{table}

\section{Ablation on Encoder Weights}
\cref{tab:abl_enc} evaluates the effect of initializing encoders $\encoder$ with different pre-trained weights on the model performance.
We train \dacUni and \methodName using DINOv2 and DINOv3 encoders on \hmThreeD and \ddad datasets.
While \dac's proposed framework is compatible with any depth estimation model, they use \iDisc~\cite{piccinelli2023idisc} for its simplicity and effectiveness.
\iDisc requires multi-scale features from the encoder for its pipeline.
Since DINO features are at a downscaled resolution, we apply consecutive upsampling via transpose convolutions to obtain multi-resolution features.

One of the major differences between DINOv2 and DINOv3 is the positional embedding scheme.
DINOv2 uses additive absolute positional embedding, whereas DINOv3 utilizes 2D-RoPE~\cite{heo2024rotary}.
 We do not modify the positional embedding scheme of the DINO encoders, and thus the overall performance of \dac and \methodName is affected by the compatibility of the positional embeddings with the respective frameworks and the task of large-FoV depth estimation.

We observe that utilizing DINOv3 as the encoder gives the best performance for both \dac and \methodName.
Since we train on small-FoV perspective images and test on large-FoV images, the absolute positional embedding of DINOv2 is not suitable for the task.
However, the RoPE in DINOv3 offers relative positional embedding, thus facilitating generalization.

\iDisc, utilized by \dac, internally uses absolute positional embedding for their proposed Internal Discretization Module.
Therefore, the proposed method in \methodName is most compatible with the DINOv3 encoder, giving the best performance.
We believe the mismatch between the RoPE in DINOv3 and the absolute positional embedding in \iDisc architecture to be one of the reasons for the performance gap between \dac-D3 and \methodName-D3.

We also observe that the performance of \dac-D3 on \kittiThreeSixty is quite low compared to \methodName-D3, underscoring the benefit of our proposed depth-guided scale estimation module.

\section{Ablation on Shift Estimation}
As mentioned in \cref{sec:scale_est}, we estimate a scale map $\scaleMap$ instead of a 1-D scalar $\scale$ to adjust for irregularities.
However, we still estimated shift $\shift$ as a 1-D scalar.
\cref{tab:abl_shift} provides an ablation on estimating a shift scalar and a shift map while keeping scale estimation in the form of a scale map.
Formally, we modify the architecture slightly to output $\{\scaleMap, \shiftMap\} \in \realDomain^{H\times W}$.
As expected, we can observe from \cref{tab:abl_shift} that there is a slight increase in performance by incorporating a shift map.
However, the improvement from adopting the shift map is still smaller than the improvement from adopting the scale map, as seen in \cref{tab:abl_scale}.


\begin{figure*}[b]
    \centering
    \begin{minipage}[t]{\linewidth}
        \centering
        \includegraphics[width=0.23\linewidth]{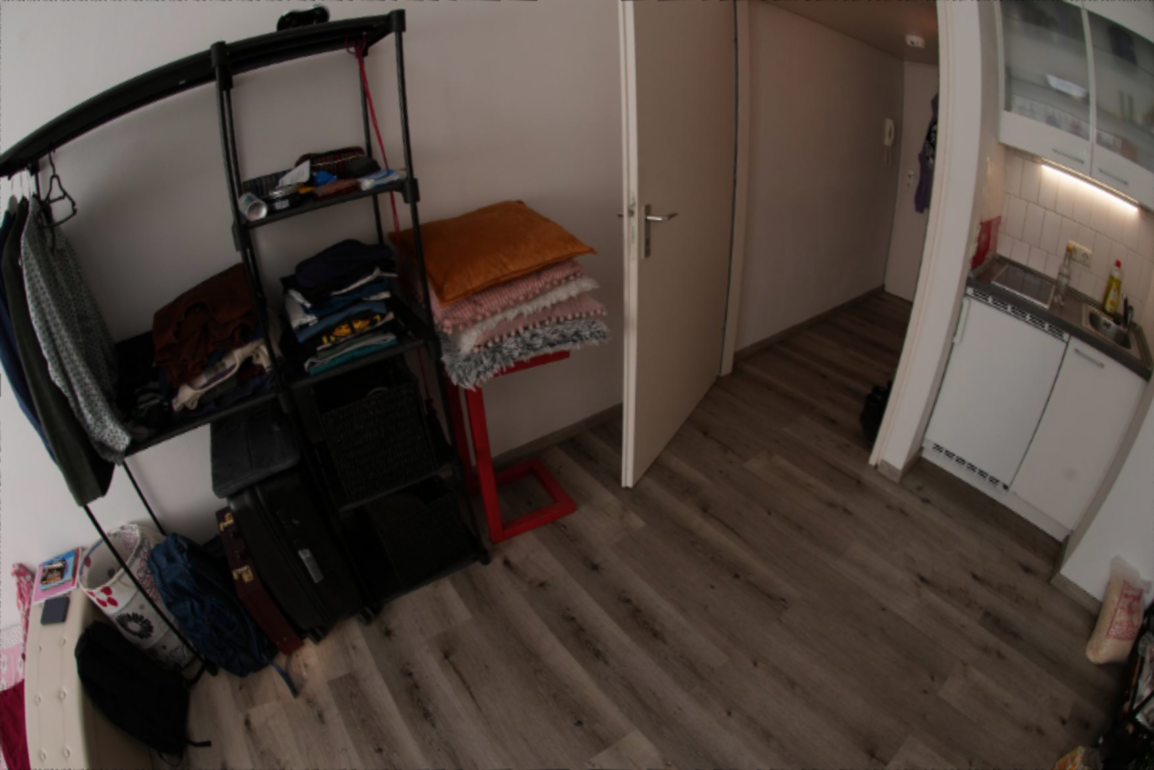}
        \includegraphics[width=0.23\linewidth]{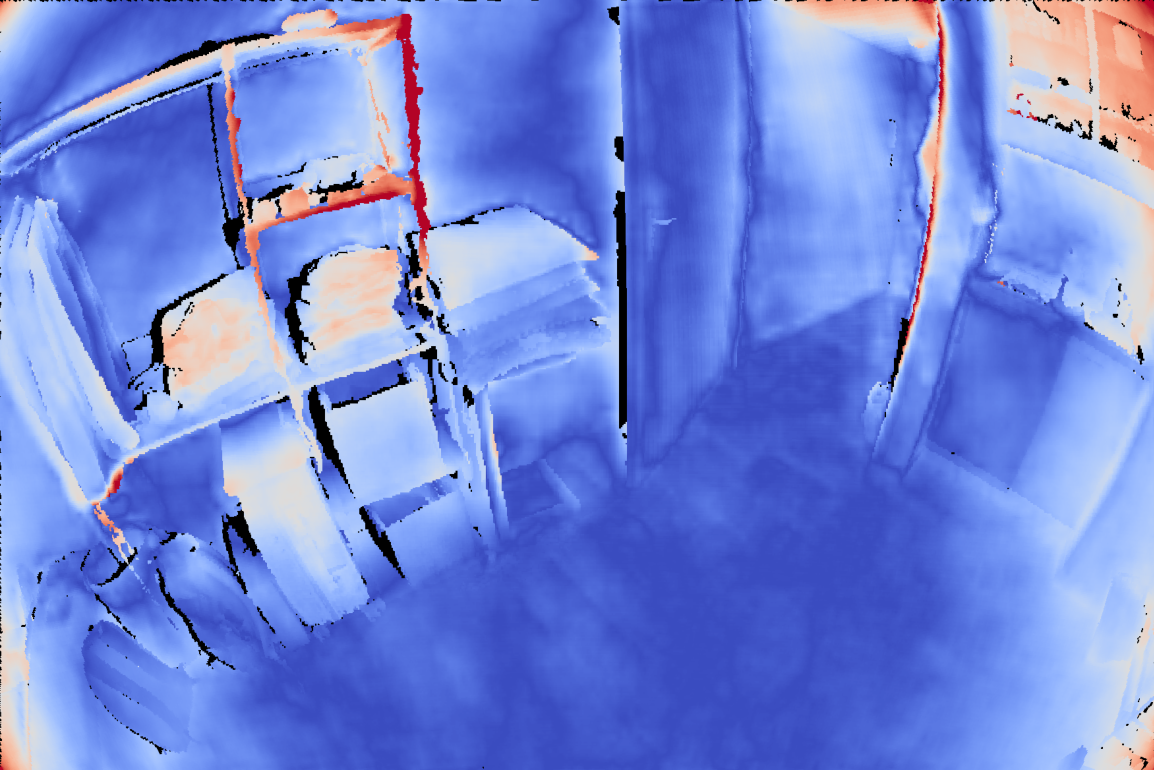}
        \includegraphics[width=0.23\linewidth]{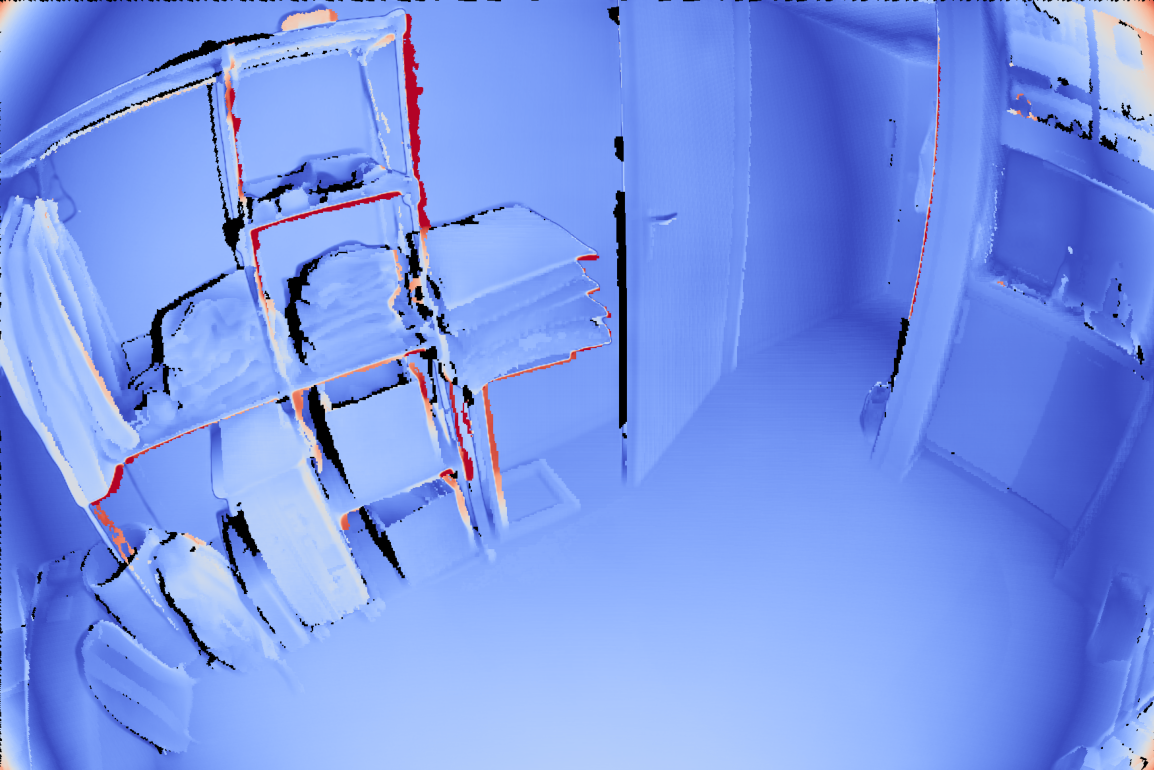}
        \includegraphics[width=0.23\linewidth]{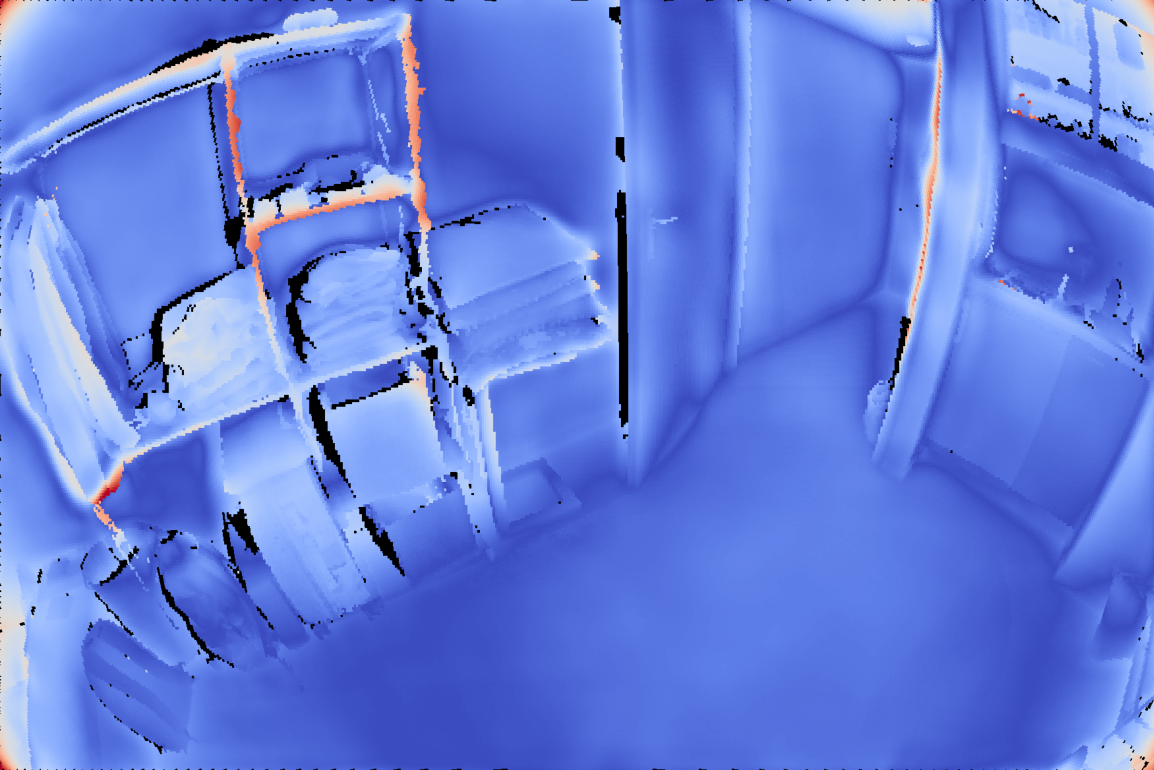}
        \raisebox{0.7ex}{\includegraphics[width=0.05\linewidth]{figures/qualitative/dac/scannetpp/arel.jpg}}
    \end{minipage}
    \begin{minipage}[t]{\linewidth}
        \centering
        \includegraphics[width=0.23\linewidth]{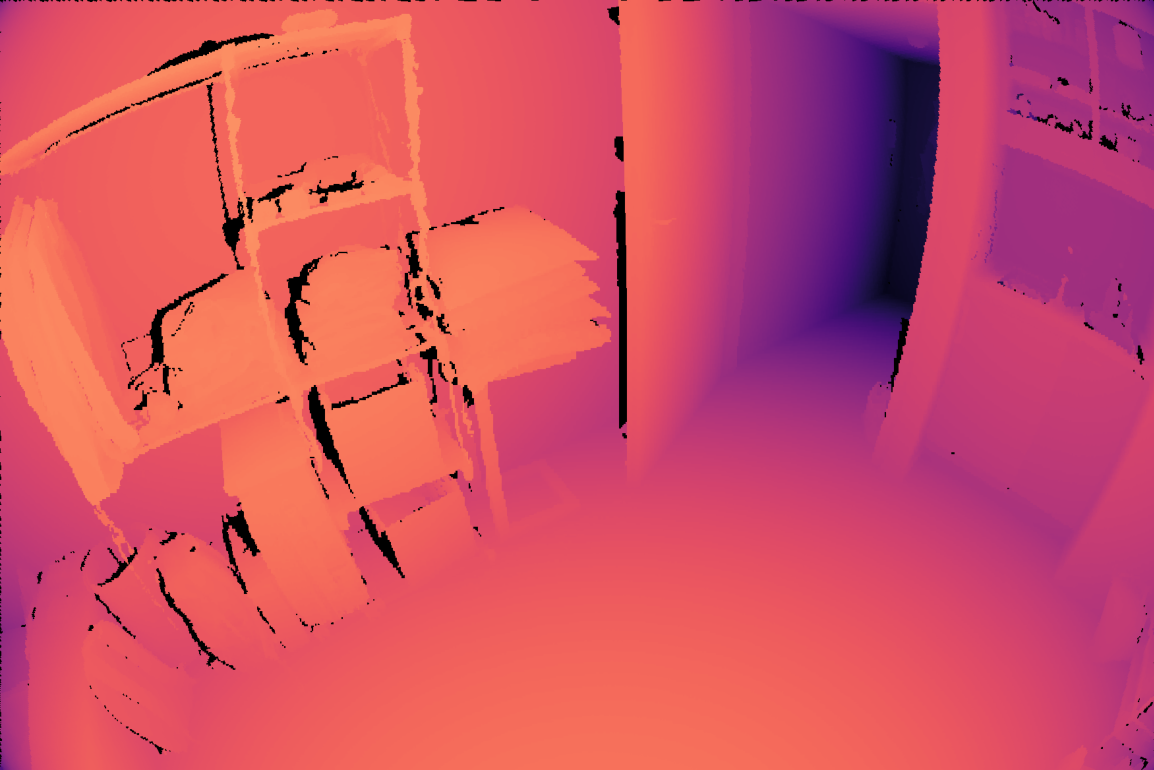}
        \includegraphics[width=0.23\linewidth]{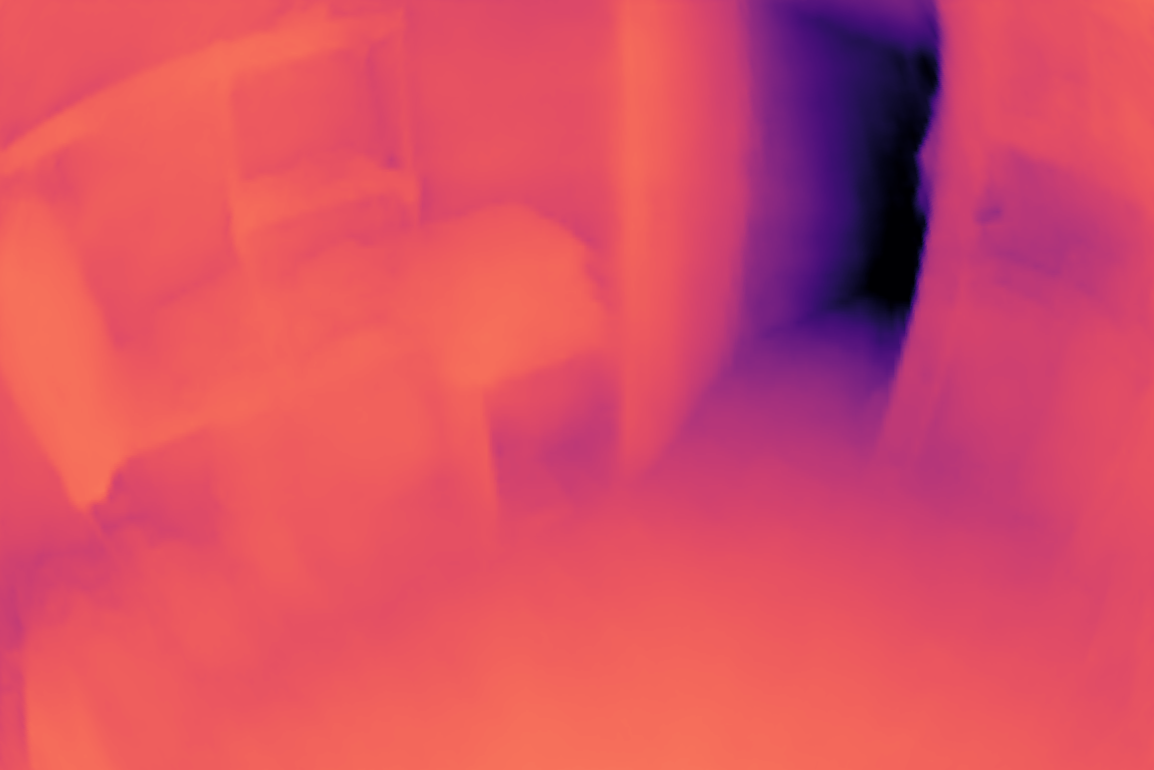}
        \includegraphics[width=0.23\linewidth]{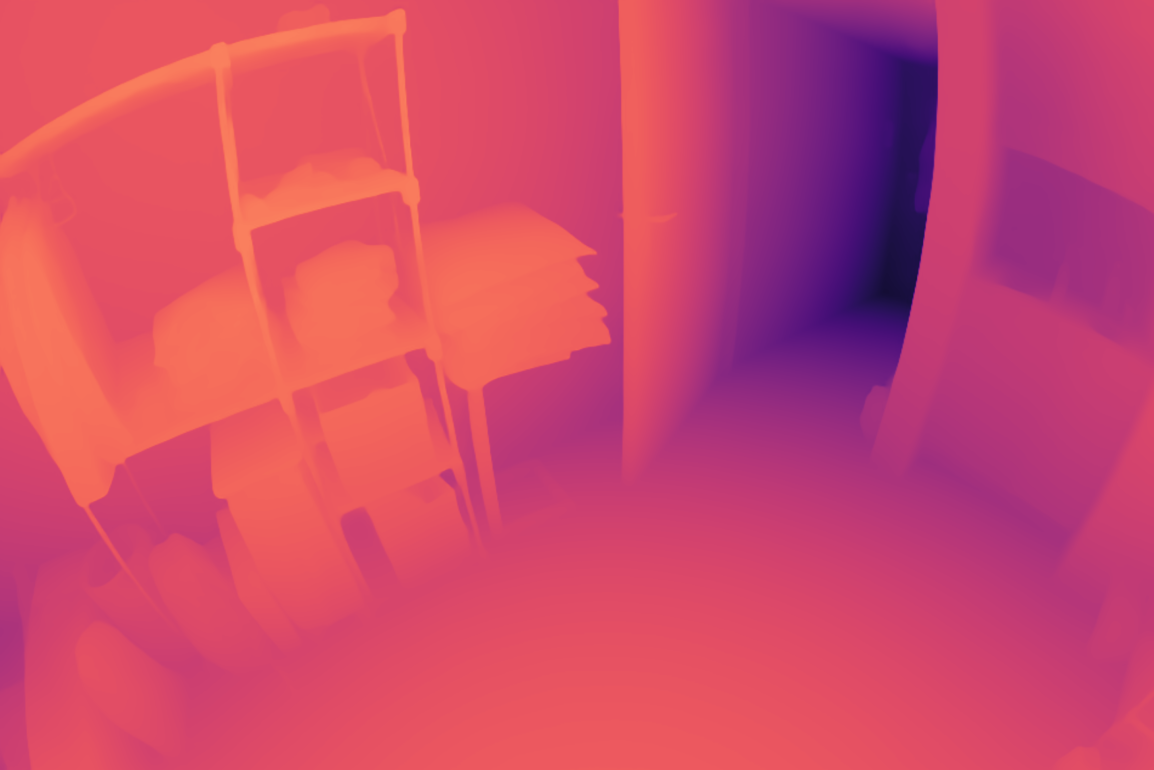}
        \includegraphics[width=0.23\linewidth]{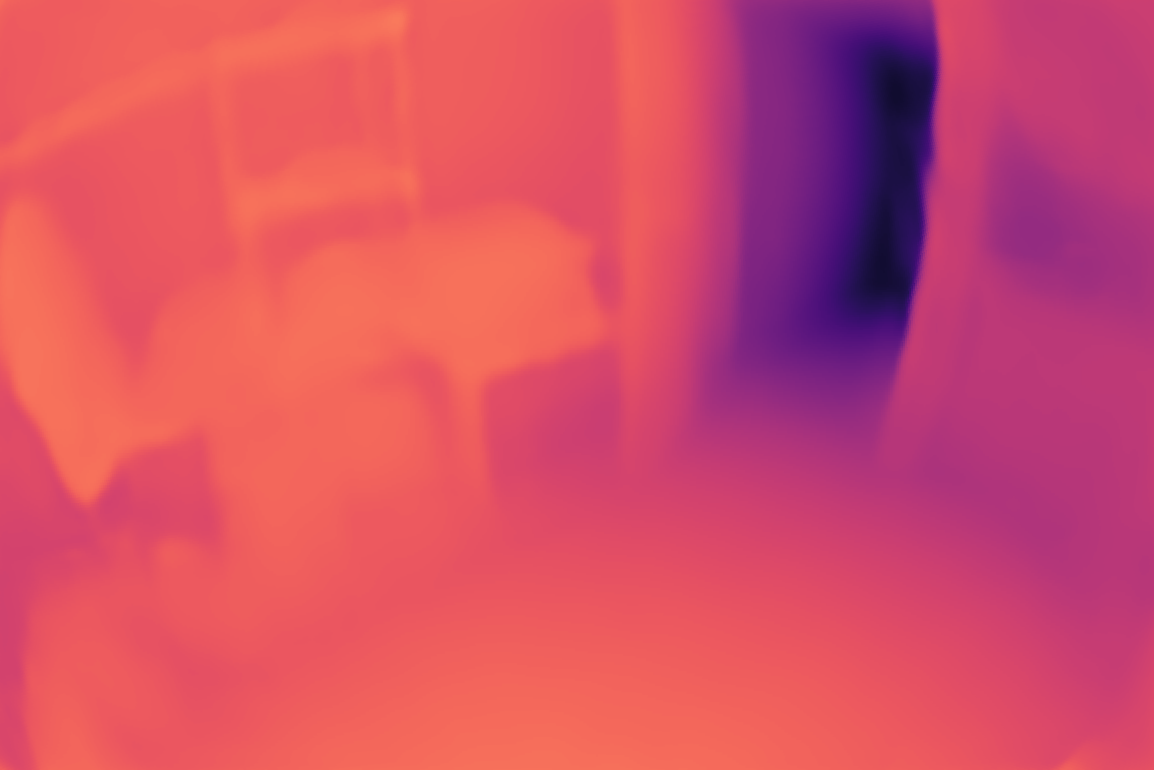}
        \raisebox{0.5ex}{\includegraphics[width=0.05\linewidth]{figures/qualitative/dac/scannetpp/meters_m.jpg}}
    \end{minipage}
    \begin{minipage}[t]{\linewidth}
        \centering
        \includegraphics[width=0.23\linewidth]{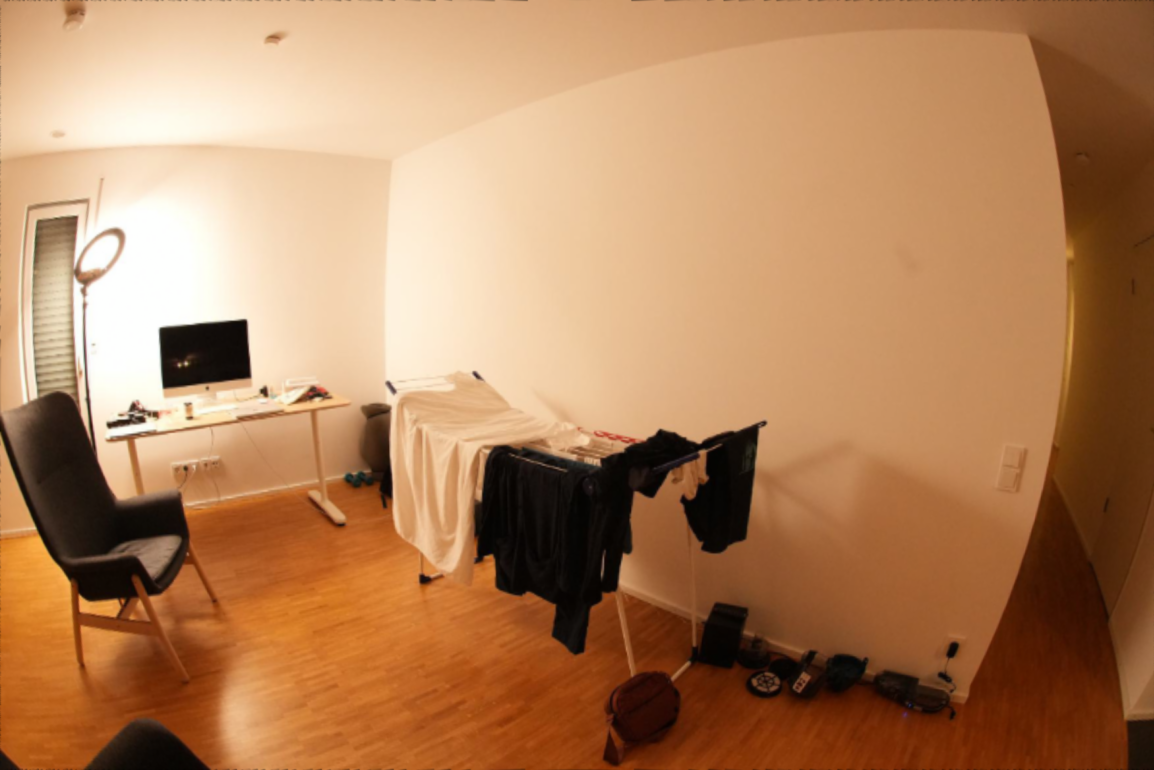}
        \includegraphics[width=0.23\linewidth]{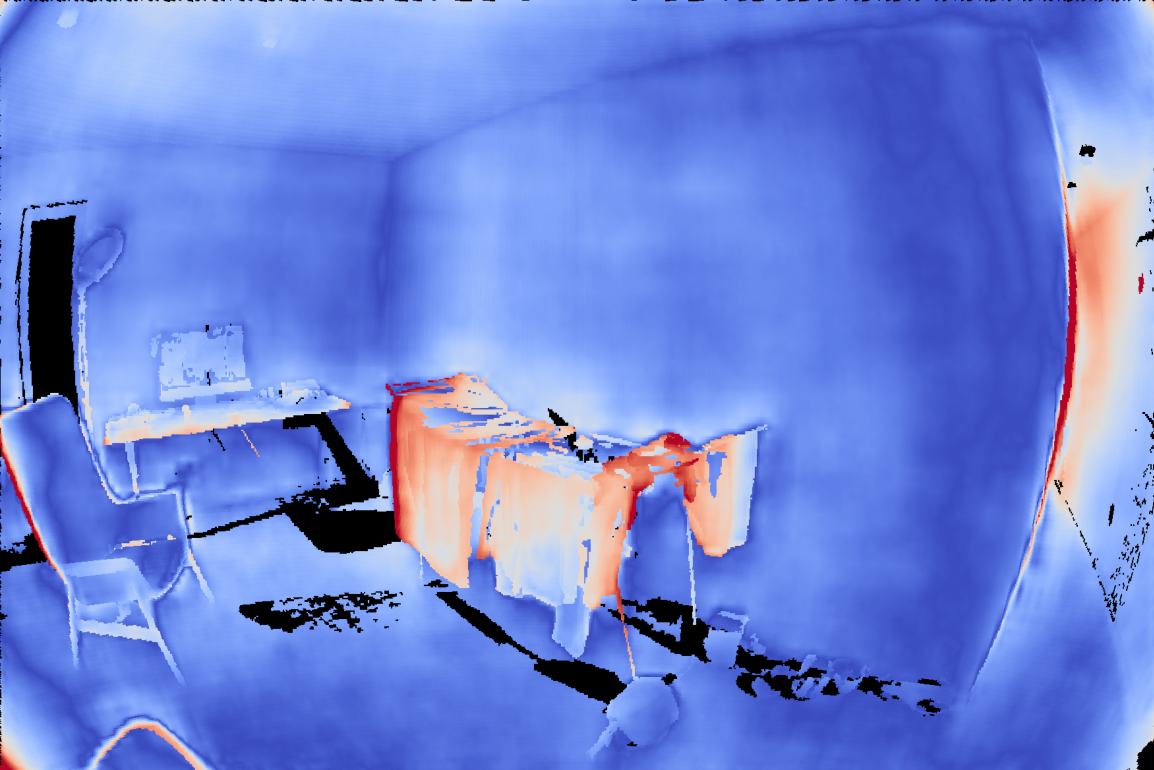}
        \includegraphics[width=0.23\linewidth]{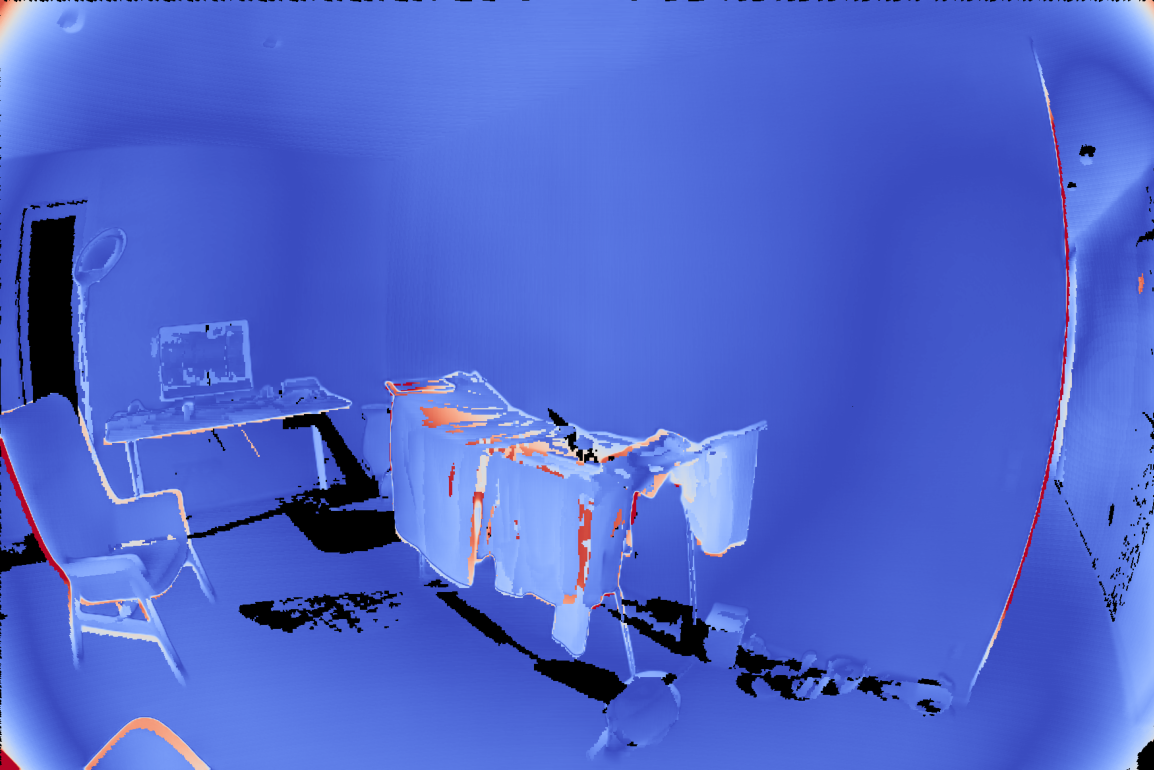}
        \includegraphics[width=0.23\linewidth]{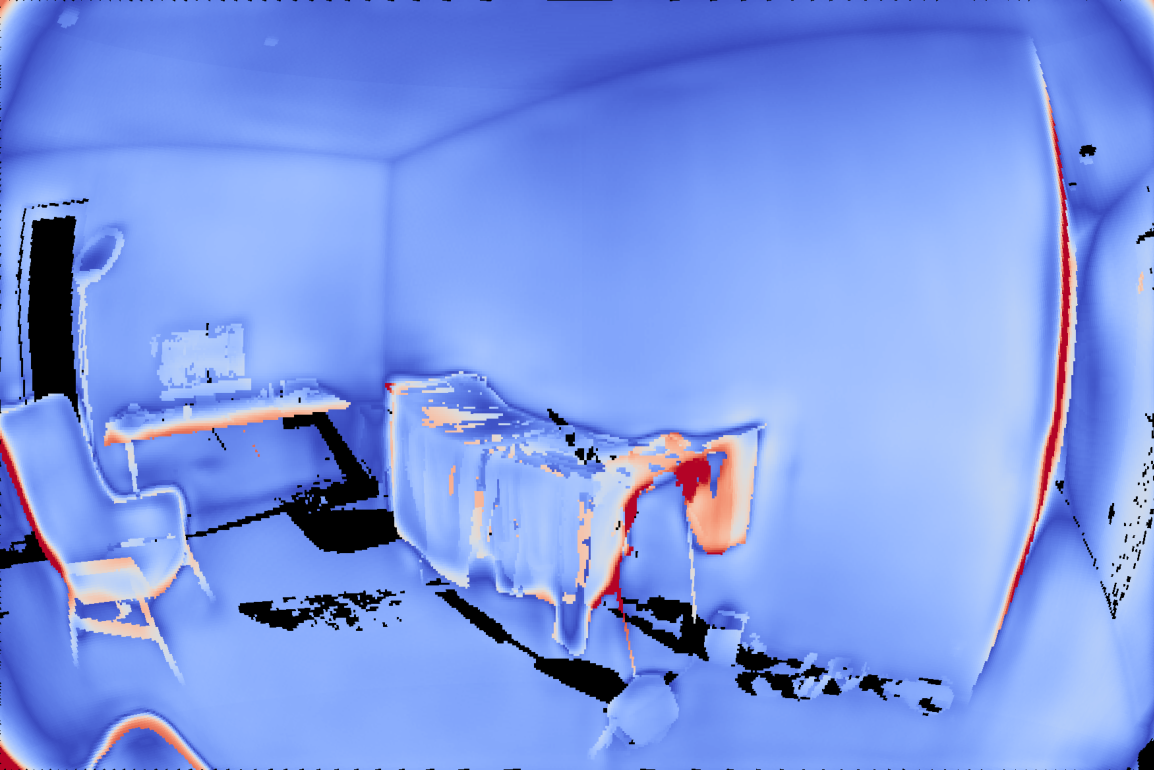}
        \raisebox{0.7ex}{\includegraphics[width=0.05\linewidth]{figures/qualitative/dac/scannetpp/arel.jpg}}
    \end{minipage}
    \begin{minipage}[t]{\linewidth}
        \centering
        \includegraphics[width=0.23\linewidth]{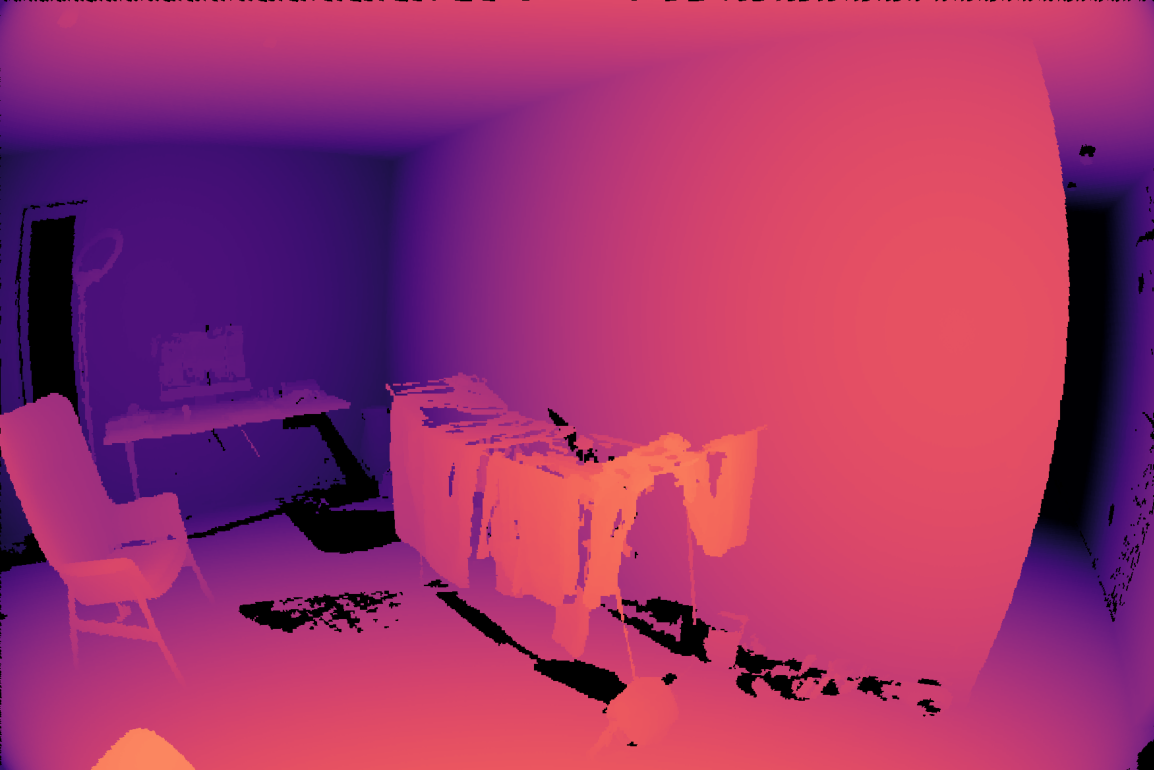}
        \includegraphics[width=0.23\linewidth]{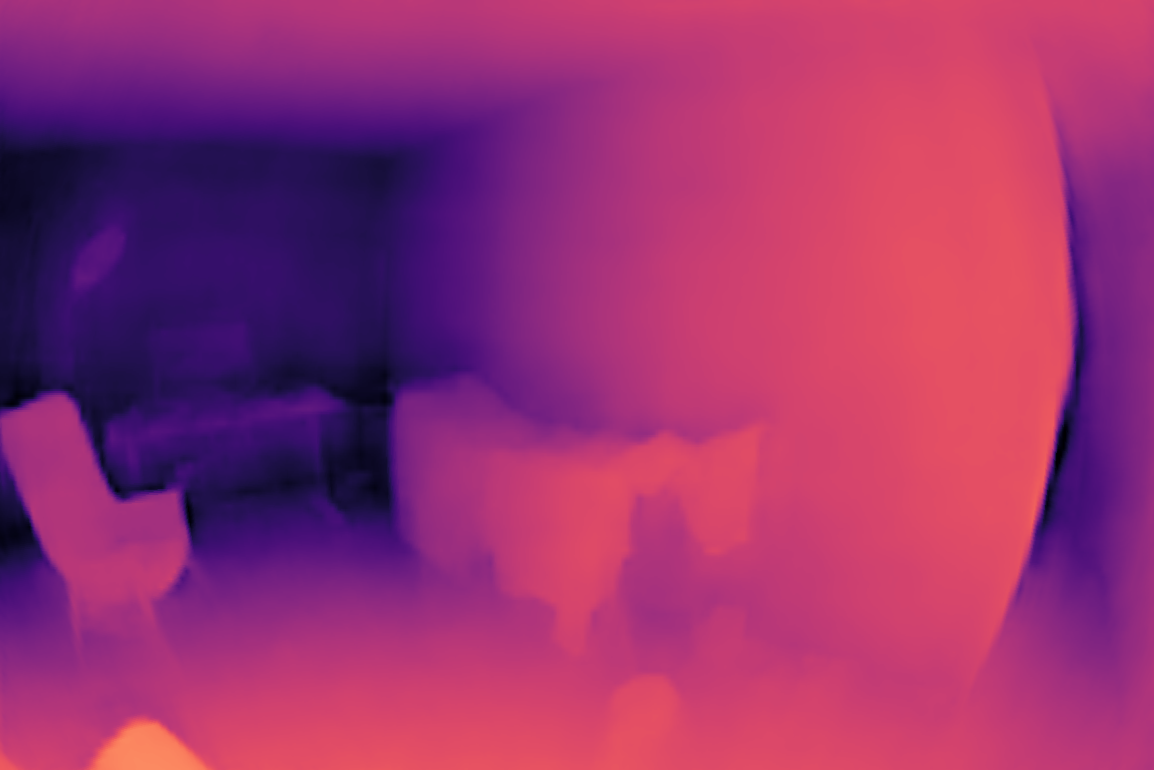}
        \includegraphics[width=0.23\linewidth]{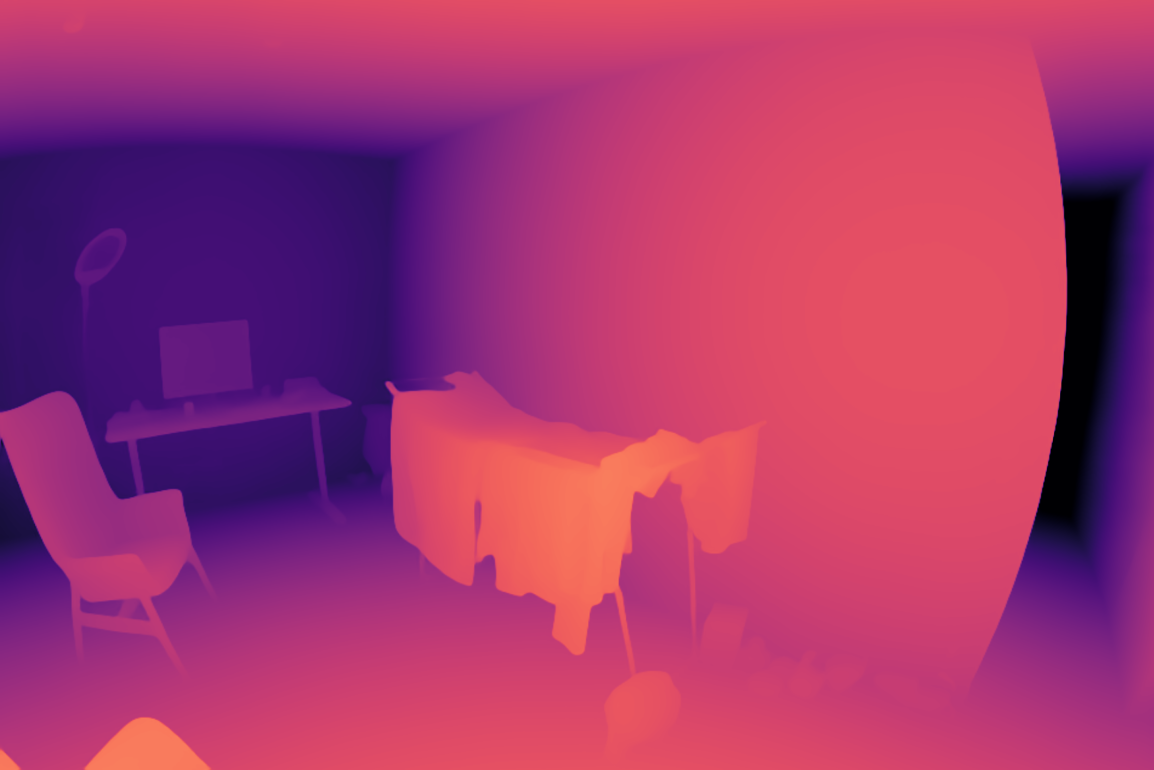}
        \includegraphics[width=0.23\linewidth]{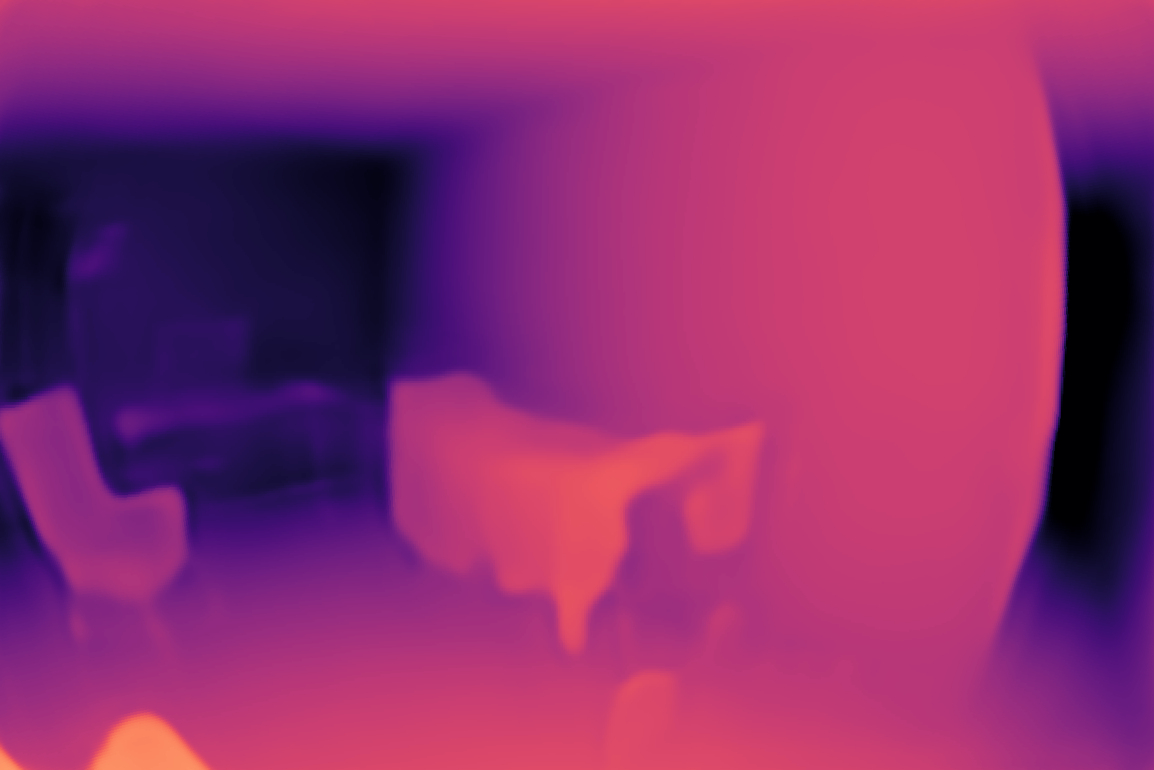}
        \raisebox{0.5ex}{\includegraphics[width=0.05\linewidth]{figures/qualitative/dac/scannetpp/meters_m.jpg}}
    \end{minipage}
    \begin{minipage}[t]{\linewidth}
        \centering
        \includegraphics[width=0.23\linewidth]{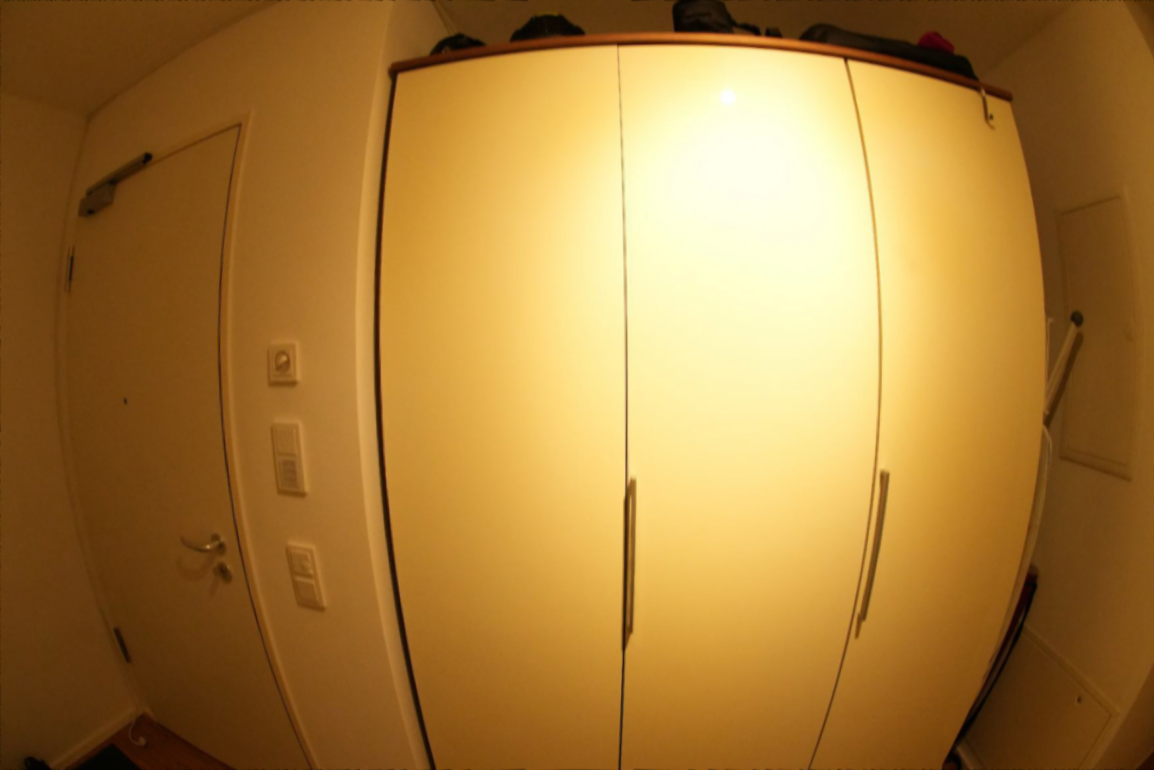}
        \includegraphics[width=0.23\linewidth]{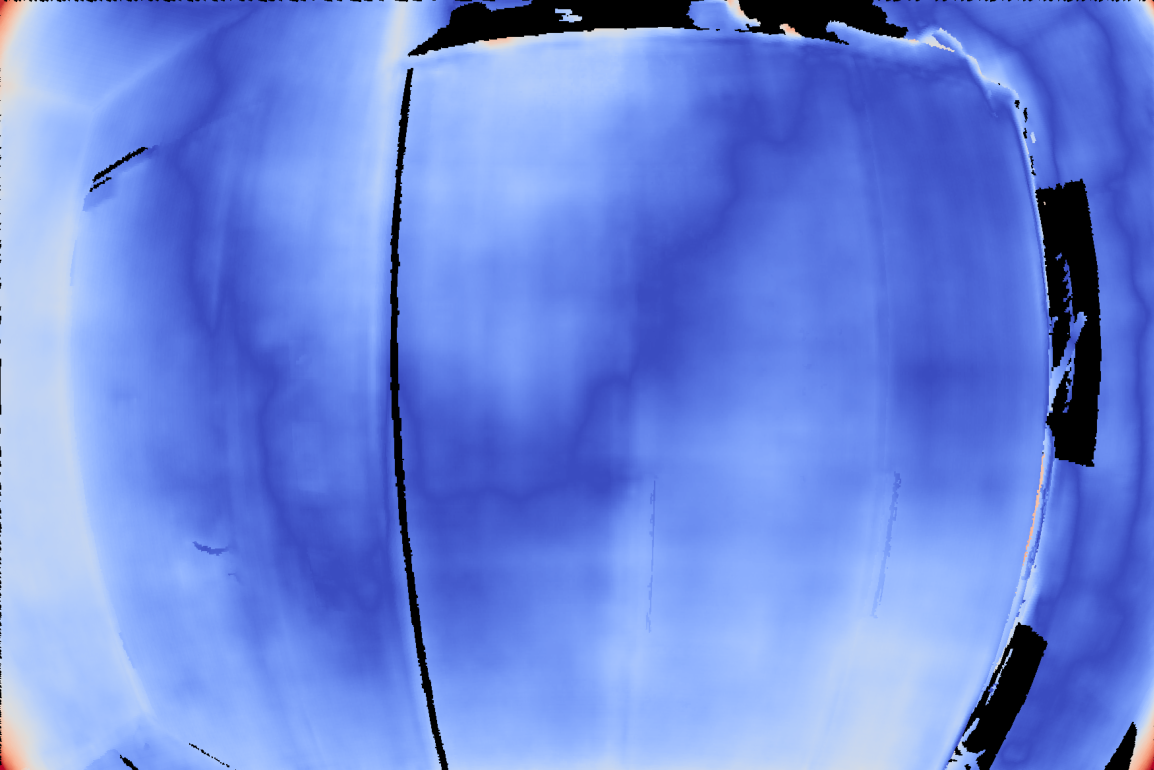}
        \includegraphics[width=0.23\linewidth]{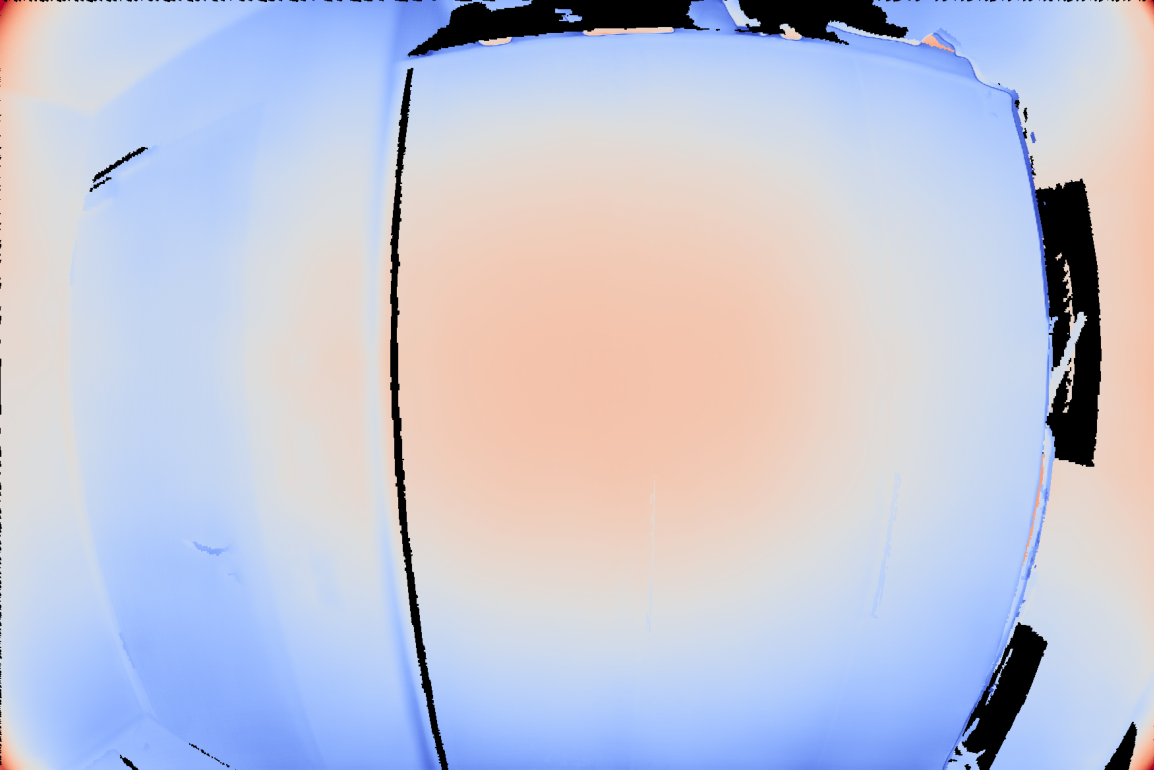}
        \includegraphics[width=0.23\linewidth]{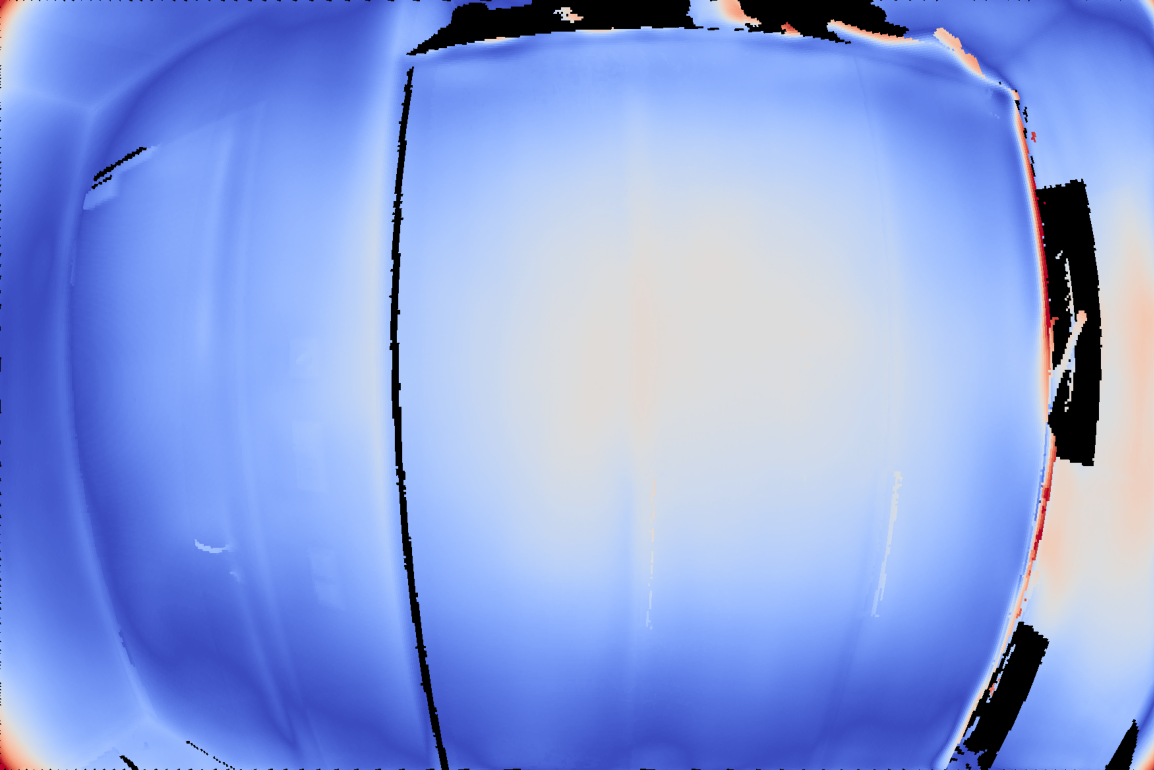}
        \raisebox{0.7ex}{\includegraphics[width=0.05\linewidth]{figures/qualitative/dac/scannetpp/arel.jpg}}
    \end{minipage}
    \begin{minipage}[t]{\linewidth}
        \centering
        \begin{tikzpicture}
            \draw (0,0) node[inner sep=0] {\includegraphics[width=0.23\linewidth]{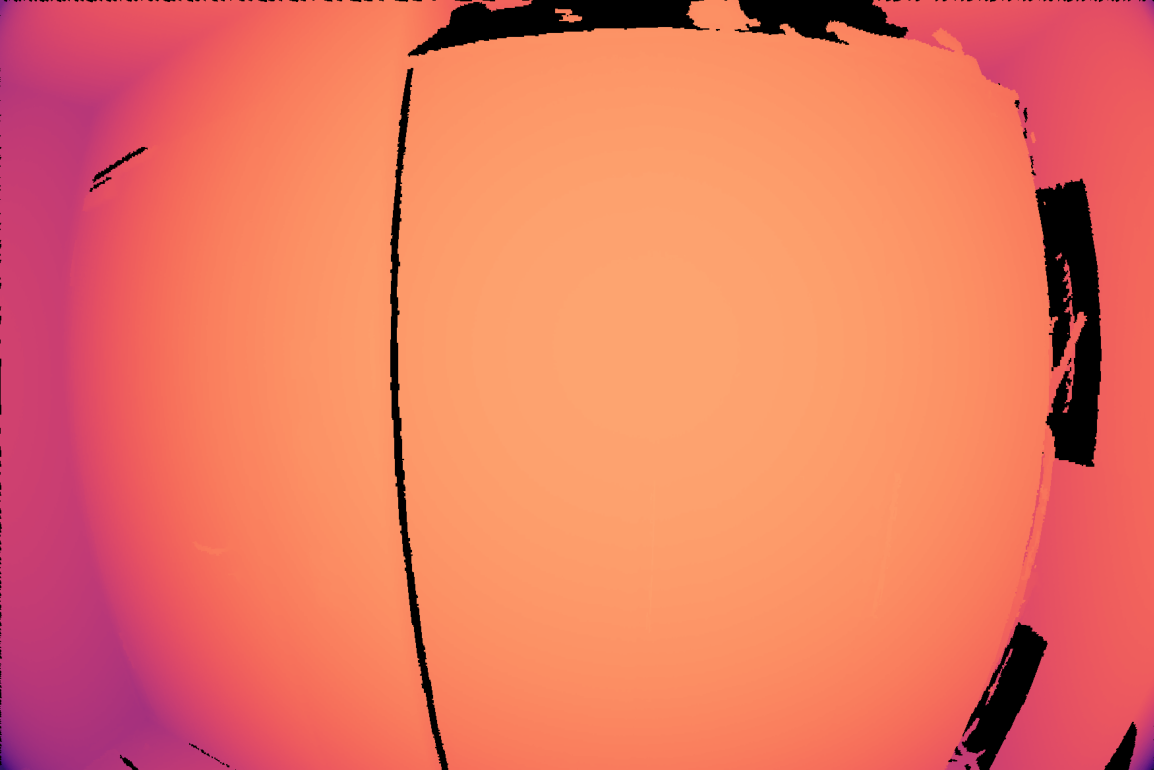}};
            \draw (0,-1.1-0.5) node[inner sep=0, align=center] {\fontsize{8.0}{10}\selectfont {RGB \& GT}};
        \end{tikzpicture}
        \begin{tikzpicture}
            \draw (0,0) node[inner sep=0] {\includegraphics[width=0.23\linewidth]{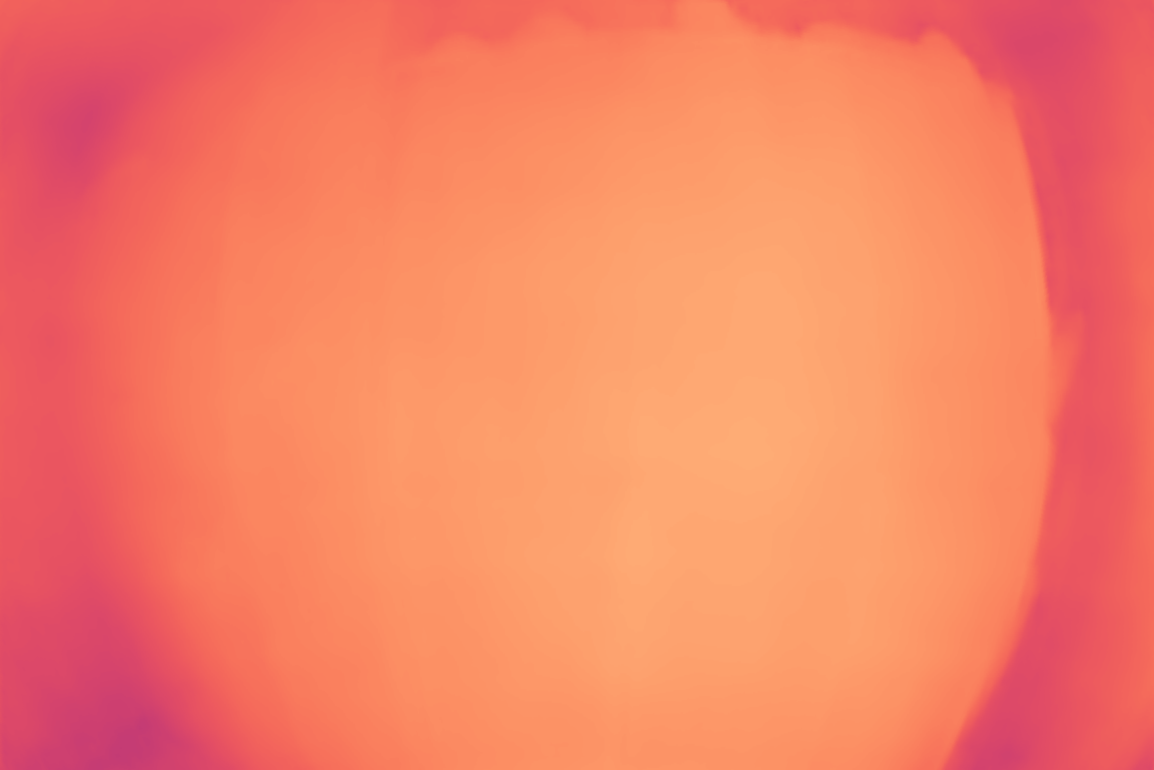}};
            \draw (0,-1.08-0.5) node[inner sep=0, align=center] {\fontsize{8.0}{10}\selectfont {\dacUni~\cite{guo2025depth}}};
        \end{tikzpicture}
        \begin{tikzpicture}
            \draw (0,0) node[inner sep=0] {\includegraphics[width=0.23\linewidth]{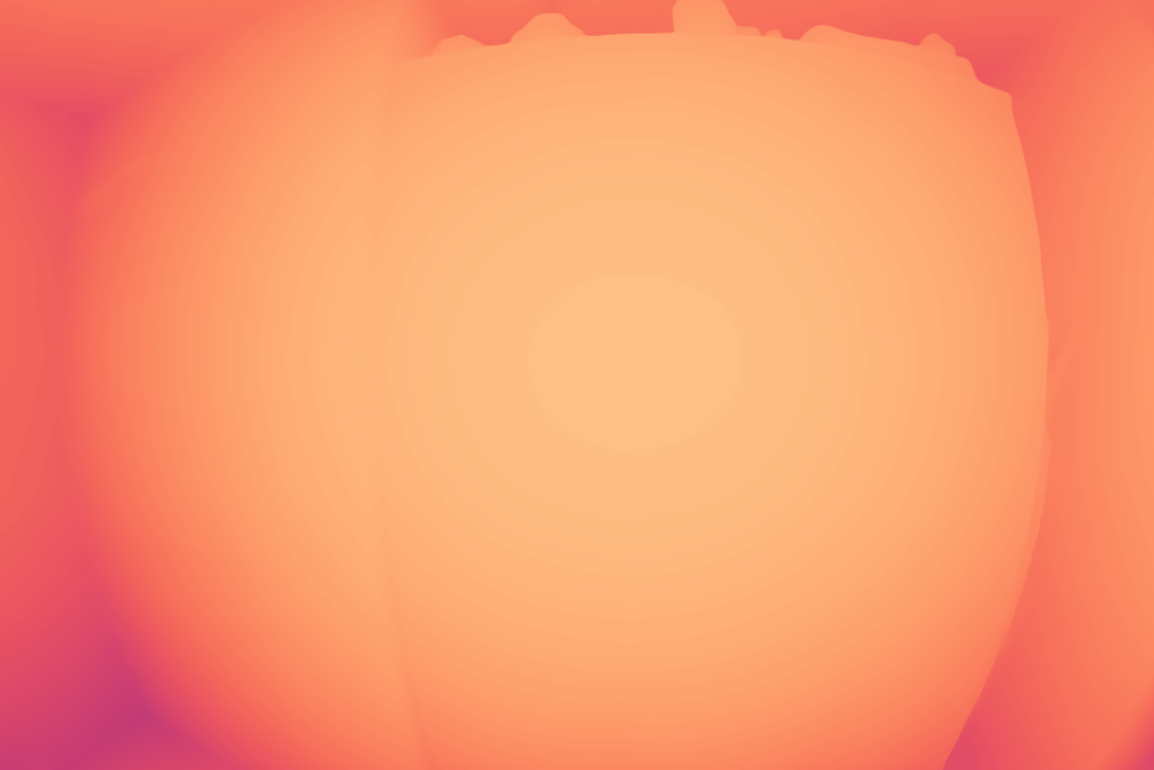}};
            \draw (0,-1.08-0.5) node[inner sep=0, align=center] {\fontsize{8.0}{10}\selectfont {\uniKThreeD~\cite{piccinelli2025unik3d}}};
        \end{tikzpicture}
        \begin{tikzpicture}
            \draw (0,0) node[inner sep=0, align=center] {\includegraphics[width=0.23\linewidth]{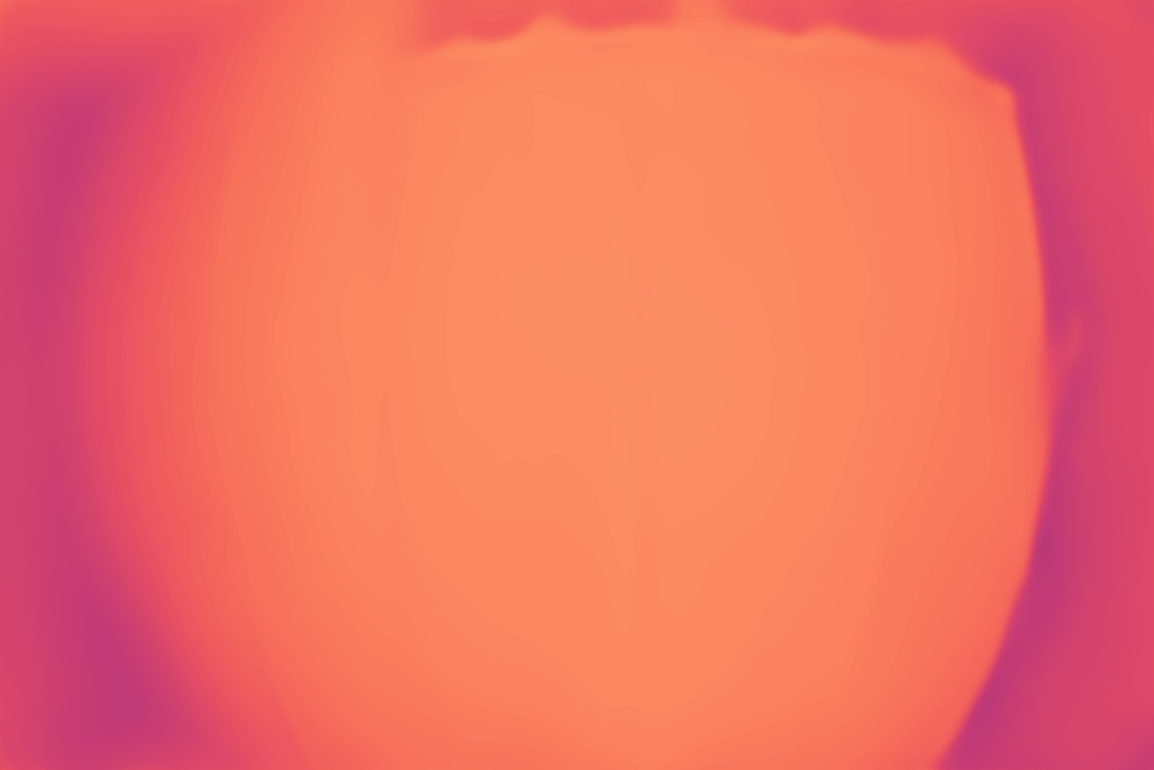}};
            \draw (0,-1.1-0.5) node[inner sep=0, align=center] {\fontsize{8.0}{10}\selectfont {\methodName}};
        \end{tikzpicture}
        \raisebox{4ex}{\includegraphics[width=0.05\linewidth]{figures/qualitative/dac/scannetpp/meters_m.jpg}}
    \end{minipage}
    \caption{\textbf{Qualitative Results on \scannetpp~\cite{yeshwanth2023scannet++}.} Every pair of consecutive rows corresponds to a single sample. Odd rows display the input RGB image, and \absRel error between predicted and GT depth maps. Even rows display the GT depth map and predicted depth maps.}
    \label{fig:supp_scannetpp}
\end{figure*}




\section{Additional Qualitative results}
\begin{figure*}
    \centering
    \begin{minipage}[t]{\linewidth}
        \centering
        \includegraphics[width=0.23\linewidth]{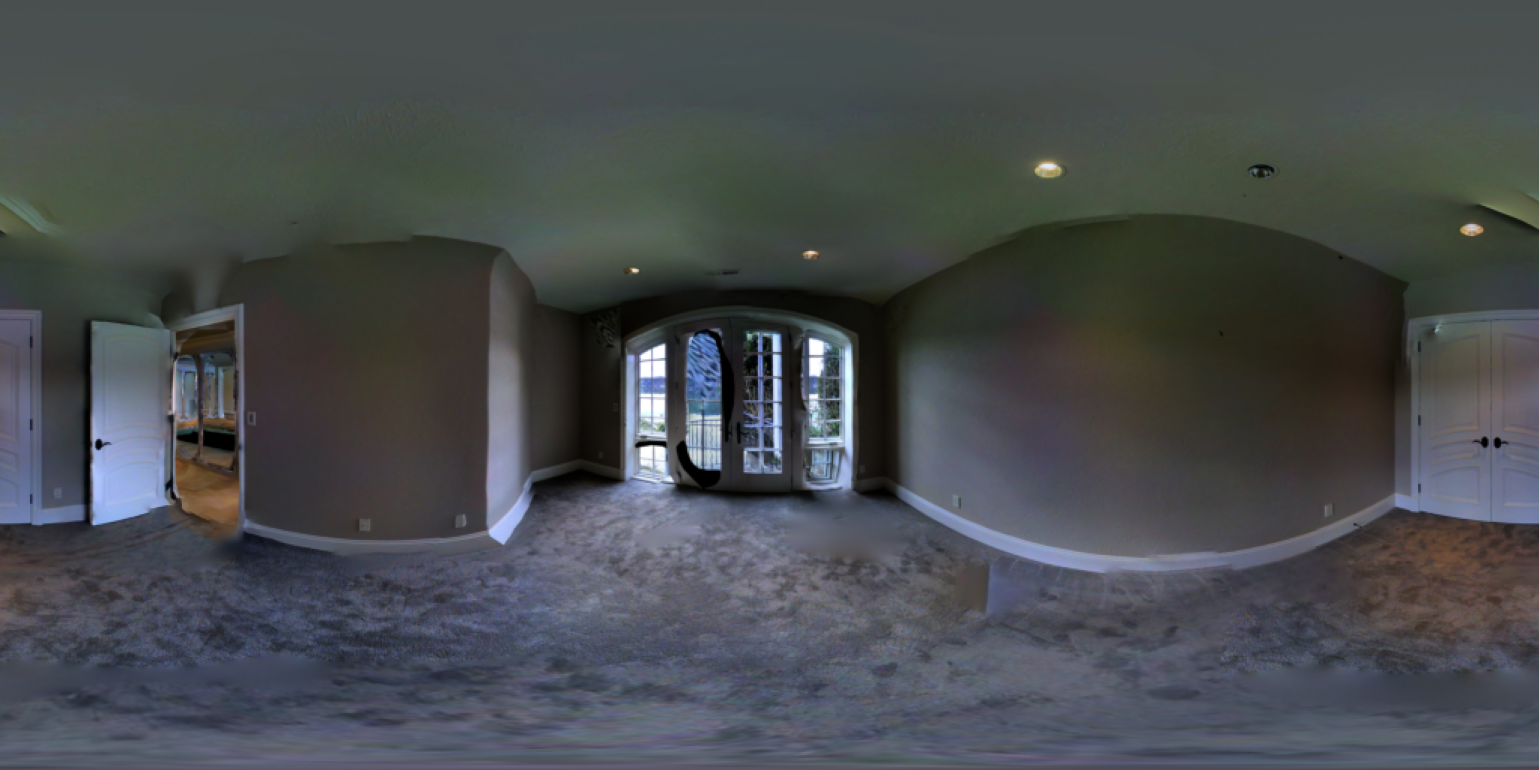}
        \includegraphics[width=0.23\linewidth]{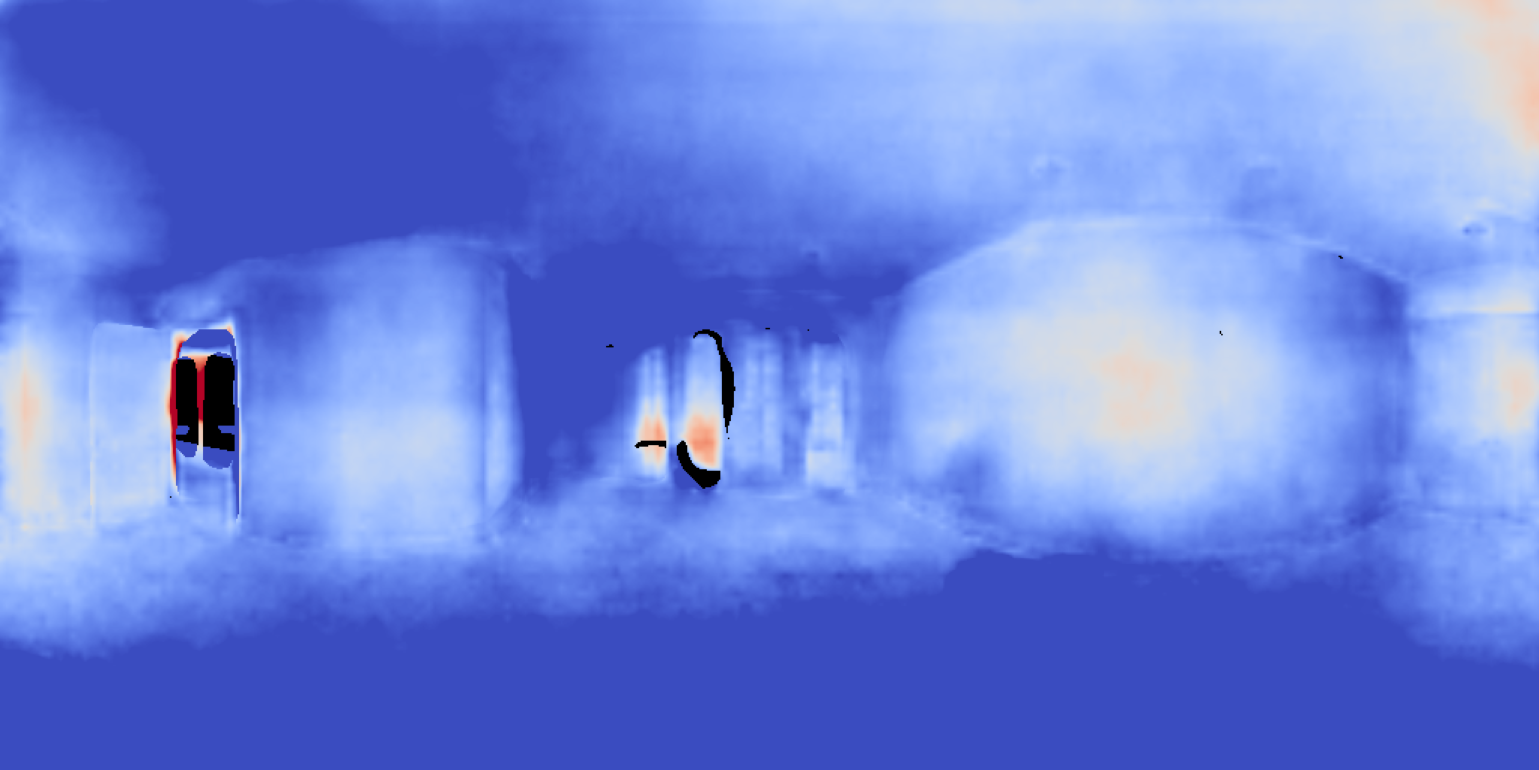}
        \includegraphics[width=0.23\linewidth]{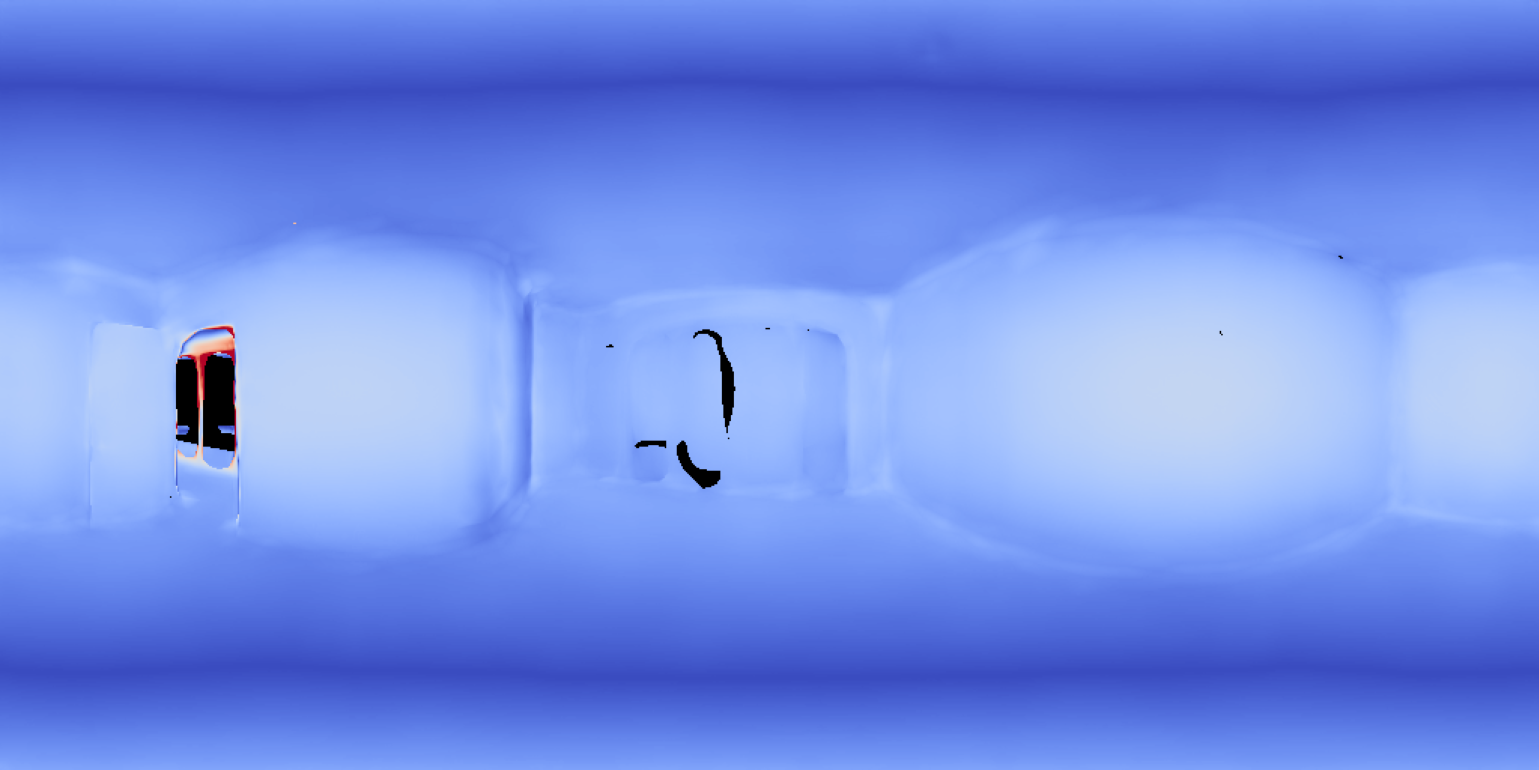}
        \includegraphics[width=0.23\linewidth]{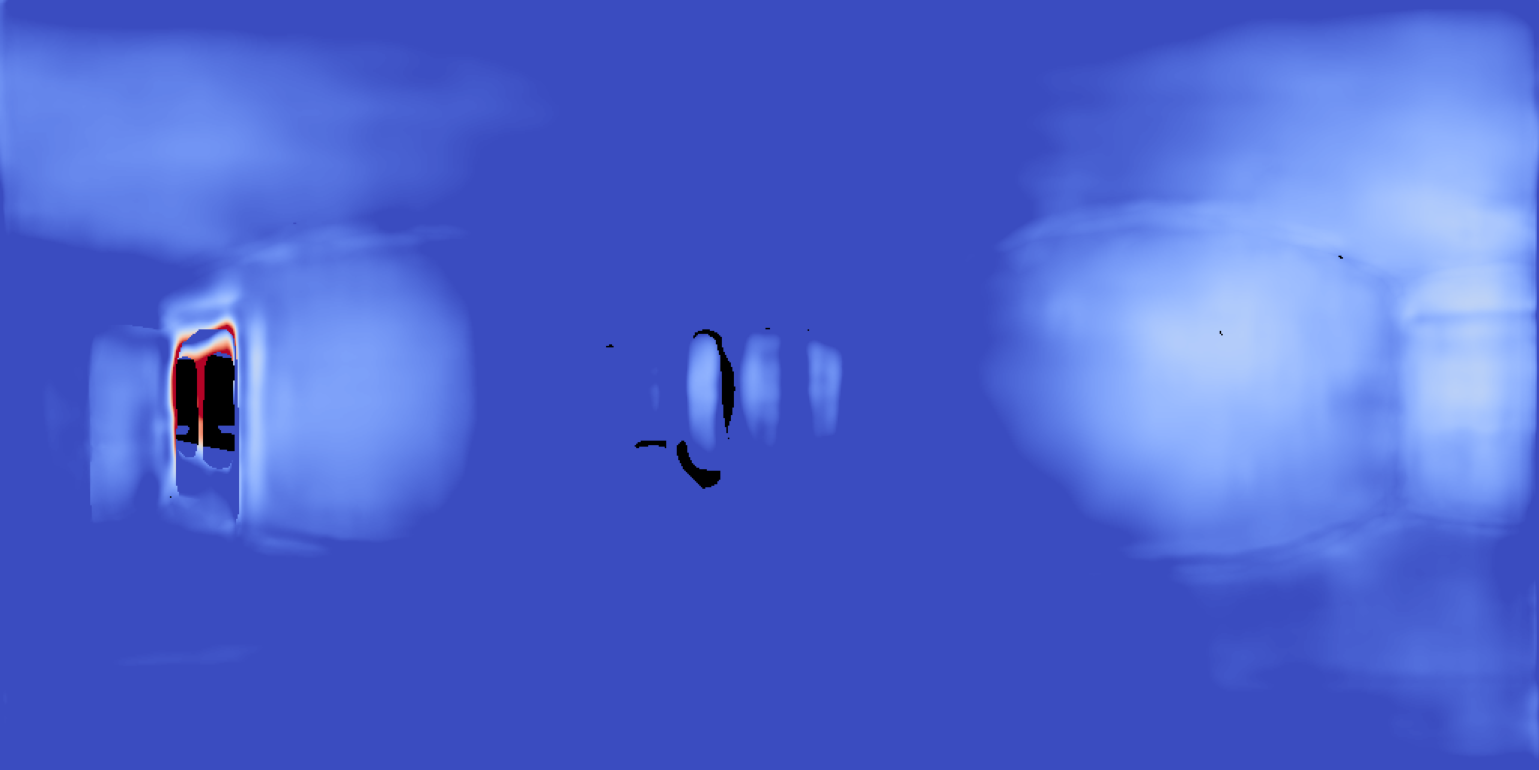}
        \raisebox{2ex}{\includegraphics[width=0.05\linewidth]{figures/qualitative/dac/kitti360/arel.jpg}}
    \end{minipage}
    \begin{minipage}[t]{\linewidth}
        \centering
        \includegraphics[width=0.23\linewidth]{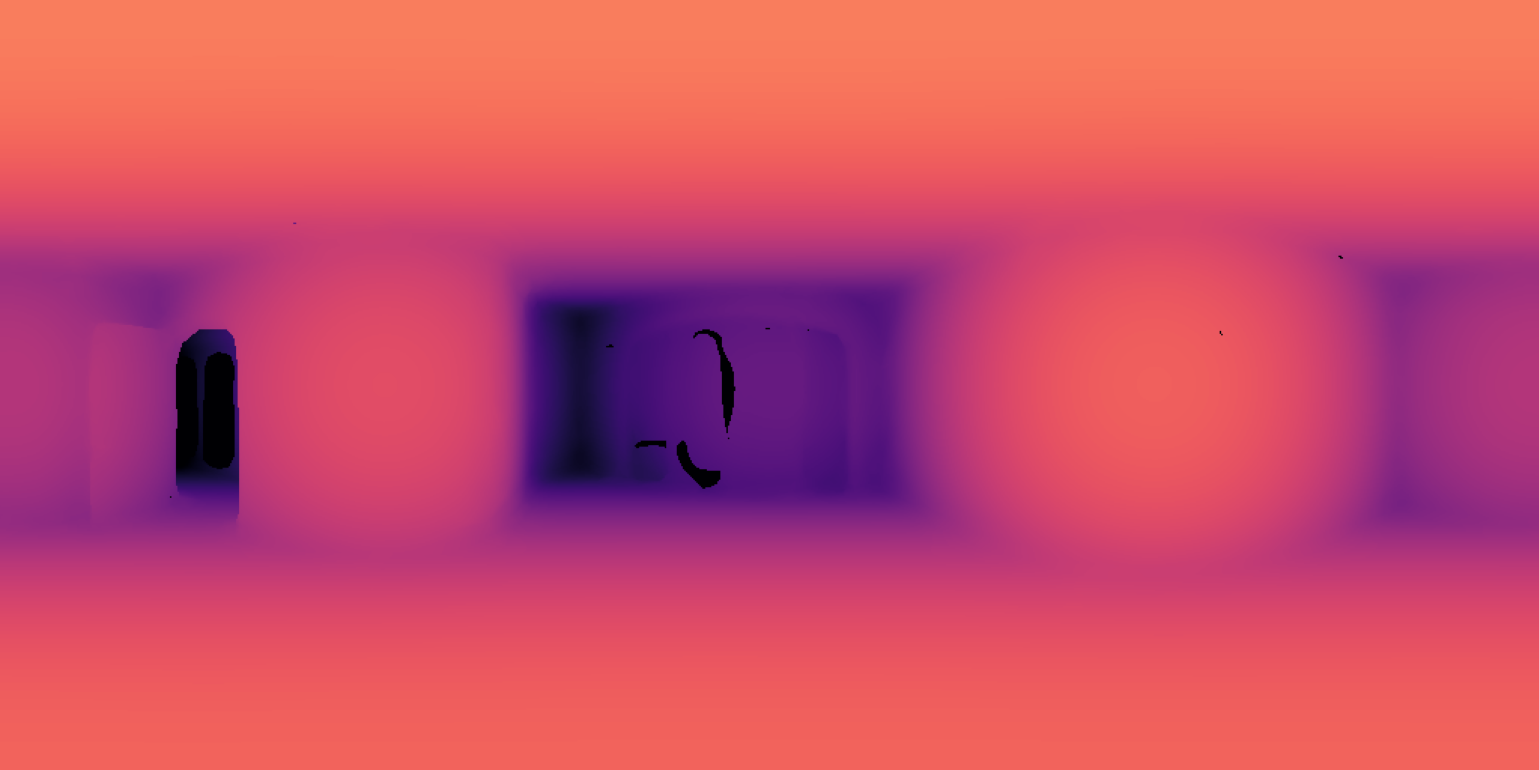}
        \includegraphics[width=0.23\linewidth]{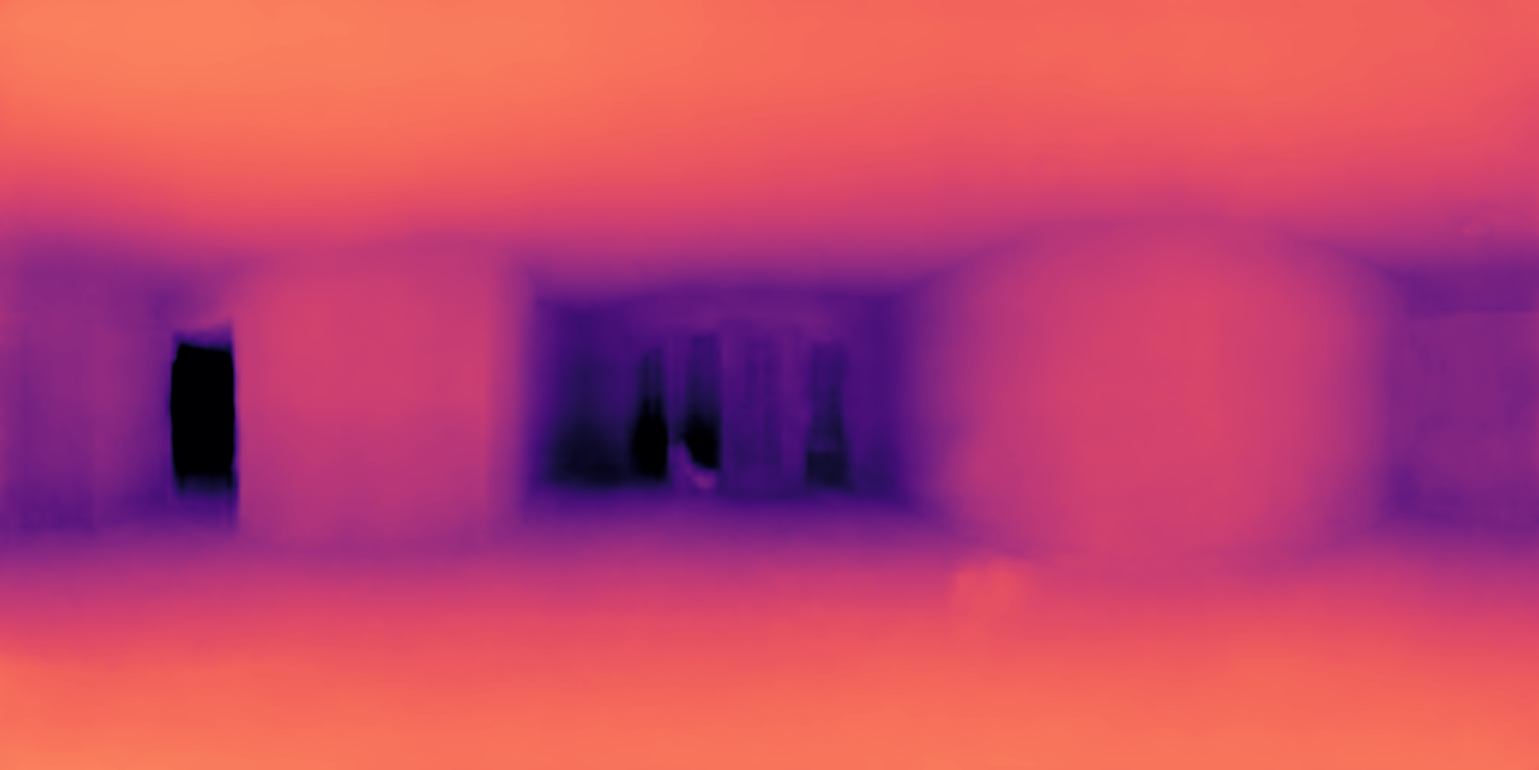}
        \includegraphics[width=0.23\linewidth]{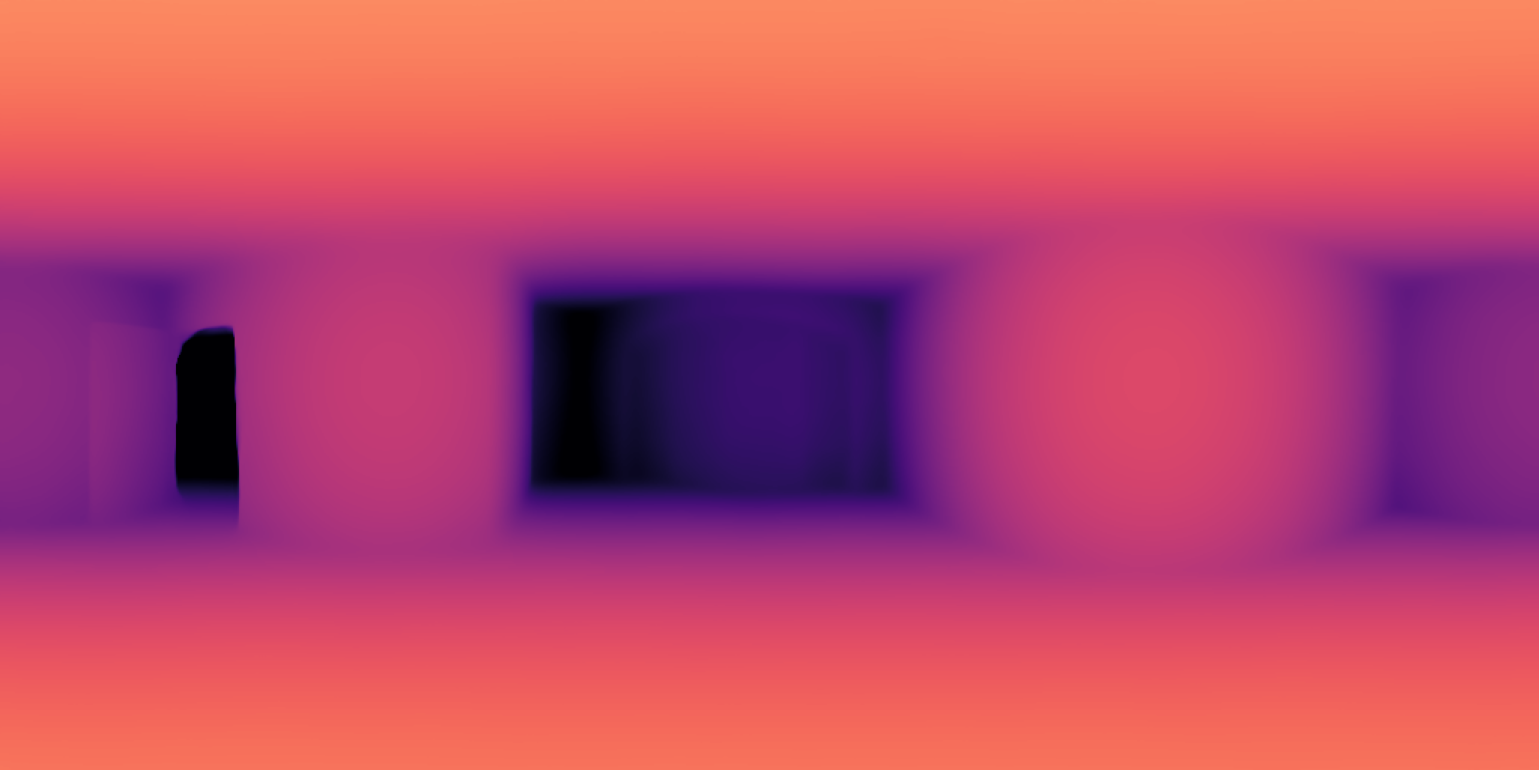}
        \includegraphics[width=0.23\linewidth]{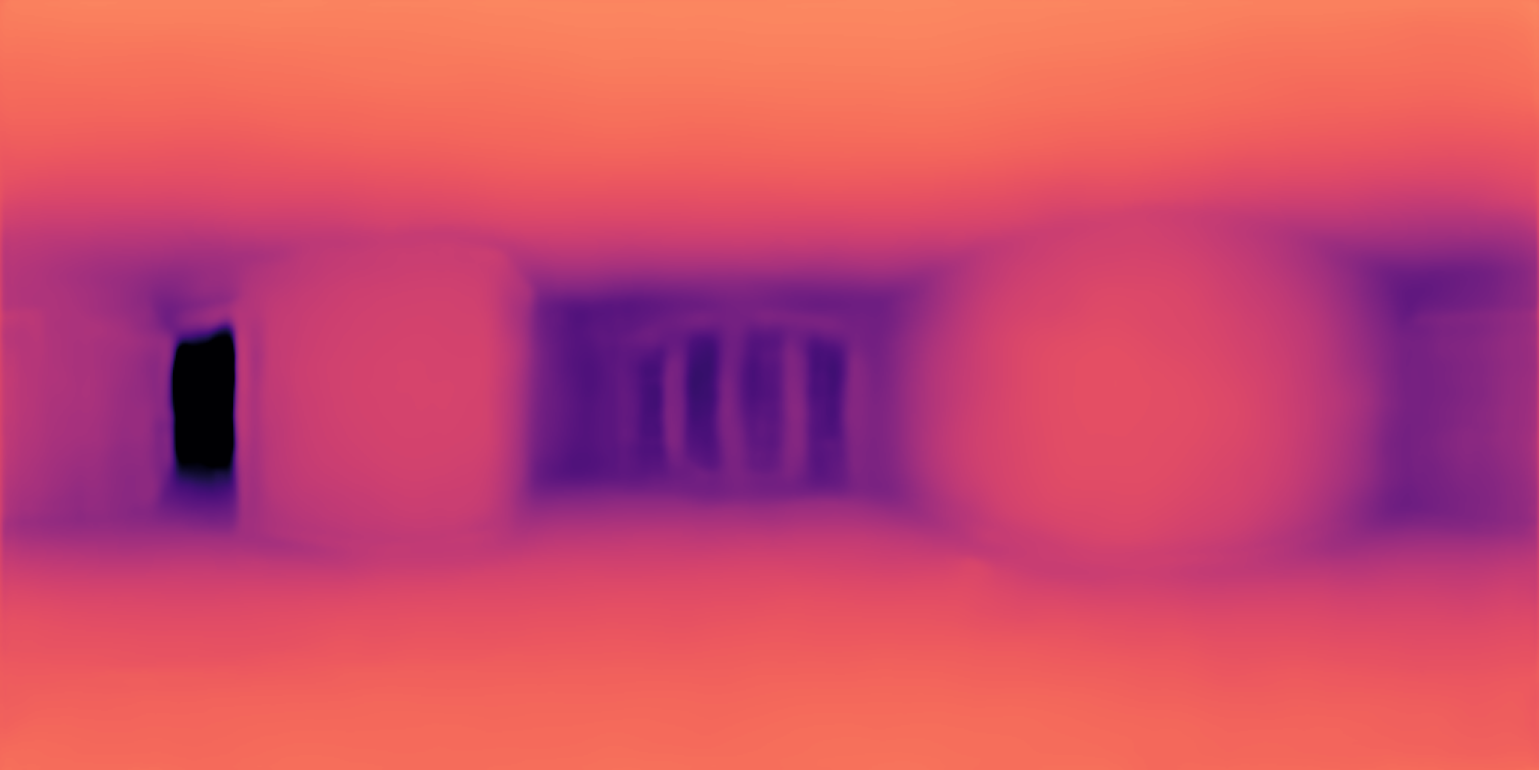}
        \raisebox{2ex}{\includegraphics[width=0.05\linewidth]{figures/qualitative/dac/gv2/meters_m.jpg}}
    \end{minipage}
   \begin{minipage}[t]{\linewidth}
        \centering
        \includegraphics[width=0.23\linewidth]{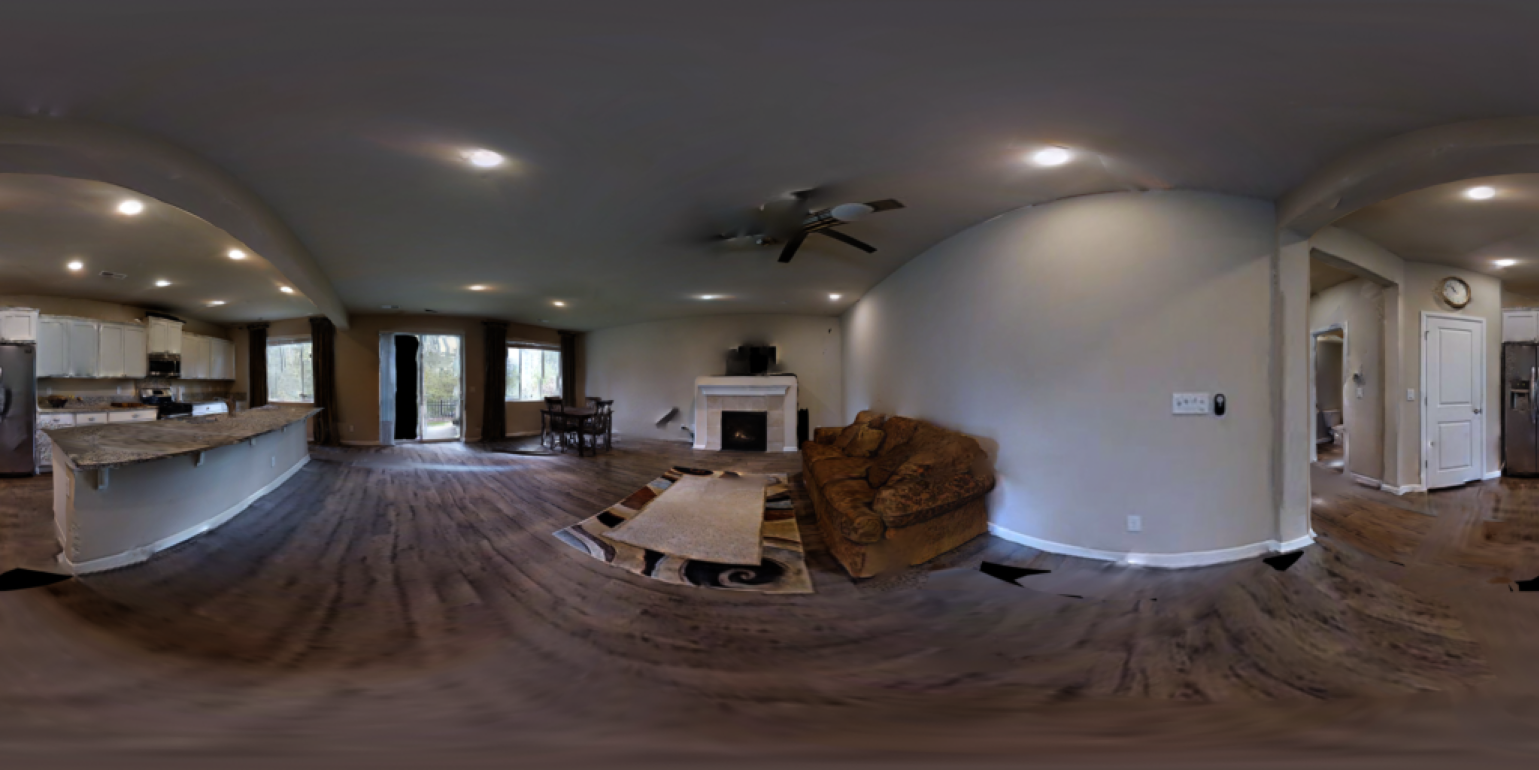}
        \includegraphics[width=0.23\linewidth]{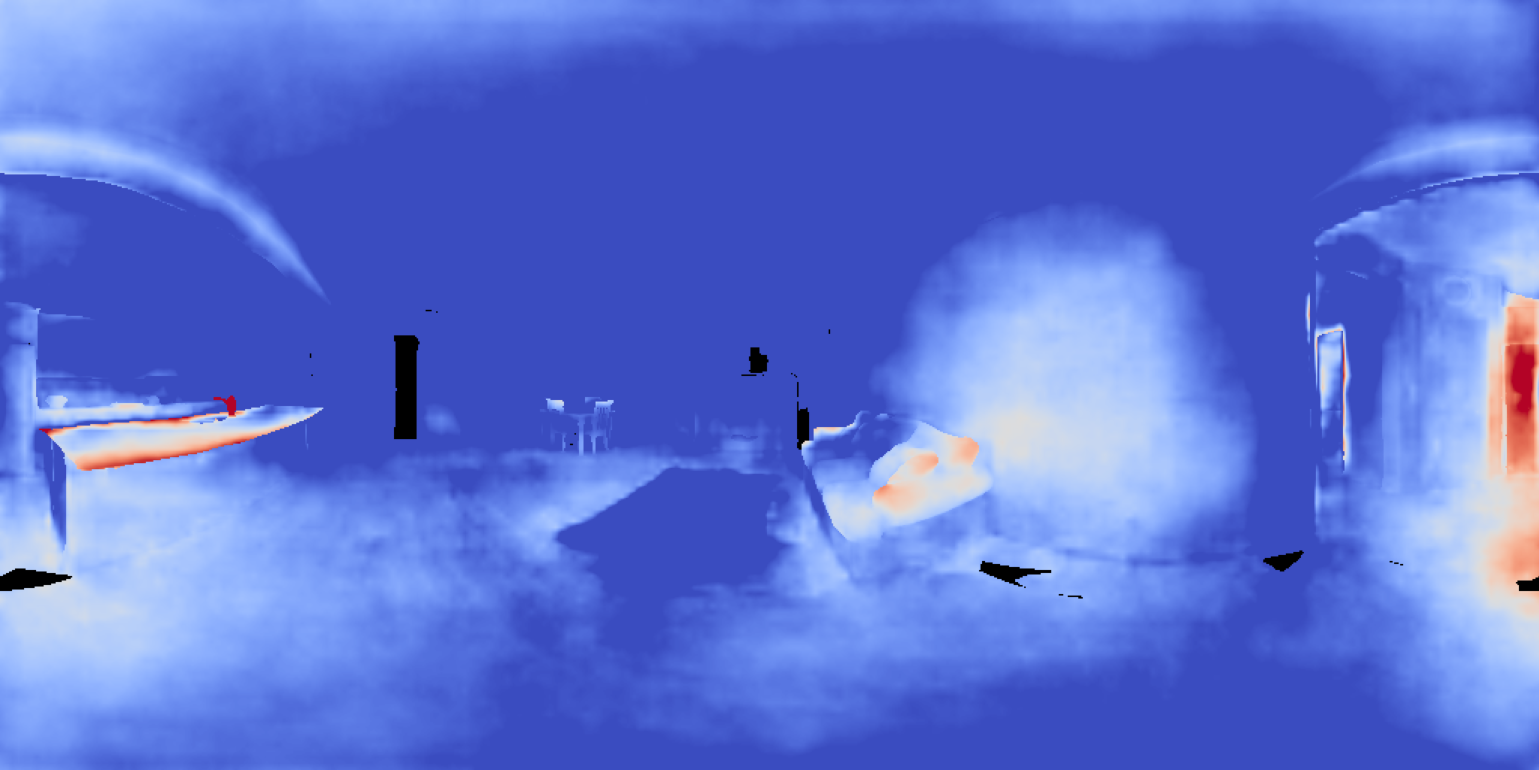}
        \includegraphics[width=0.23\linewidth]{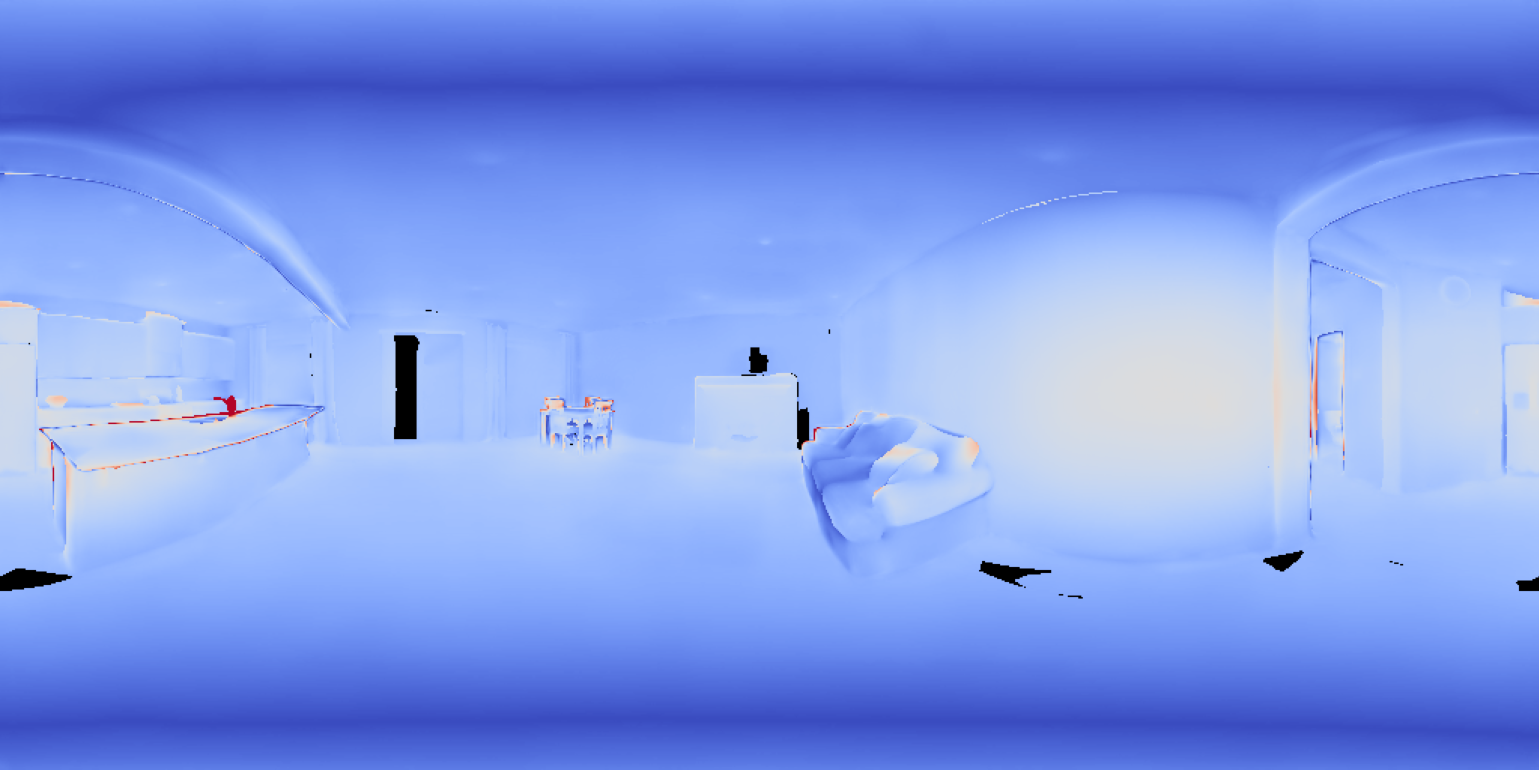}
        \includegraphics[width=0.23\linewidth]{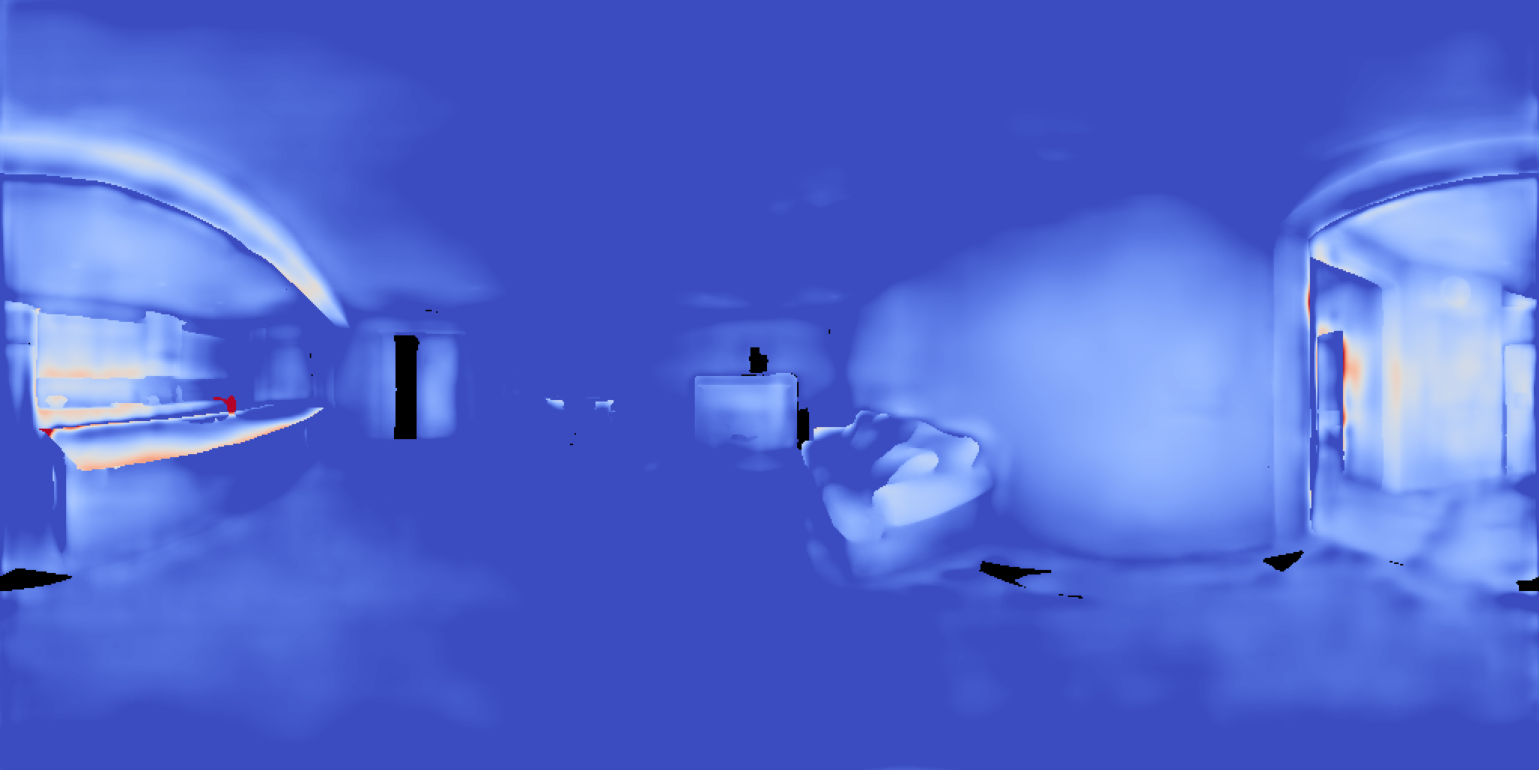}
        \raisebox{2ex}{\includegraphics[width=0.05\linewidth]{figures/qualitative/dac/kitti360/arel.jpg}}
    \end{minipage}
    \begin{minipage}[t]{\linewidth}
        \centering
        \includegraphics[width=0.23\linewidth]{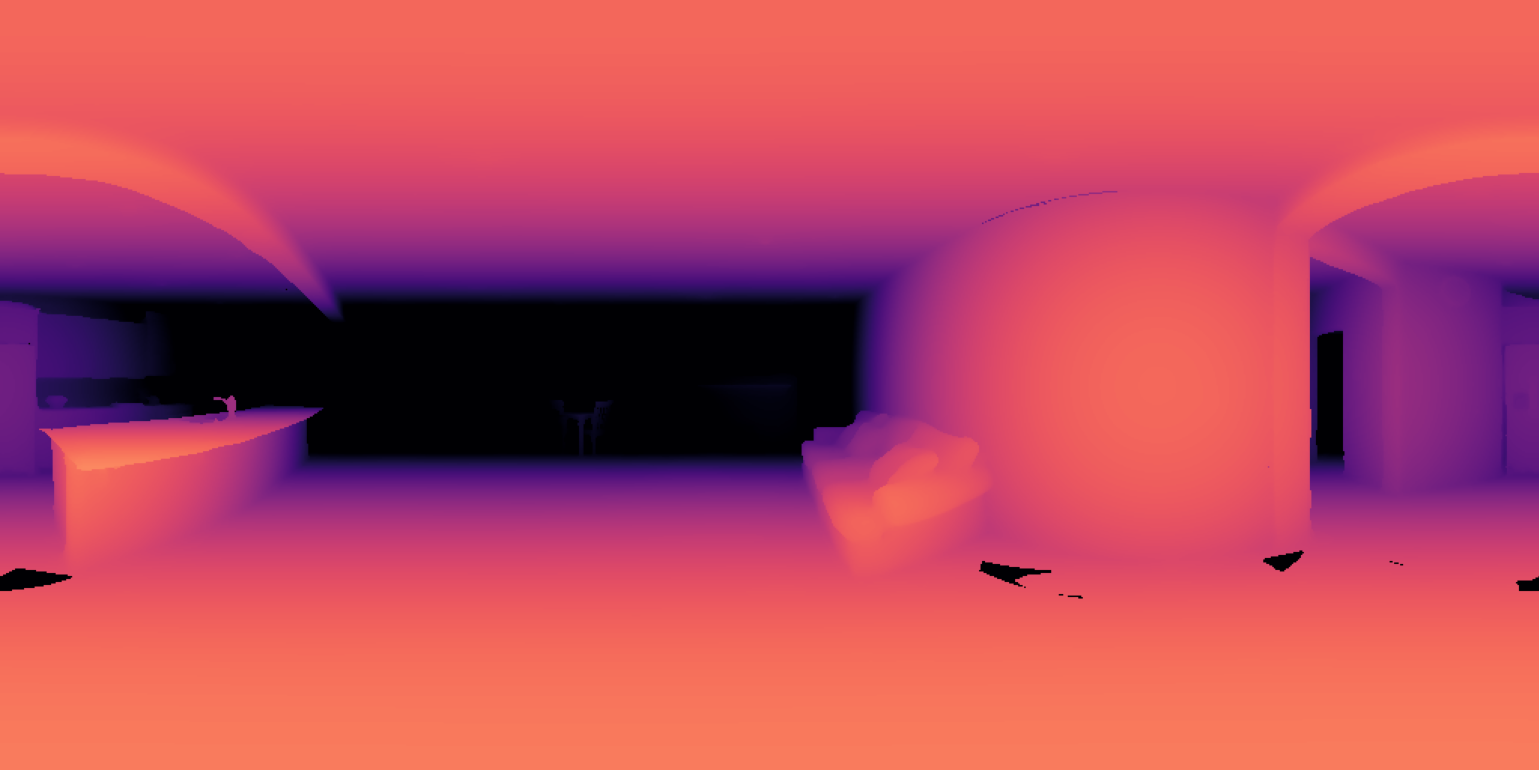}
        \includegraphics[width=0.23\linewidth]{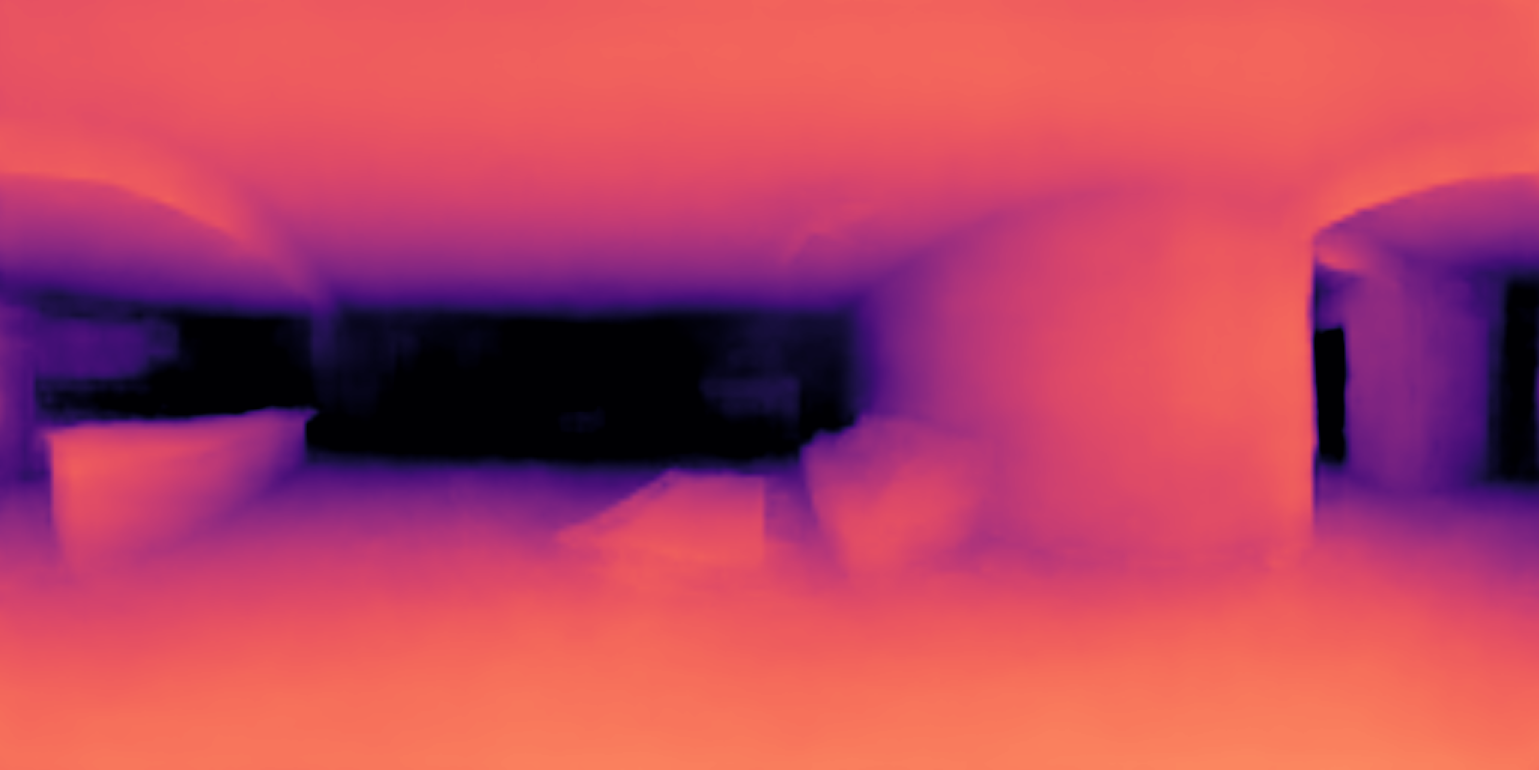}
        \includegraphics[width=0.23\linewidth]{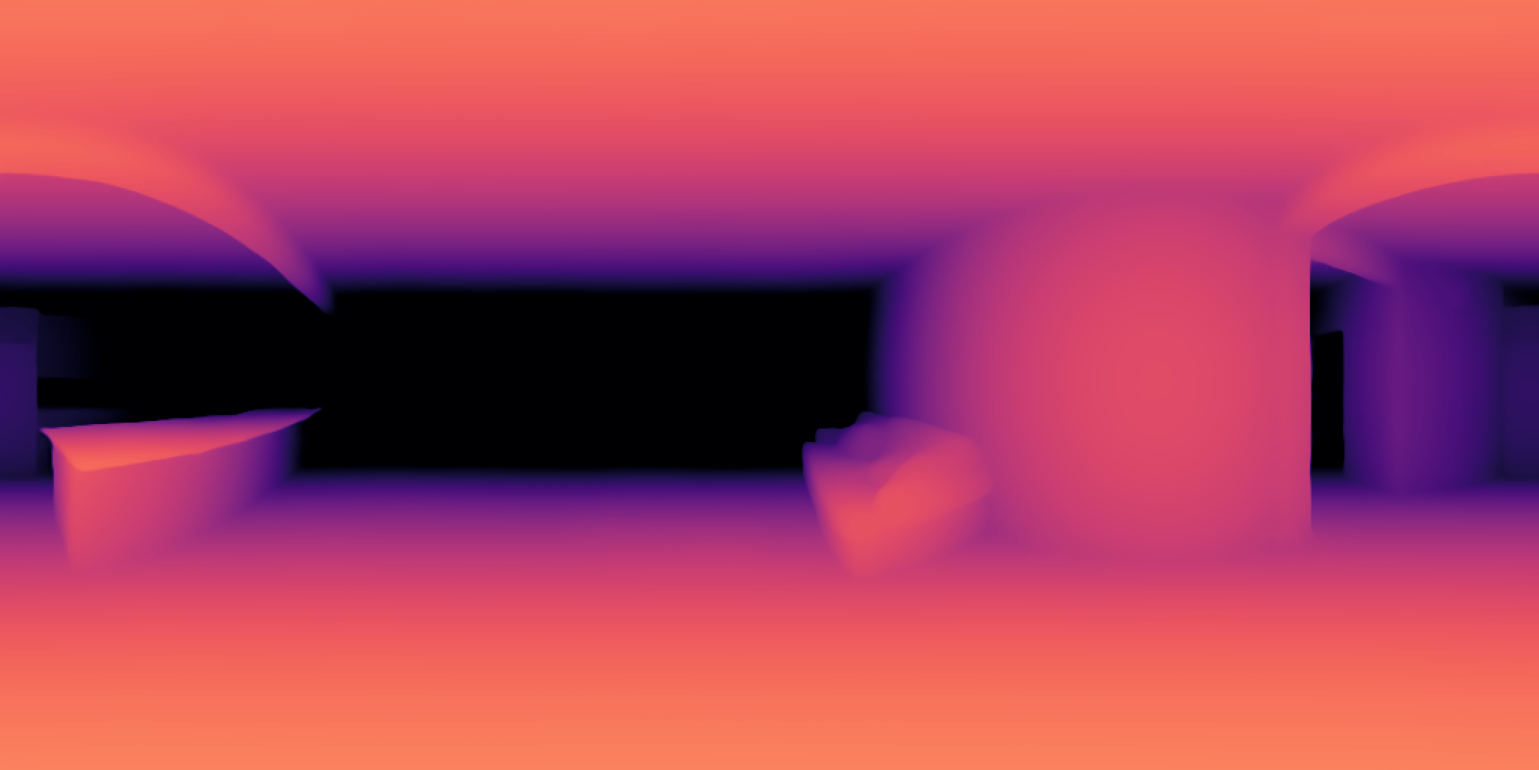}
        \includegraphics[width=0.23\linewidth]{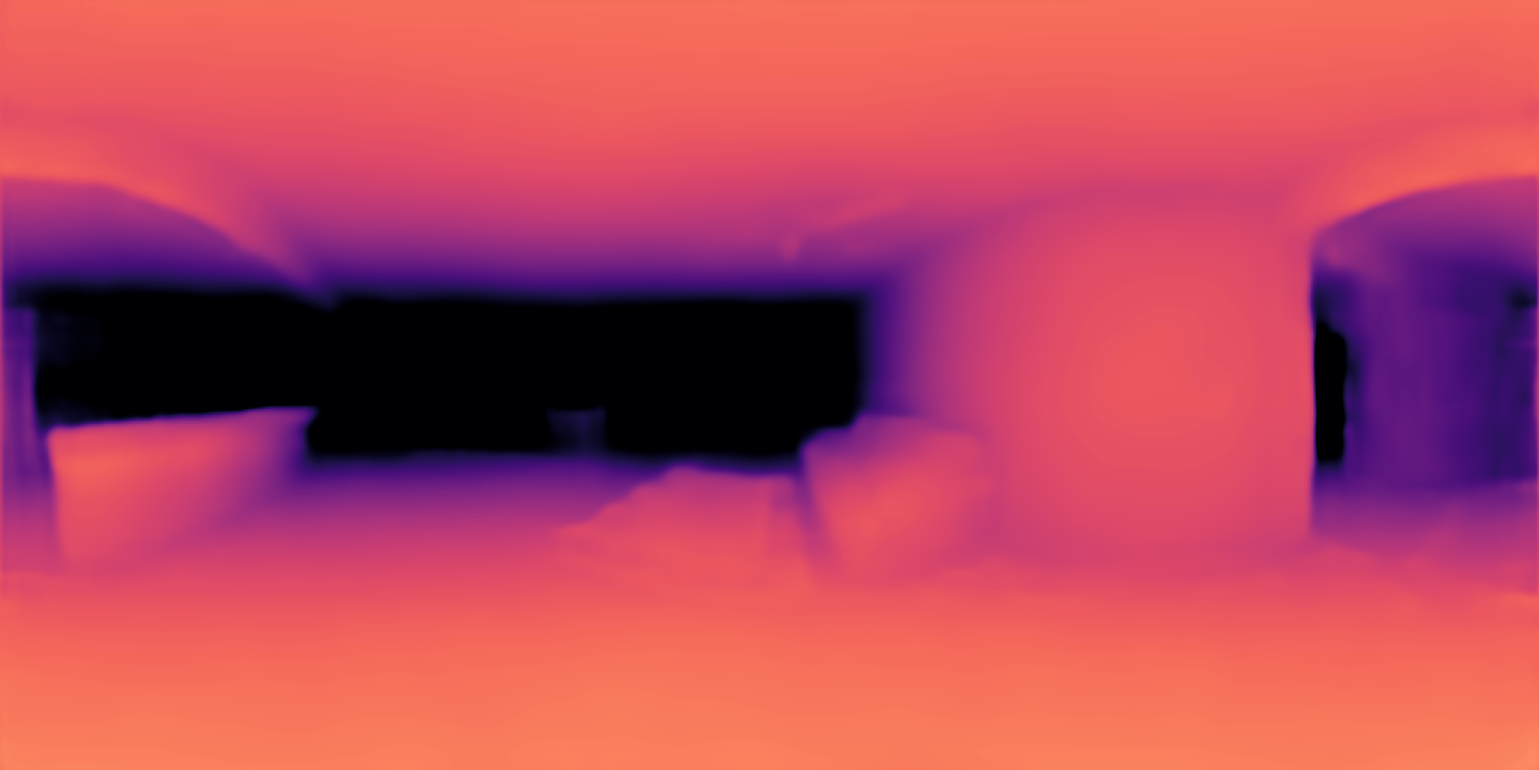}
        \raisebox{2ex}{\includegraphics[width=0.05\linewidth]{figures/qualitative/dac/gv2/meters_m.jpg}}
    \end{minipage}
    \begin{minipage}[t]{\linewidth}
        \centering
        \includegraphics[width=0.23\linewidth]{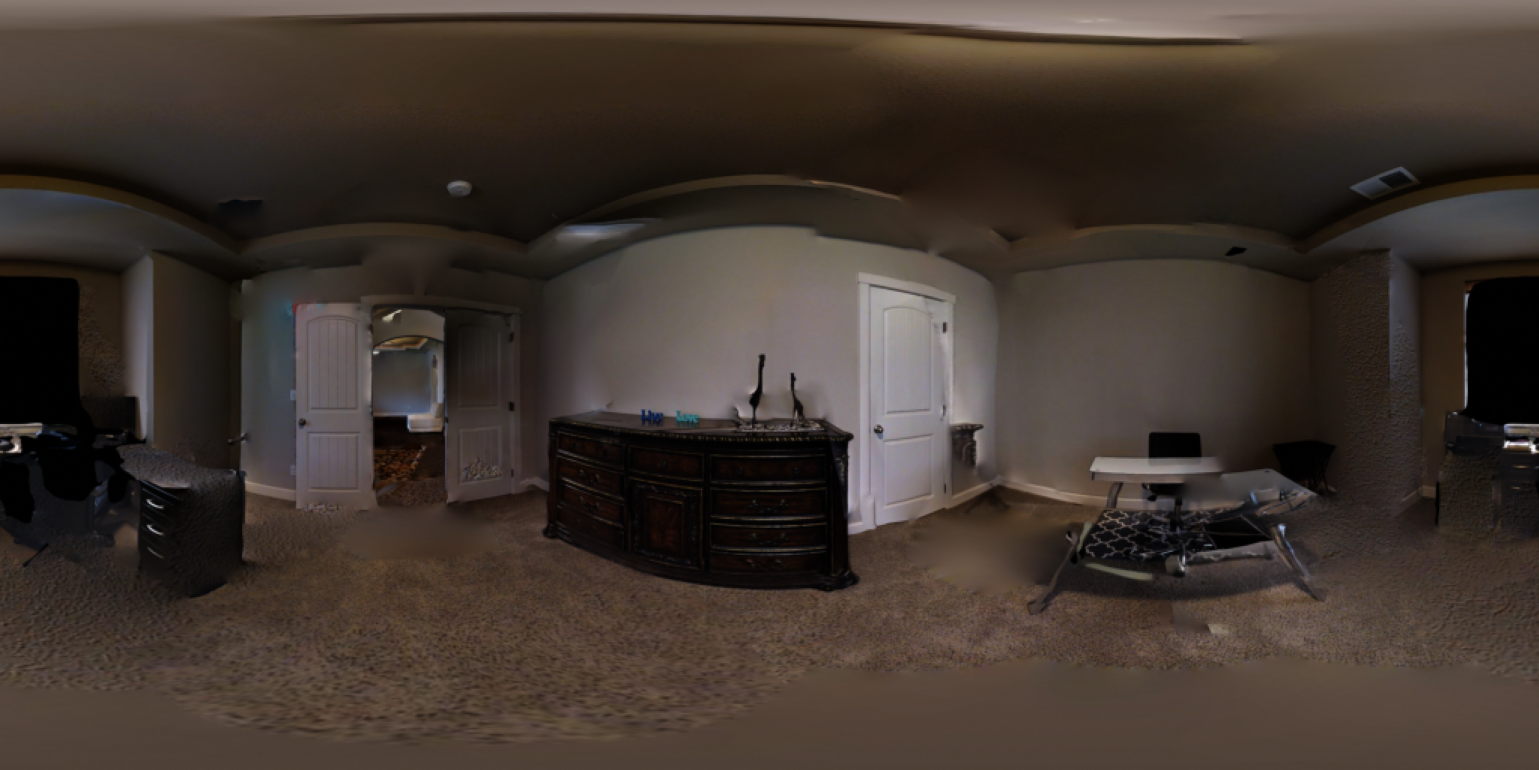}
        \includegraphics[width=0.23\linewidth]{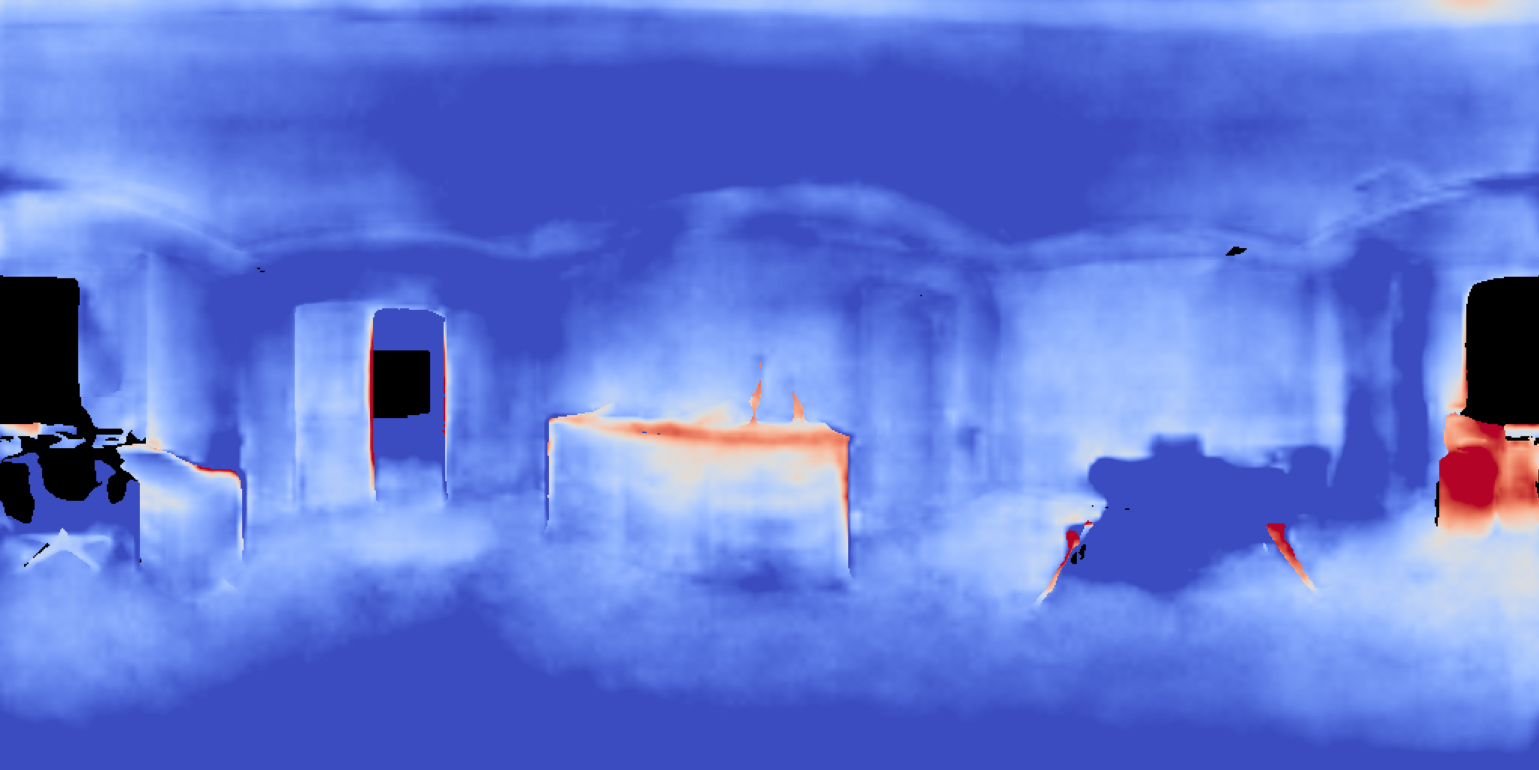}
        \includegraphics[width=0.23\linewidth]{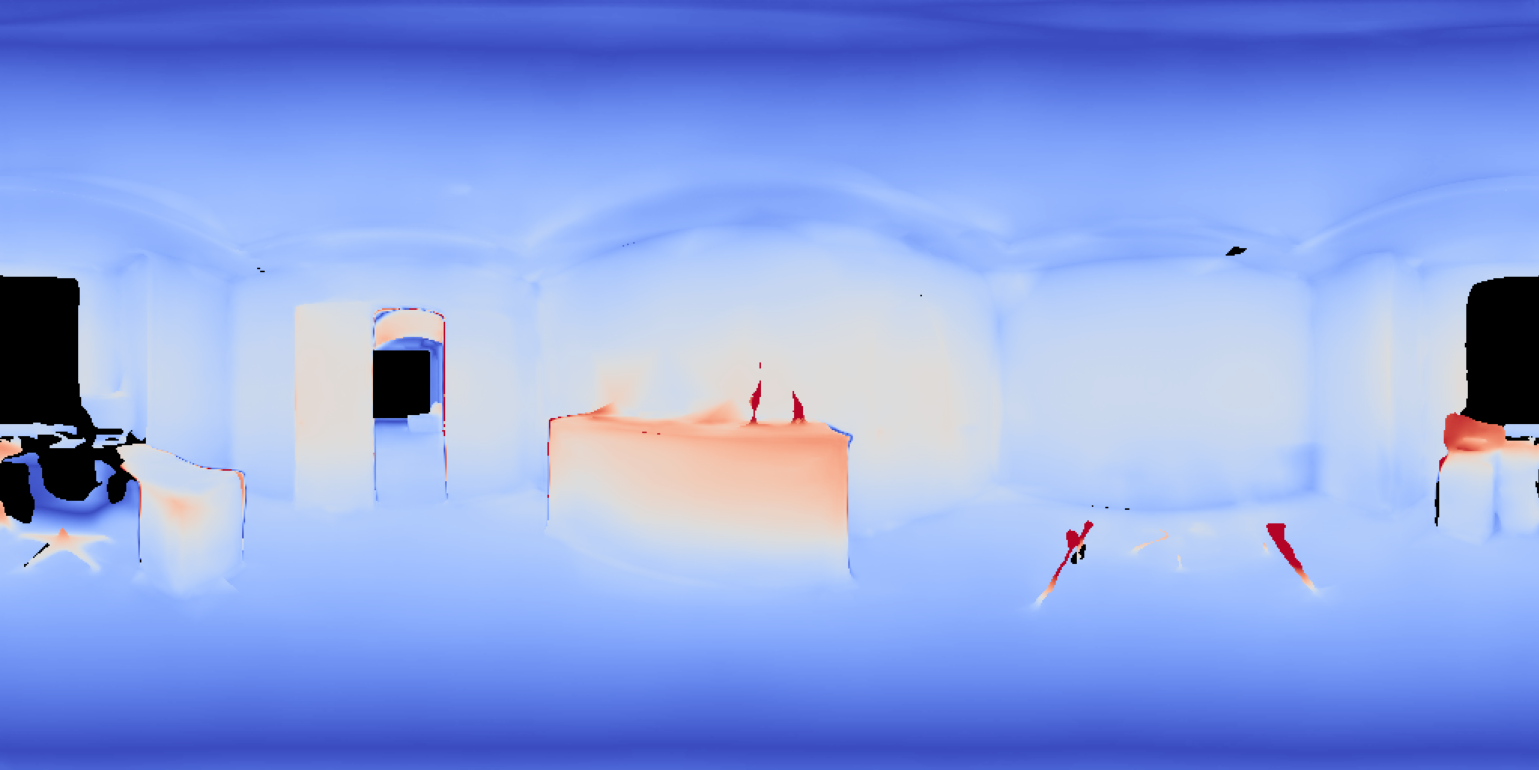}
        \includegraphics[width=0.23\linewidth]{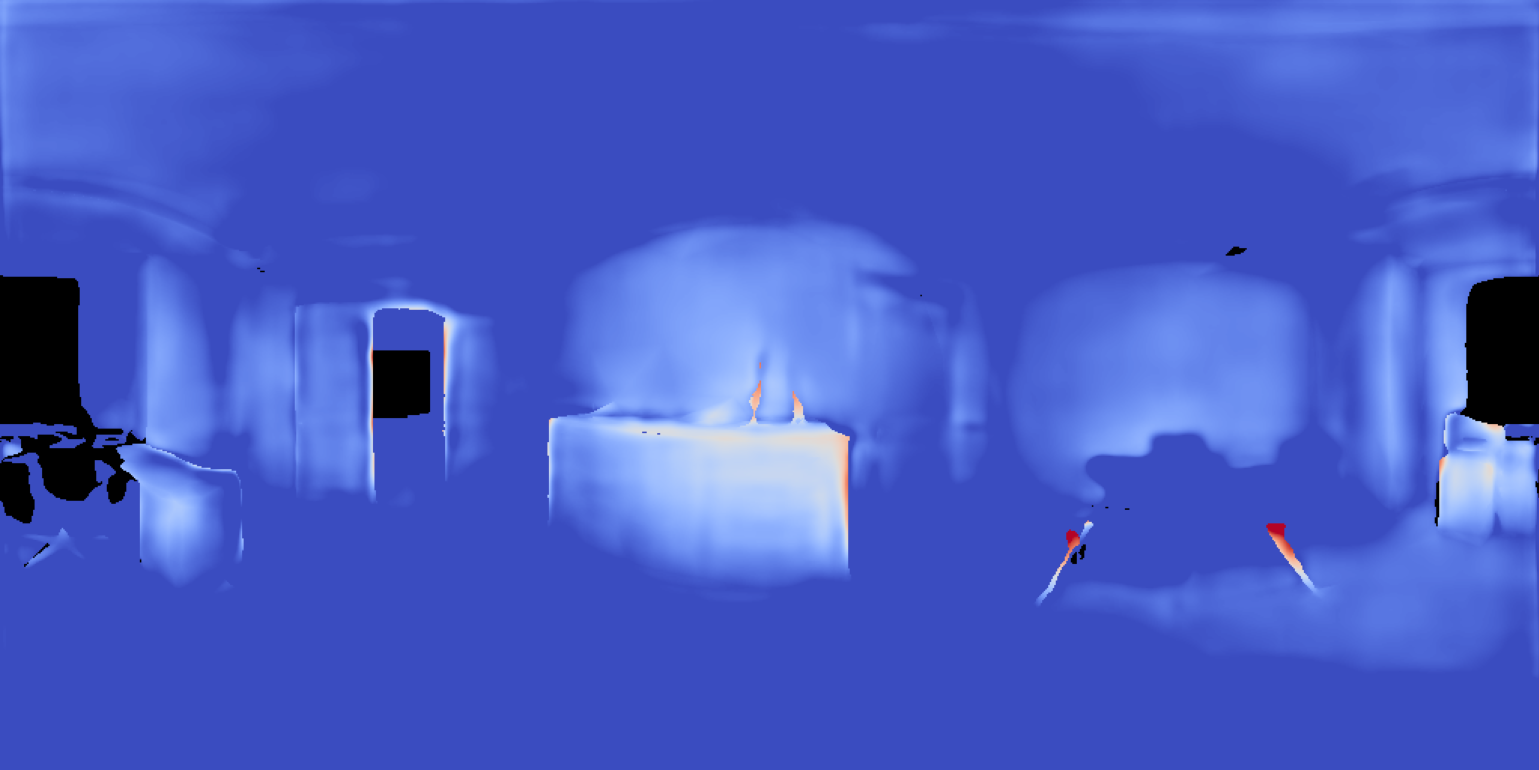}
        \raisebox{2ex}{\includegraphics[width=0.05\linewidth]{figures/qualitative/dac/kitti360/arel.jpg}}
    \end{minipage}
    \begin{minipage}[t]{\linewidth}
        \centering
        \includegraphics[width=0.23\linewidth]{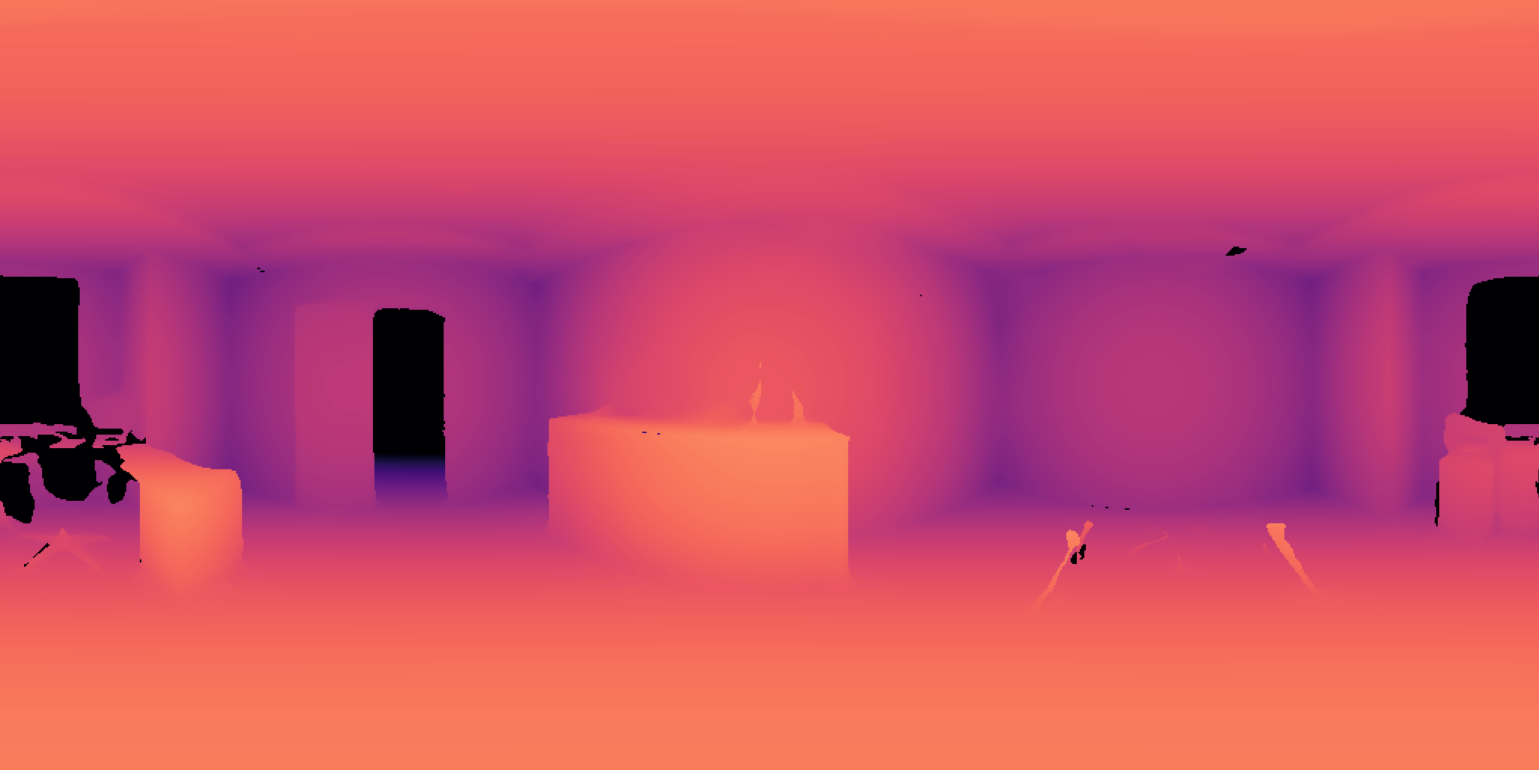}
        \includegraphics[width=0.23\linewidth]{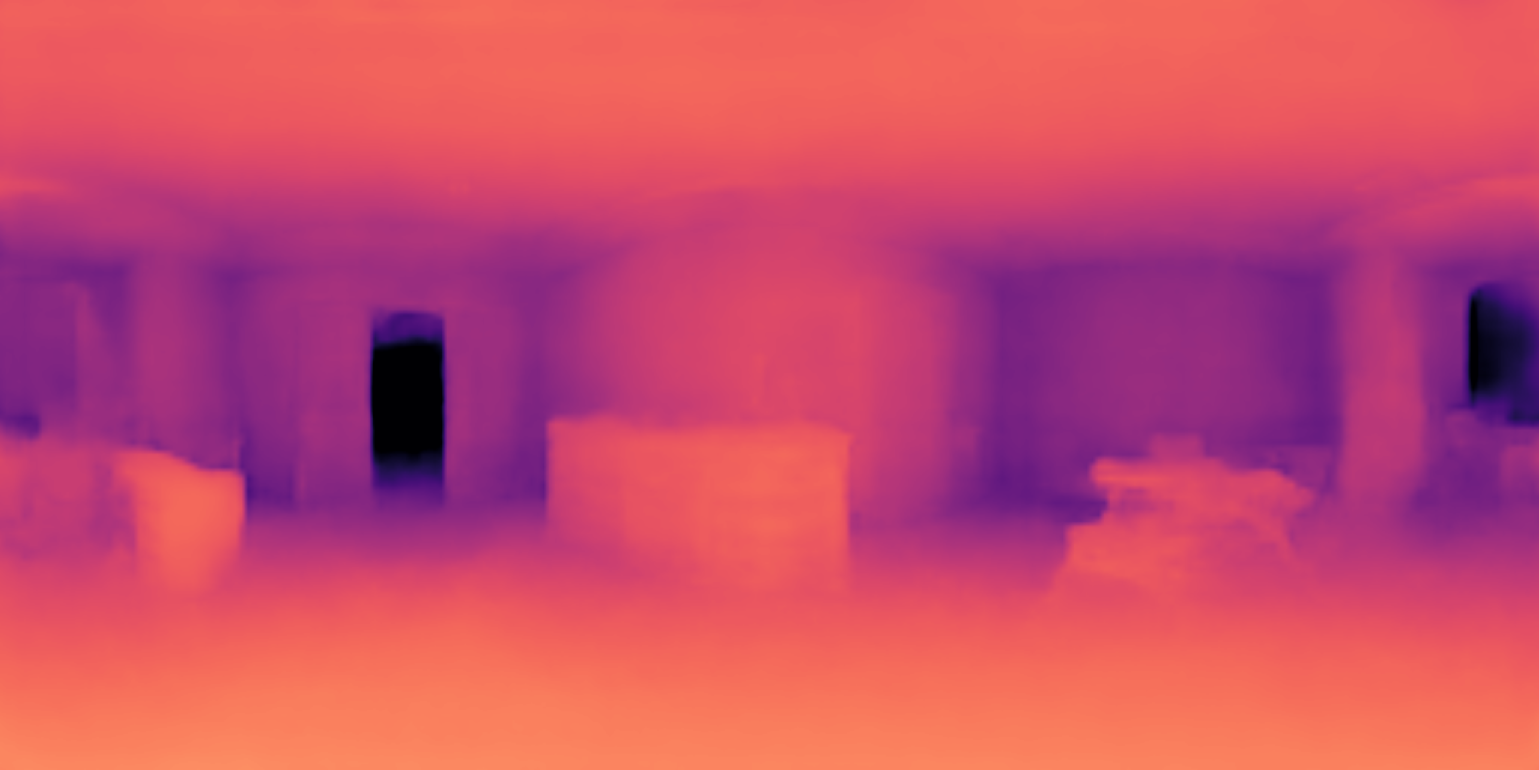}
        \includegraphics[width=0.23\linewidth]{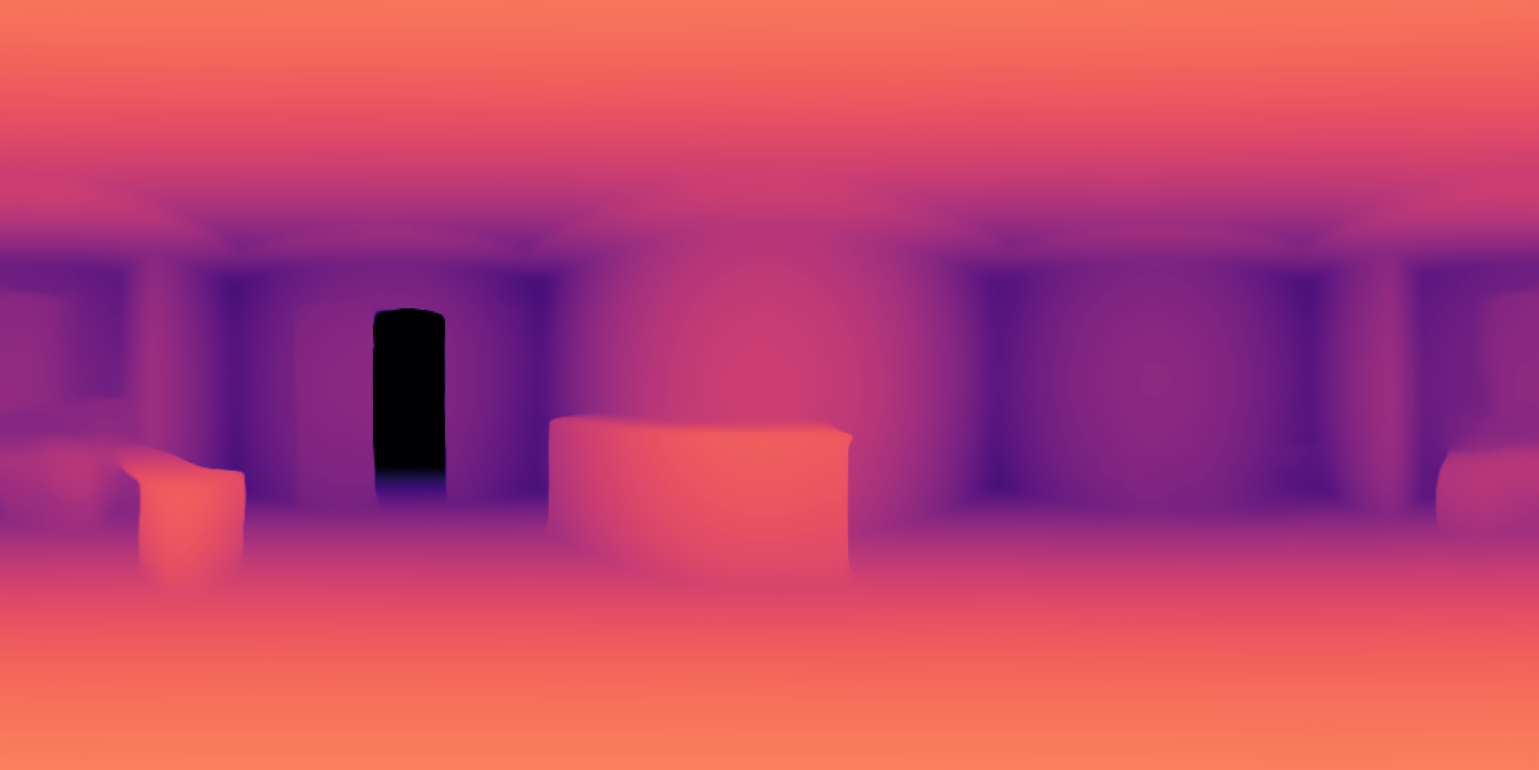}
        \includegraphics[width=0.23\linewidth]{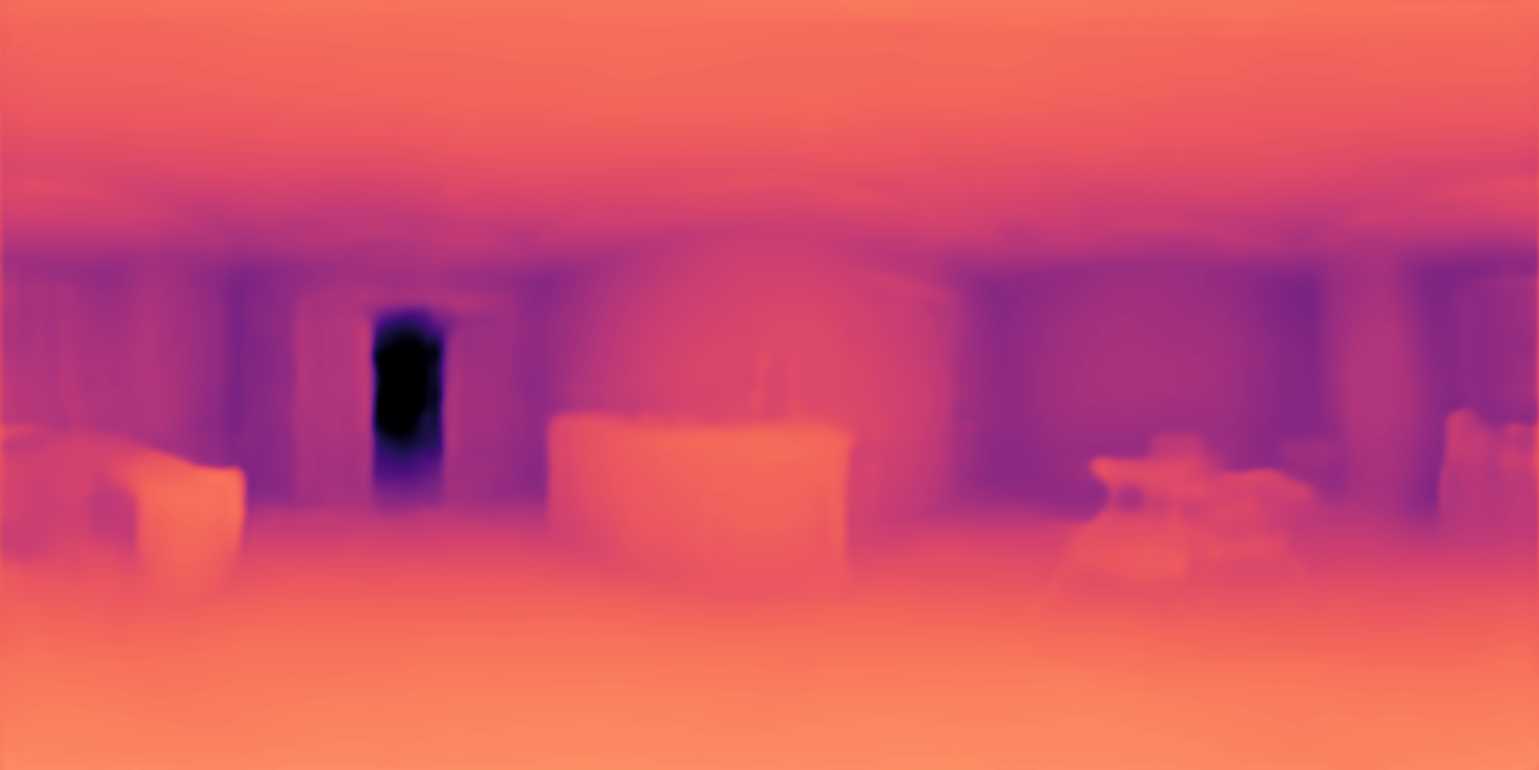}
        \raisebox{2ex}{\includegraphics[width=0.05\linewidth]{figures/qualitative/dac/gv2/meters_m.jpg}}
    \end{minipage}
    \begin{minipage}[t]{\linewidth}
        \centering
        \includegraphics[width=0.23\linewidth]{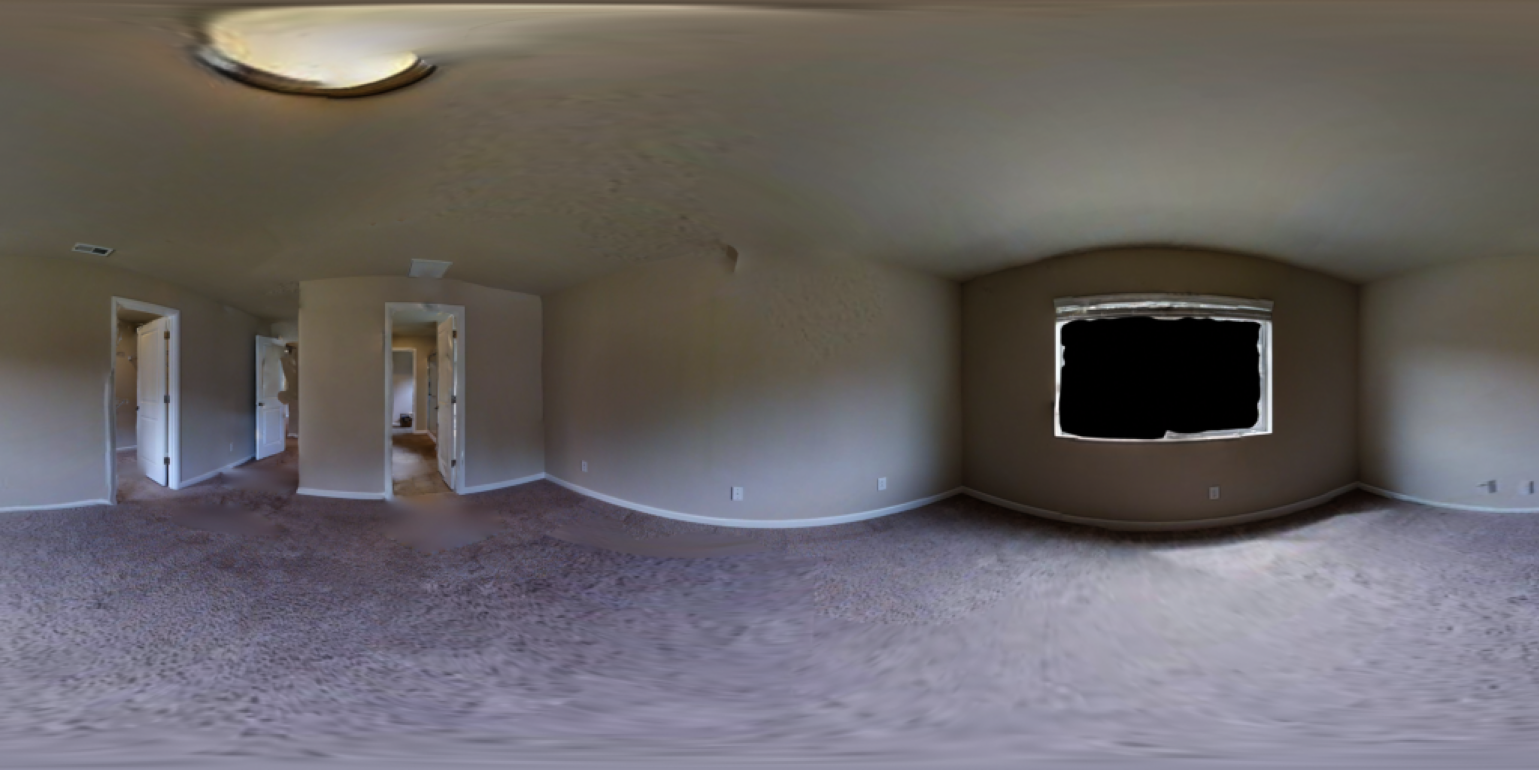}
        \includegraphics[width=0.23\linewidth]{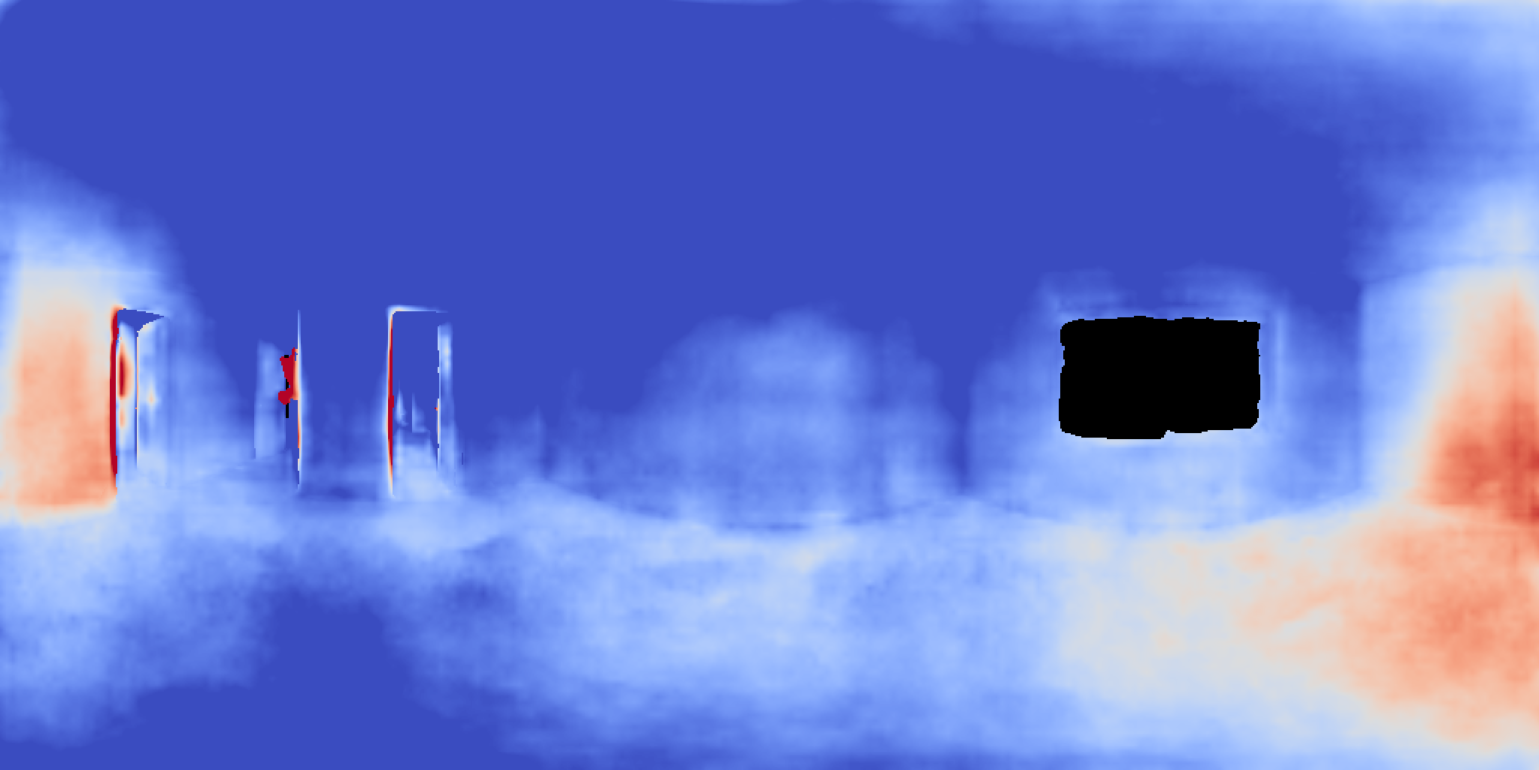}
        \includegraphics[width=0.23\linewidth]{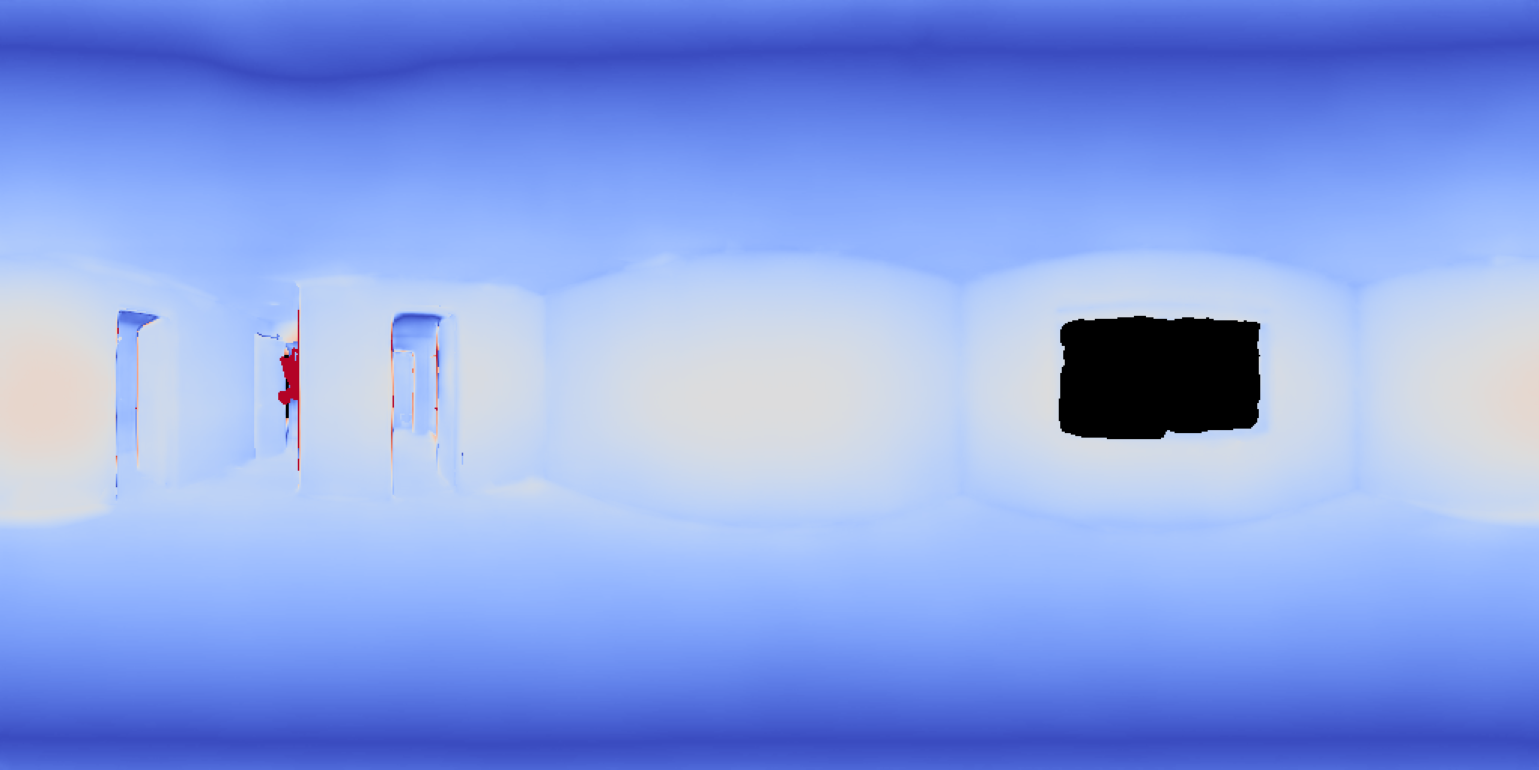}
        \includegraphics[width=0.23\linewidth]{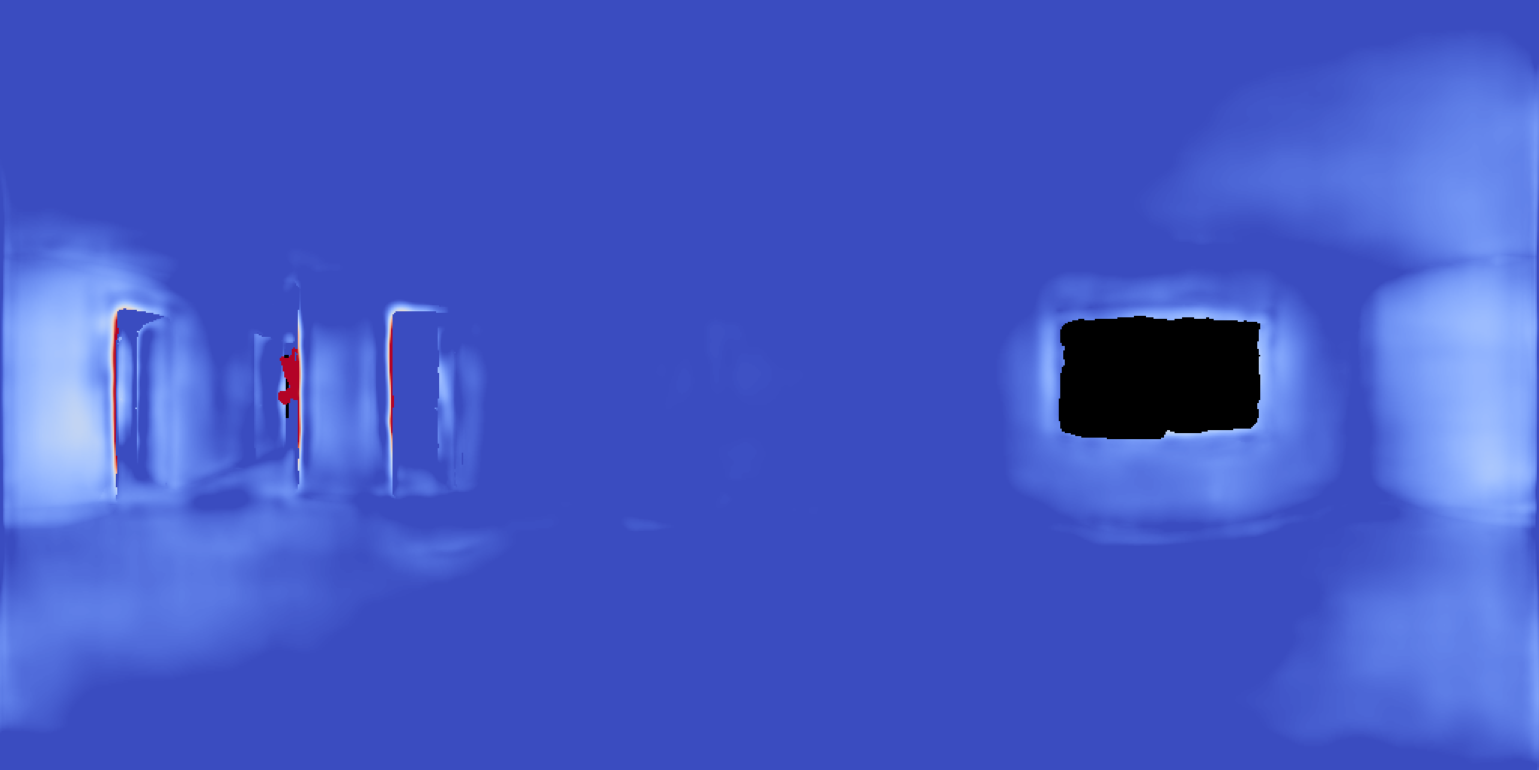}
        \raisebox{2ex}{\includegraphics[width=0.05\linewidth]{figures/qualitative/dac/kitti360/arel.jpg}}
    \end{minipage}
    \begin{minipage}[t]{\linewidth}
        \centering
        \begin{tikzpicture}
            \draw (0,0) node[inner sep=0] {\includegraphics[width=0.23\linewidth]{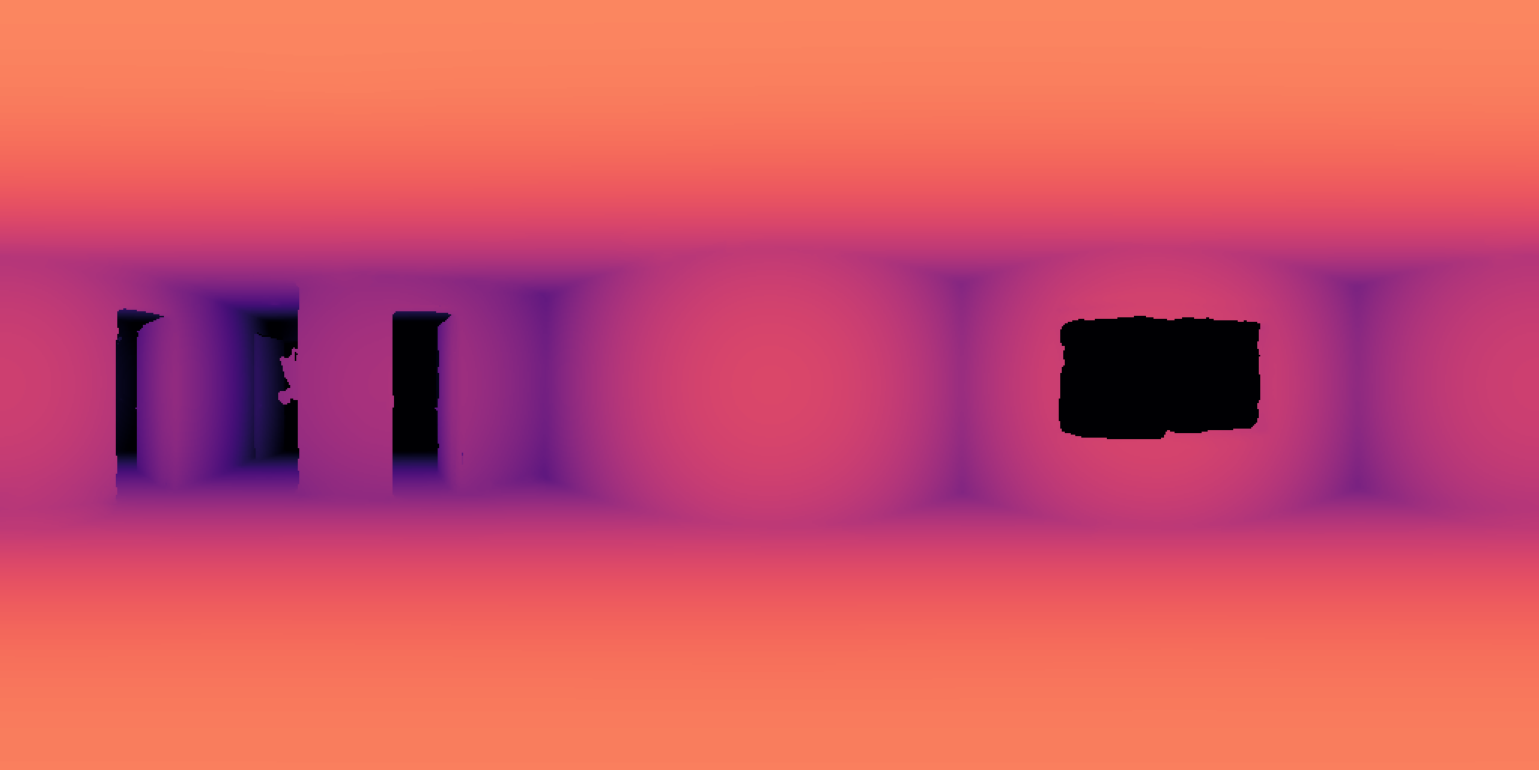}};
            \draw (0,-1.1-0.5) node[inner sep=0, align=center] {\fontsize{8.0}{10}\selectfont {RGB \& GT}};
        \end{tikzpicture}
        \begin{tikzpicture}
            \draw (0,0) node[inner sep=0] {\includegraphics[width=0.23\linewidth]{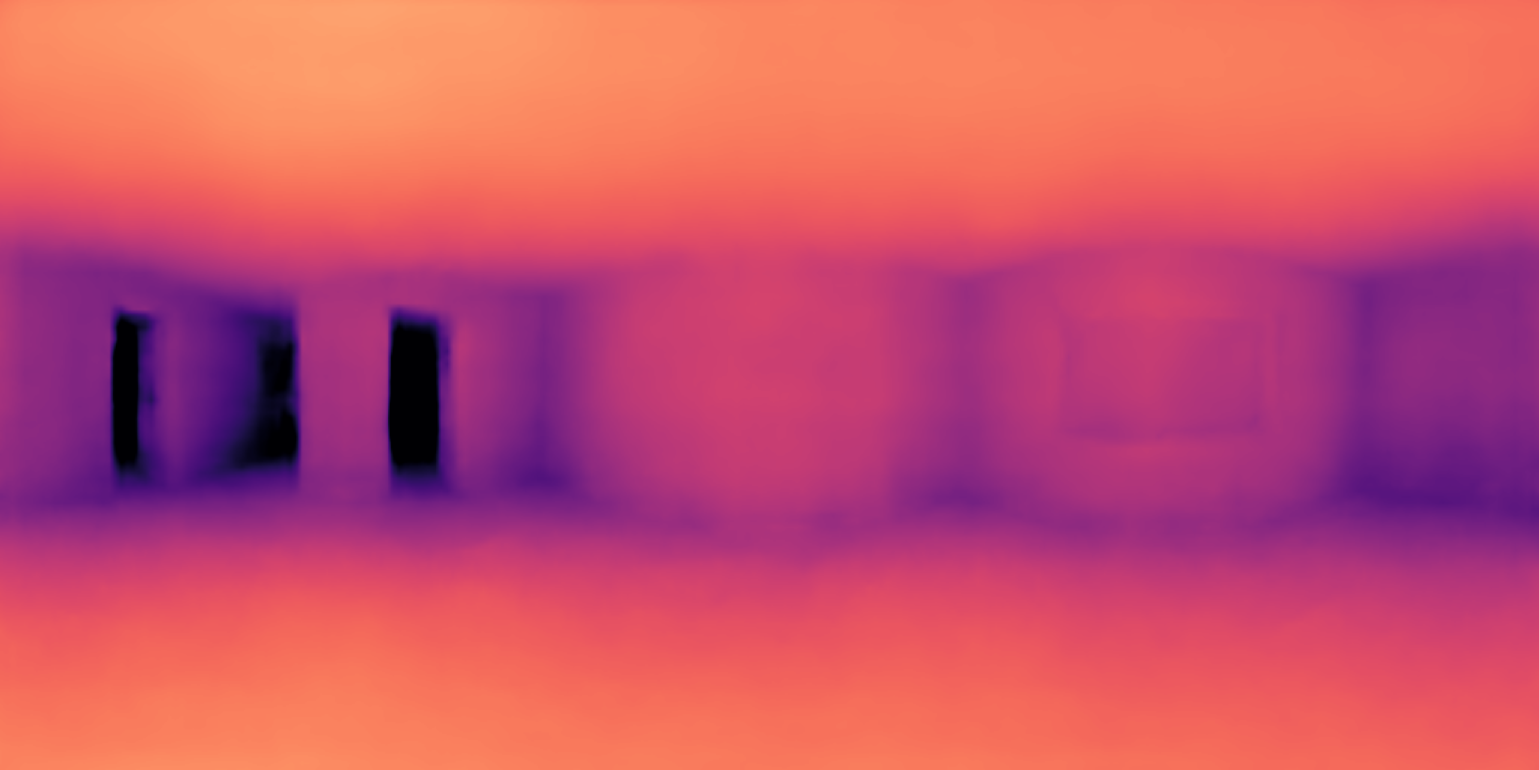}};
            \draw (0,-1.08-0.5) node[inner sep=0, align=center] {\fontsize{8.0}{10}\selectfont {\dacUni~\cite{guo2025depth}}};
        \end{tikzpicture}
        \begin{tikzpicture}
            \draw (0,0) node[inner sep=0] {\includegraphics[width=0.23\linewidth]{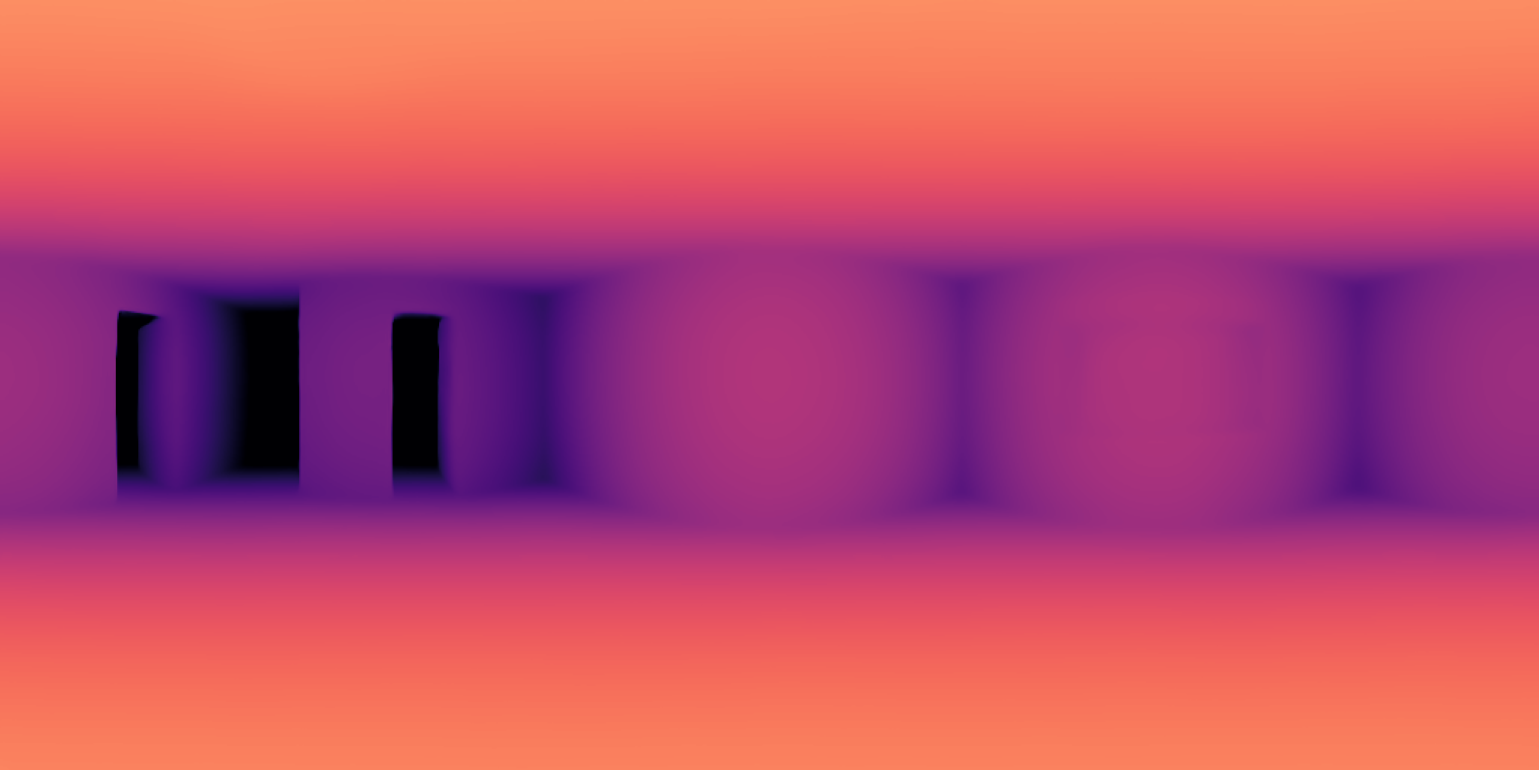}};
            \draw (0,-1.08-0.5) node[inner sep=0, align=center] {\fontsize{8.0}{10}\selectfont {\uniKThreeD~\cite{piccinelli2025unik3d}}};
        \end{tikzpicture}
        \begin{tikzpicture}
            \draw (0,0) node[inner sep=0, align=center] {\includegraphics[width=0.23\linewidth]{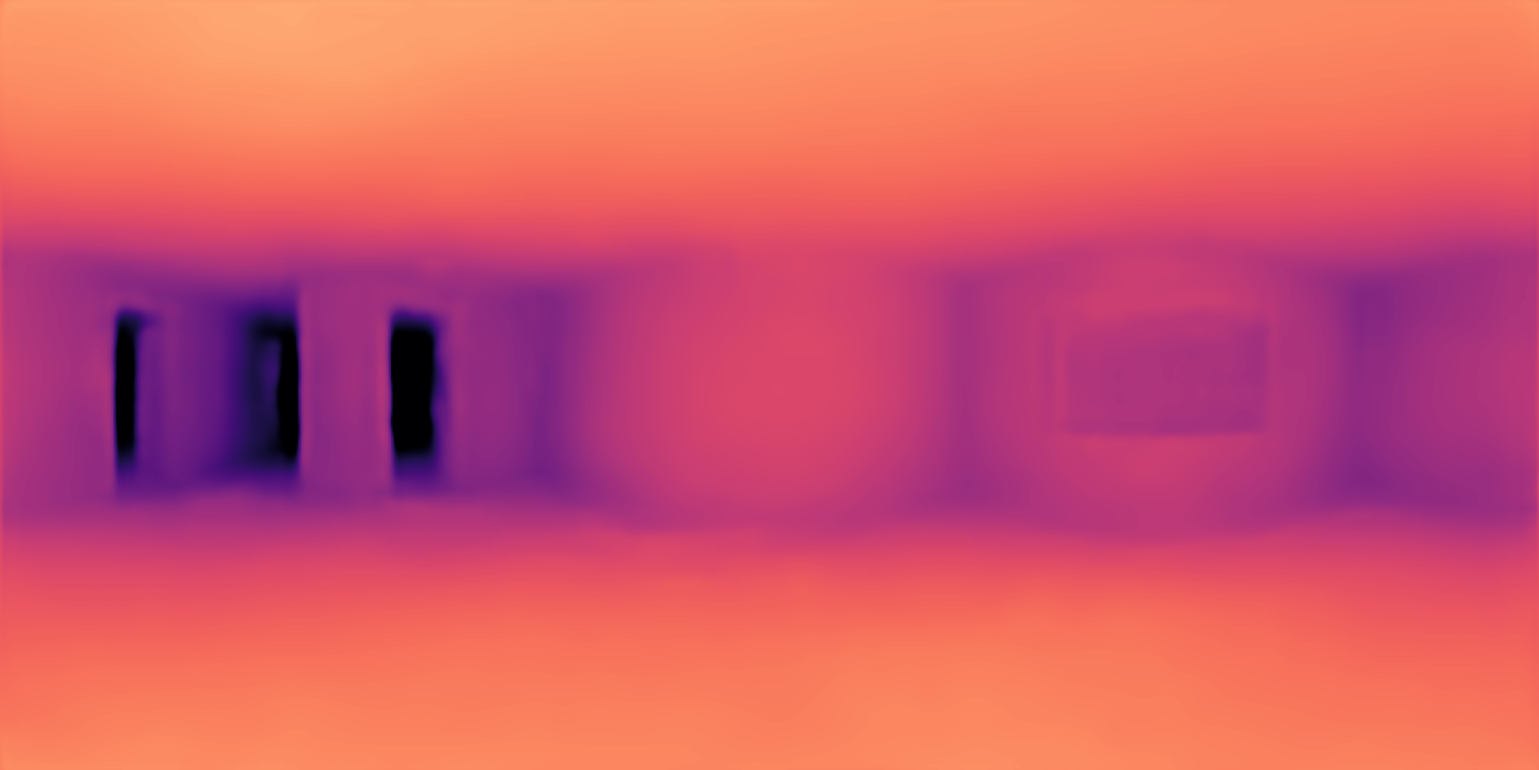}};
            \draw (0,-1.1-0.5) node[inner sep=0, align=center] {\fontsize{8.0}{10}\selectfont {\methodName}};
        \end{tikzpicture}
        \raisebox{7ex}{\includegraphics[width=0.05\linewidth]{figures/qualitative/dac/gv2/meters_m.jpg}}
    \end{minipage}
    \caption{\textbf{Qualitative Results on \panoGVTwo~\cite{albanis2021pano3d}.} Every pair of consecutive rows corresponds to a single sample. Odd rows display the input RGB image, and \absRel error between predicted and GT depth maps. Even rows display the GT depth map and predicted depth maps.}
    \label{fig:supp_gv2}
\end{figure*}

\begin{figure*}
    \centering
    \begin{minipage}[t]{\linewidth}
        \centering
        \includegraphics[width=0.23\linewidth]{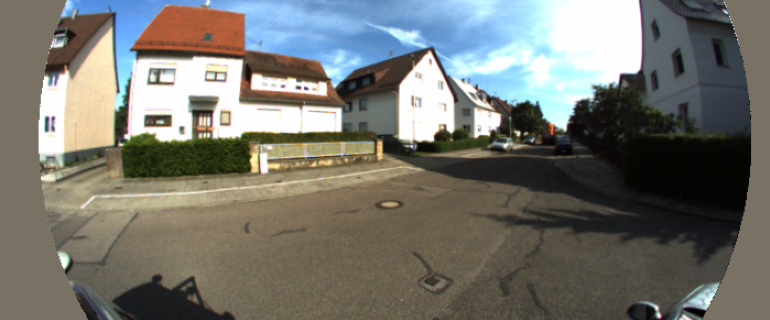}
        \includegraphics[width=0.23\linewidth]{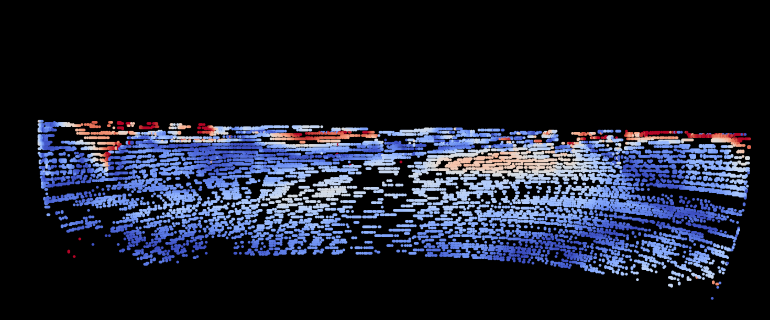}
        \includegraphics[width=0.23\linewidth]{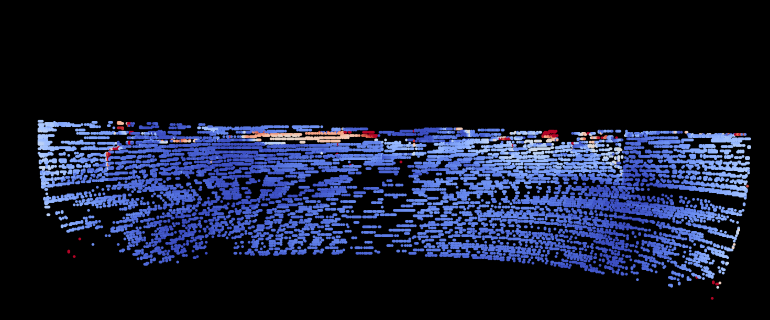}
        \includegraphics[width=0.23\linewidth]{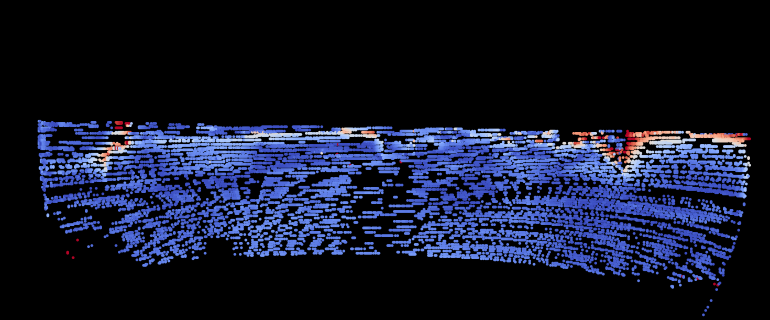}
        \raisebox{1.2ex}{\includegraphics[width=0.05\linewidth]{figures/qualitative/dac/kitti360/arel.jpg}}
    \end{minipage}
    \begin{minipage}[t]{\linewidth}
        \centering
        \includegraphics[width=0.23\linewidth]{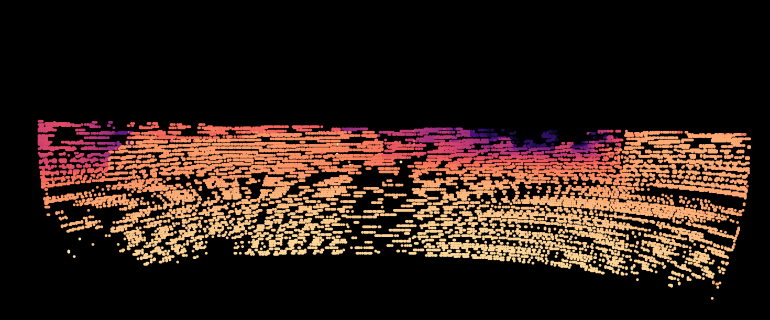}
        \includegraphics[width=0.23\linewidth]{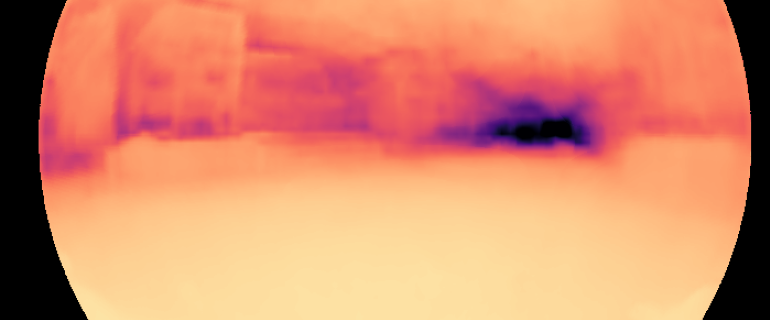}
        \includegraphics[width=0.23\linewidth]{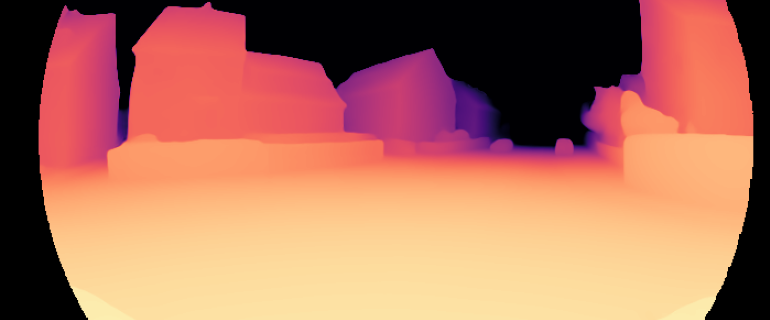}
        \includegraphics[width=0.23\linewidth]{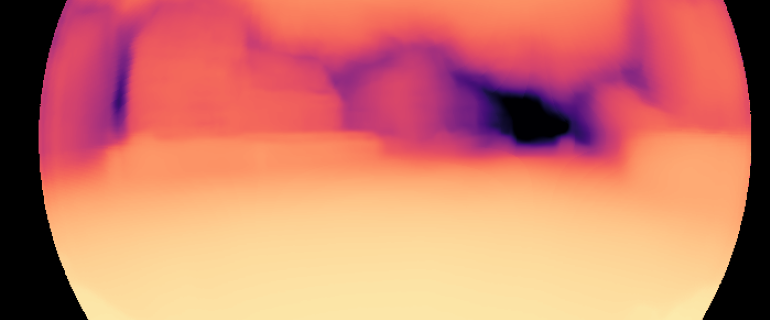}
        \raisebox{1.2ex}{\includegraphics[width=0.05\linewidth]{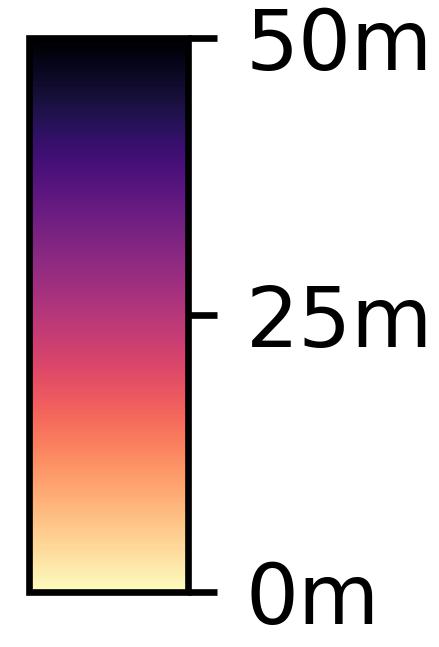}}
    \end{minipage}
   \begin{minipage}[t]{\linewidth}
        \centering
        \includegraphics[width=0.23\linewidth]{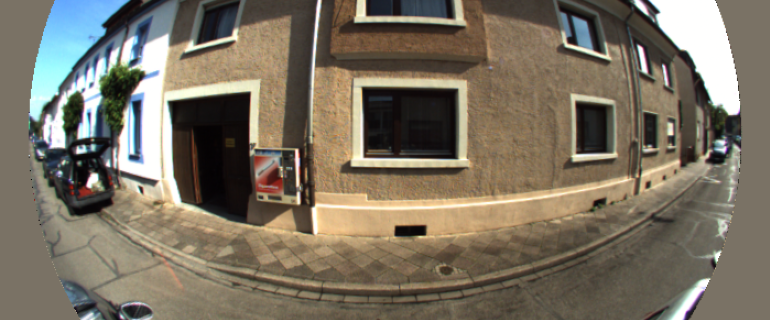}
        \includegraphics[width=0.23\linewidth]{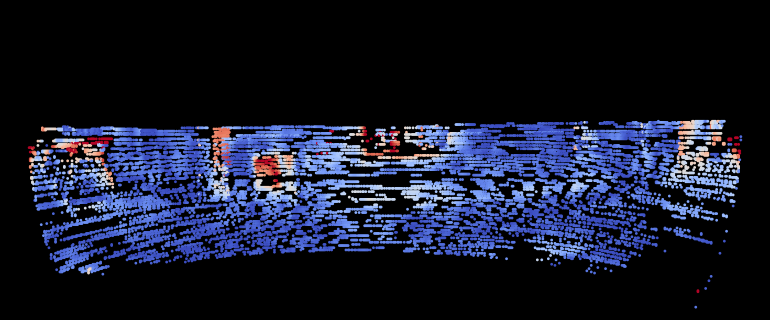}
        \includegraphics[width=0.23\linewidth]{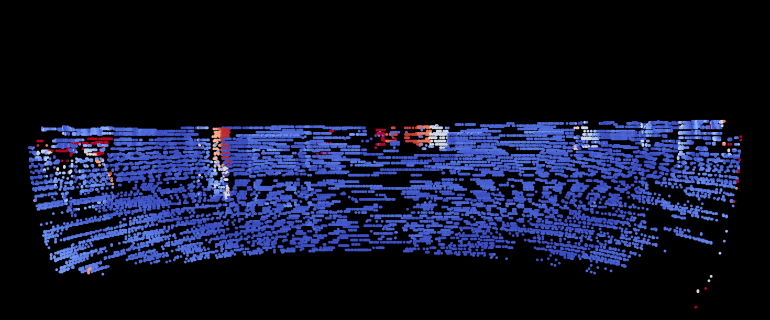}
        \includegraphics[width=0.23\linewidth]{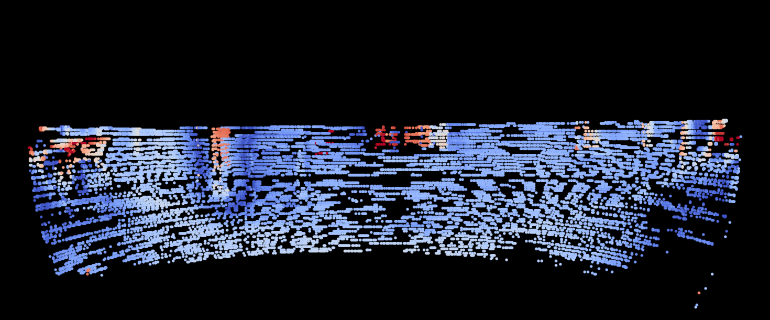}
        \raisebox{1.2ex}{\includegraphics[width=0.05\linewidth]{figures/qualitative/dac/kitti360/arel.jpg}}
    \end{minipage}
    \begin{minipage}[t]{\linewidth}
        \centering
        \includegraphics[width=0.23\linewidth]{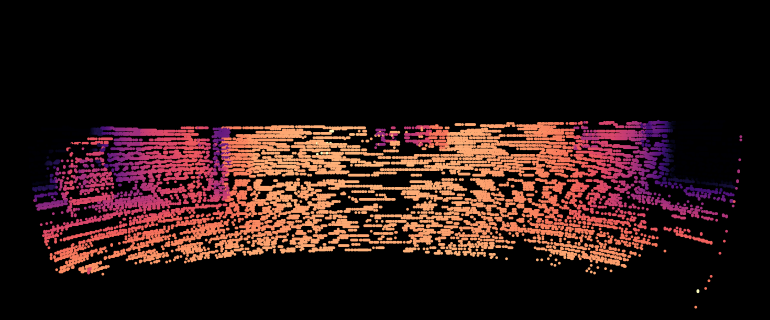}
        \includegraphics[width=0.23\linewidth]{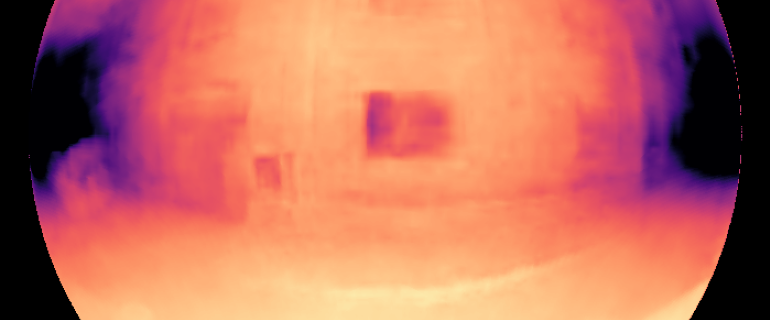}
        \includegraphics[width=0.23\linewidth]{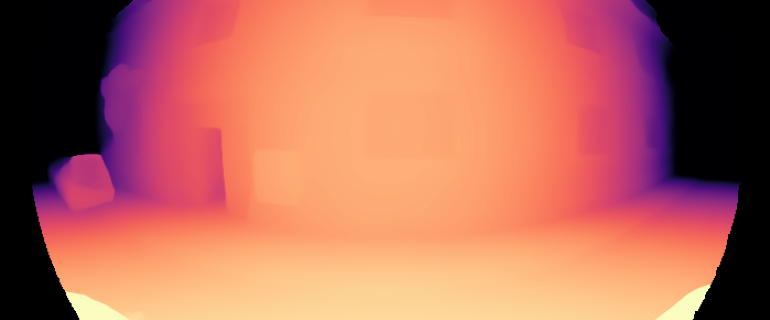}
        \includegraphics[width=0.23\linewidth]{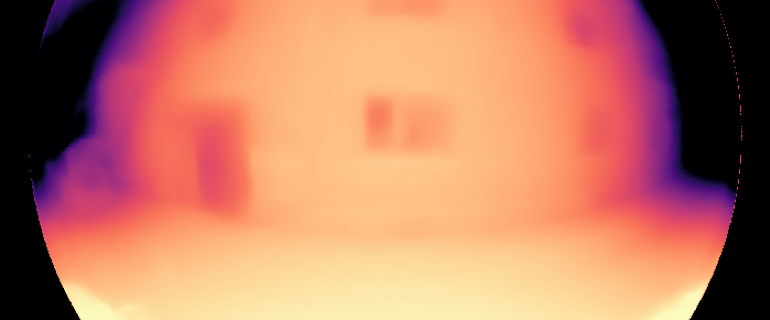}
        \raisebox{1.2ex}{\includegraphics[width=0.05\linewidth]{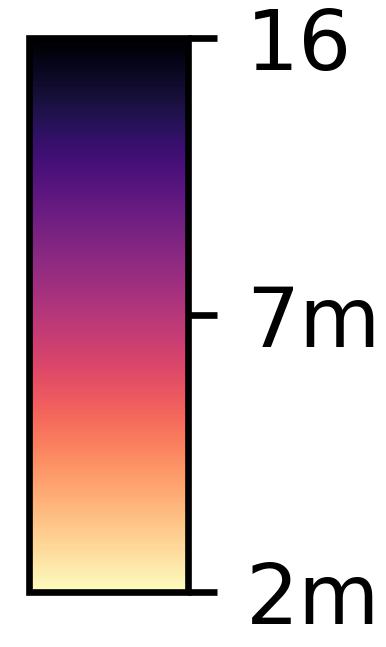}}
    \end{minipage}
    \begin{minipage}[t]{\linewidth}
        \centering
        \includegraphics[width=0.23\linewidth]{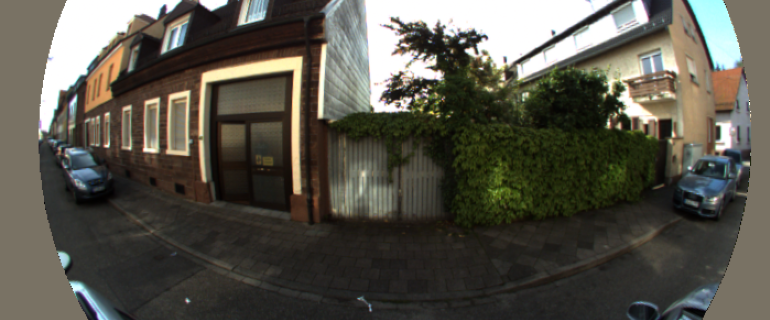}
        \includegraphics[width=0.23\linewidth]{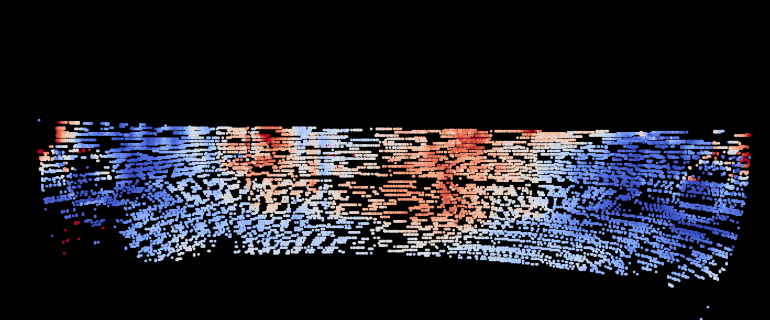}
        \includegraphics[width=0.23\linewidth]{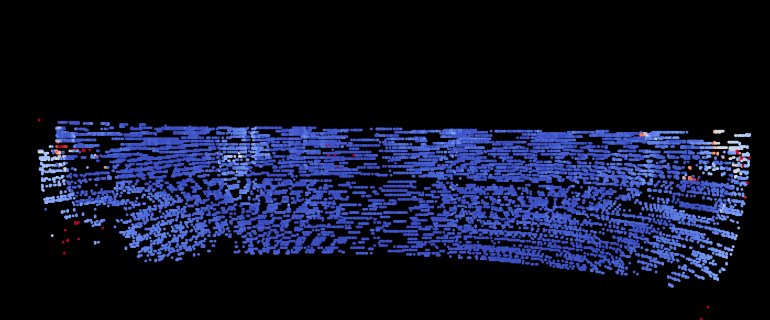}
        \includegraphics[width=0.23\linewidth]{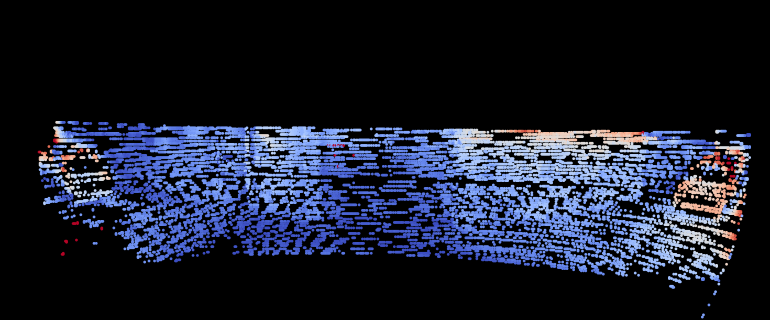}
        \raisebox{1.2ex}{\includegraphics[width=0.05\linewidth]{figures/qualitative/dac/kitti360/arel.jpg}}
    \end{minipage}
    \begin{minipage}[t]{\linewidth}
        \centering
        \includegraphics[width=0.23\linewidth]{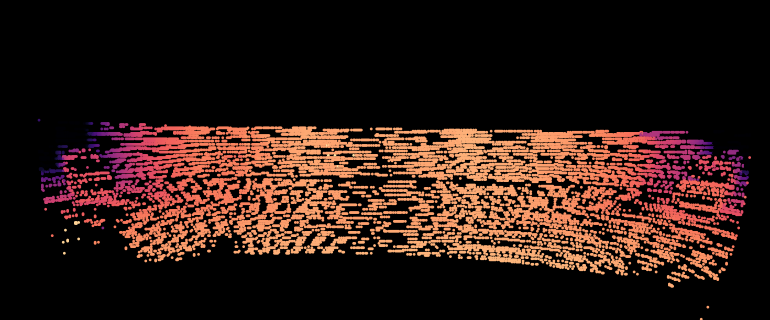}
        \includegraphics[width=0.23\linewidth]{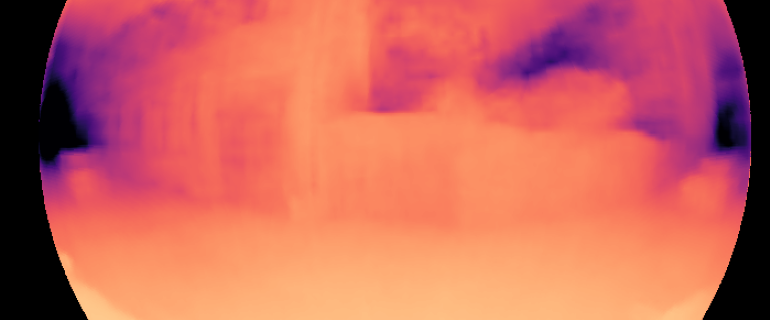}
        \includegraphics[width=0.23\linewidth]{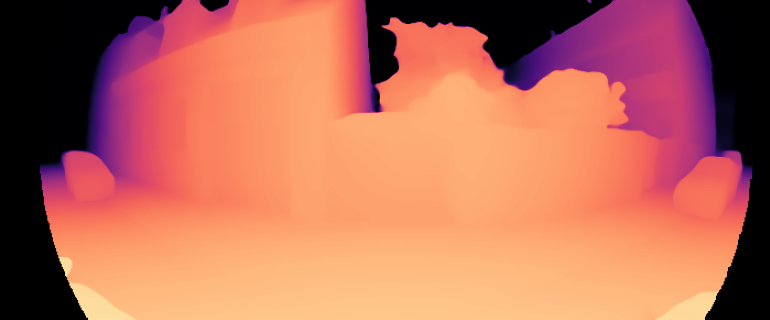}
        \includegraphics[width=0.23\linewidth]{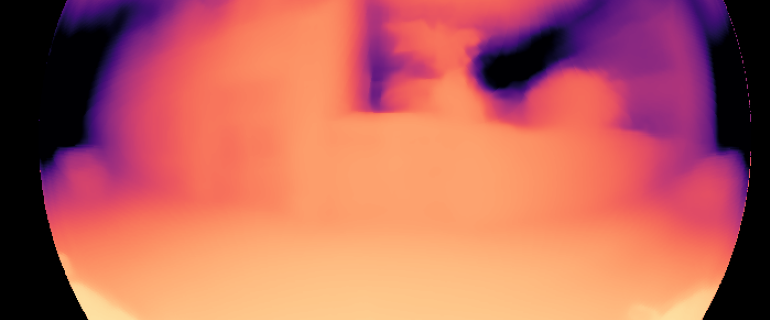}
        \raisebox{1.2ex}{\includegraphics[width=0.05\linewidth]{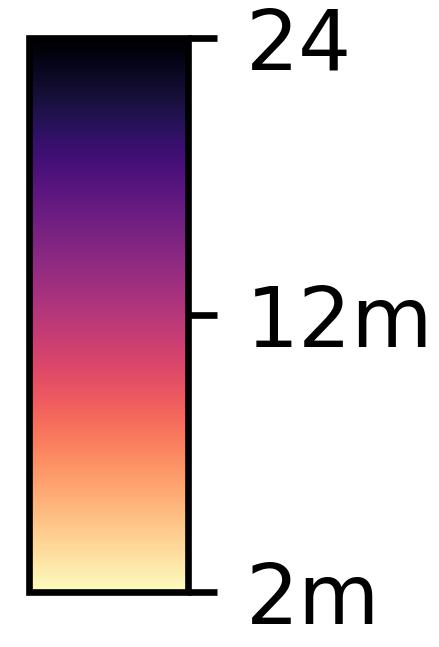}}
    \end{minipage}
    \begin{minipage}[t]{\linewidth}
        \centering
        \includegraphics[width=0.23\linewidth]{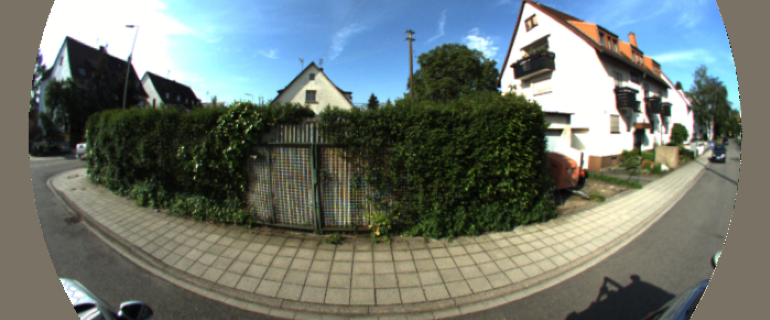}
        \includegraphics[width=0.23\linewidth]{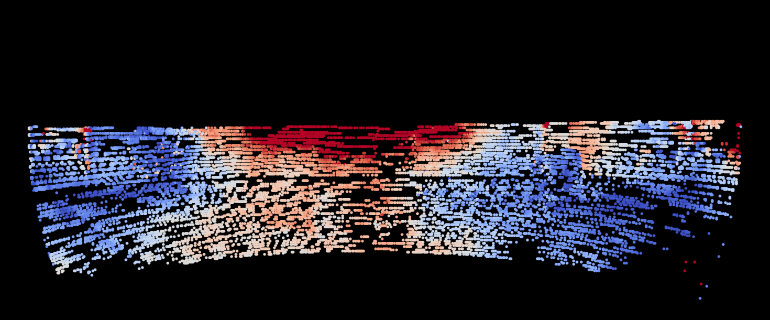}
        \includegraphics[width=0.23\linewidth]{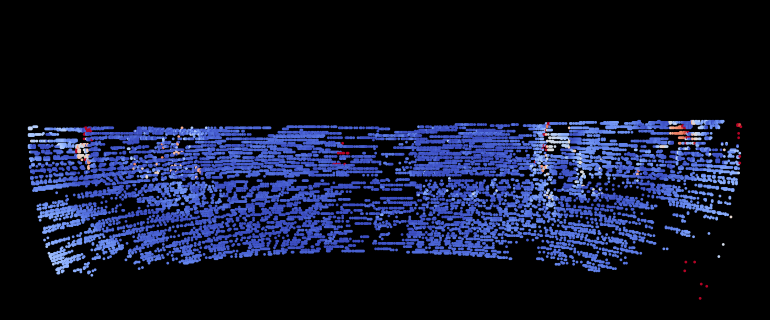}
        \includegraphics[width=0.23\linewidth]{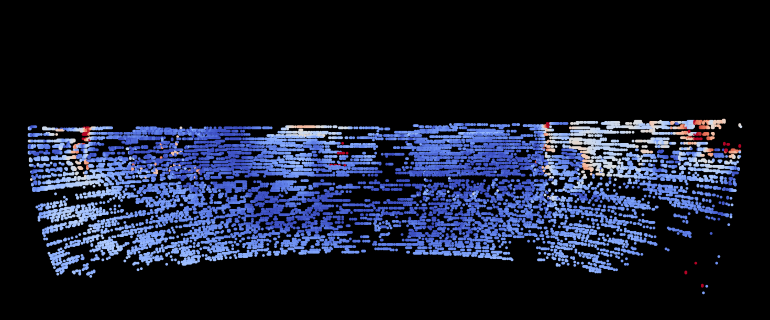}
        \raisebox{1.2ex}{\includegraphics[width=0.05\linewidth]{figures/qualitative/dac/kitti360/arel.jpg}}
    \end{minipage}
    \begin{minipage}[t]{\linewidth}
        \centering
        \begin{tikzpicture}
            \draw (0,0) node[inner sep=0] {\includegraphics[width=0.23\linewidth]{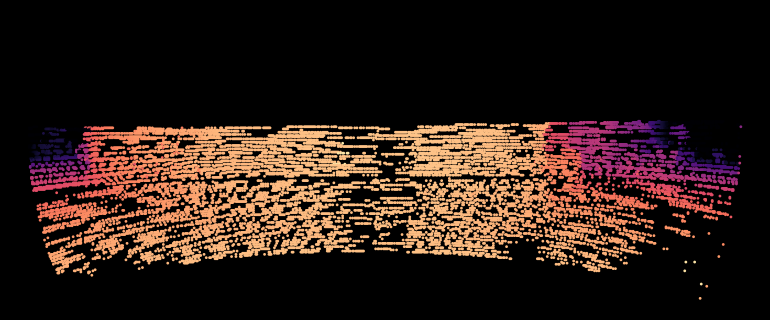}};
            \draw (0,-1.1-0.5) node[inner sep=0, align=center] {\fontsize{8.0}{10}\selectfont {RGB \& GT}};
        \end{tikzpicture}
        \begin{tikzpicture}
            \draw (0,0) node[inner sep=0] {\includegraphics[width=0.23\linewidth]{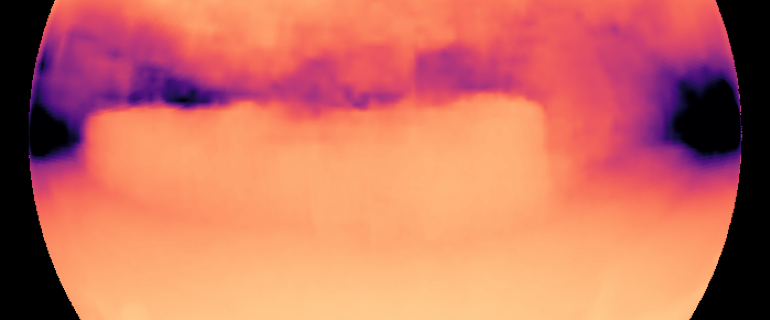}};
            \draw (0,-1.08-0.5) node[inner sep=0, align=center] {\fontsize{8.0}{10}\selectfont {\dacUni~\cite{guo2025depth}}};
        \end{tikzpicture}
        \begin{tikzpicture}
            \draw (0,0) node[inner sep=0] {\includegraphics[width=0.23\linewidth]{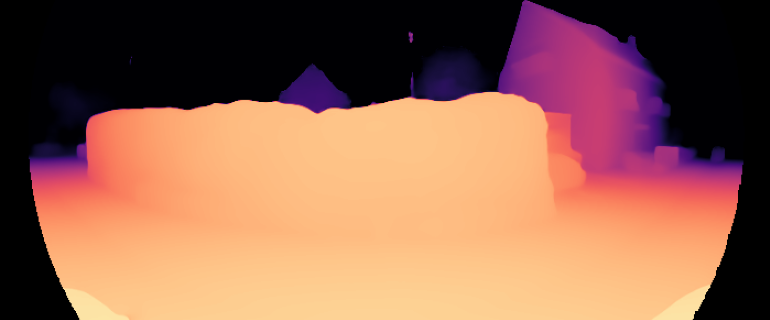}};
            \draw (0,-1.08-0.5) node[inner sep=0, align=center] {\fontsize{8.0}{10}\selectfont {\uniKThreeD~\cite{piccinelli2025unik3d}}};
        \end{tikzpicture}
        \begin{tikzpicture}
            \draw (0,0) node[inner sep=0, align=center] {\includegraphics[width=0.23\linewidth]{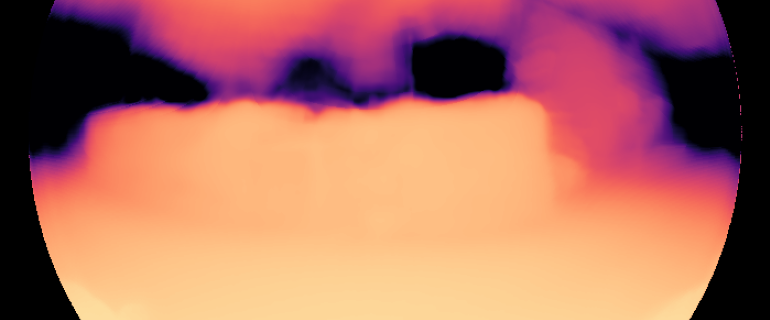}};
            \draw (0,-1.1-0.5) node[inner sep=0, align=center] {\fontsize{8.0}{10}\selectfont {\methodName}};
        \end{tikzpicture}
        \raisebox{7ex}{\includegraphics[width=0.05\linewidth]{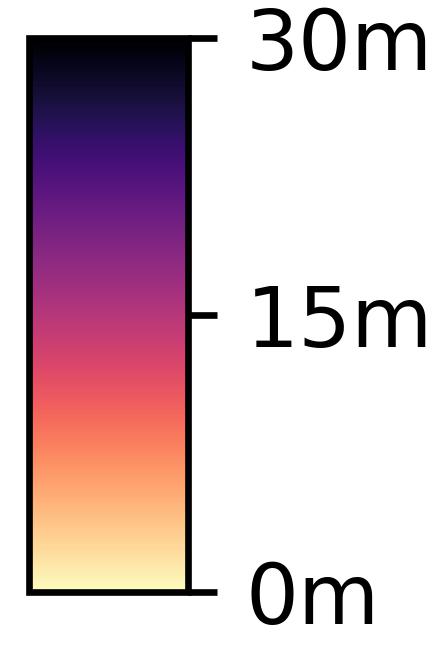}}
    \end{minipage}
    \caption{\textbf{Qualitative Results on \kittiThreeSixty~\cite{liao2022kitti}.} Every pair of consecutive rows corresponds to a single sample. Odd rows display the input RGB image, and \absRel error between predicted and GT depth maps. Even rows display the GT depth map and predicted depth maps.}
    \label{fig:supp_kitti360}
\end{figure*}

We provide additional qualitative results on \scannetpp~\cite{yeshwanth2023scannet++}, \panoGVTwo~\cite{albanis2021pano3d}, and \kittiThreeSixty~\cite{liao2022kitti} for visual comparison in \cref{fig:supp_scannetpp}, \cref{fig:supp_gv2} and \cref{fig:supp_kitti360} respectively.

